\documentclass[acmlarge]{acmart}
\makeatletter
\newcommand{\myconfshort}{\acmConference@shortname}
\newcommand{\myconffull}{\acmConference@name}
\newcommand{\myconfdate}{\acmConference@date}
\newcommand{\myconfloc}{\acmConference@venue}
\AtBeginDocument{
  \fancypagestyle{firstpagestyle}{
    \fancyhead{}%
    \fancyfoot[C]{}%
  }
  \fancyhf{}
  \fancyhead[LO]{\@headfootfont\shorttitle}%
  \fancyhead[RE]{\@headfootfont\@shortauthors}%
  \fancyhead[LE]{\@headfootfont\footnotesize \myconfshort, \myconfdate, \myconfloc}%
  \fancyhead[RO]{\@headfootfont\footnotesize \myconfshort, \myconfdate, \myconfloc}%
  \fancyfoot[C]{}%
}
\makeatother
\acmBooktitle{\conffull\@ (\confshort), \confdate, \confloc}

\AtBeginDocument{%
  }

\copyrightyear{2026}
\acmYear{2026}
\setcopyright{cc}
\setcctype{by}
\acmConference[FAccT '26]{The 2026 ACM Conference on Fairness, Accountability, and Transparency}{June 25--28, 2026}{Montreal, QC, Canada}
\acmBooktitle{The 2026 ACM Conference on Fairness, Accountability, and Transparency (FAccT '26), June 25--28, 2026, Montreal, QC, Canada}
\acmDOI{10.1145/3805689.3812410}
\acmISBN{979-8-4007-2596-8/2026/06}

\hypersetup{
    colorlinks=true,
    urlcolor=blue
}

\usepackage[dvipsnames]{xcolor}
\usepackage[x11names]{xcolor}
\usepackage[table]{xcolor}
\definecolor{red}{RGB}{228, 26, 28}          %
\definecolor{green}{RGB}{55, 165, 65}        %
\definecolor{yellow}{RGB}{255, 180, 0}       %
\definecolor{lightgreen}{RGB}{140, 240, 180} %
\usepackage{pifont}
\usepackage{subcaption} 

\usepackage{xspace}
\newcommand{\cface}{\textsc{CounterFace}\xspace}

\usepackage{microtype}
\usepackage{graphicx}
\usepackage{booktabs} 
\usepackage{bm}
\usepackage{mathtools}
\usepackage{xfrac}
\usepackage{graphicx}
\usepackage{subcaption}
\usepackage{tikz}
\usepackage{amsmath,amsfonts,bm}

\usepackage{amssymb,amsthm}
\usepackage{newfloat}
\usepackage{booktabs}
\usepackage{soul}
\usepackage{multirow}
\usepackage{microtype}
\usepackage{tcolorbox}

\newcommand{\refsection}[2]{Section\ref{#1}~`#2'}

\newcommand{\reffigure}[1]{Figure~\ref{#1}}

\newcommand{\attribute}{\mathsf{a}}
\newcommand{\attributes}{\vec{a}}
\newcommand{\attributeSpace}{\mathcal{A}}
\newcommand{\attributeInducement}{\mathrm{g}}

\renewcommand{\vec}{\bm}

\DeclarePairedDelimiterX{\infdivx}[2]{(}{)}{%
  #1\;\delimsize\|\;#2
}

\newcommand{\sampleElement}{x}
\newcommand{\labelElement}{y}

\newcommand{\labelSpace}{\mathcal{Y}}

\usepackage{dsfont}

\DeclarePairedDelimiter{\abs}{\lvert}{\rvert}

\newcommand{\sample}{\vec{\sampleElement}}

\newcommand{\realNumbers}{\mathbb{R}}

\newcommand\MTkillspecial[1]{%
\bgroup
\catcode`\&=9
\let\\\relax%
\scantokens{#1}%
\egroup
}
\DeclarePairedDelimiter\brparen
\lparen\rparen
\reDeclarePairedDelimiterInnerWrapper\brparen{star}{
\mathopen{#1\vphantom{\MTkillspecial{#2}}\kern-\nulldelimiterspace\right.}
#2
\mathclose{\left.\kern-\nulldelimiterspace\vphantom{\MTkillspecial{#2}}#3}}

\usepackage{clipboard}

\makeatletter
\newcommand\RedeclareMathOperator{%
  \@ifstar{\def\rmo@s{m}\rmo@redeclare}{\def\rmo@s{o}\rmo@redeclare}%
}
\newcommand\rmo@redeclare[2]{%
  \begingroup \escapechar\m@ne\xdef\@gtempa{{\string#1}}\endgroup
  \expandafter\@ifundefined\@gtempa
     {\@latex@error{\noexpand#1undefined}\@ehc}%
     \relax
  \expandafter\rmo@declmathop\rmo@s{#1}{#2}}
\newcommand\rmo@declmathop[3]{%
  \DeclareRobustCommand{#2}{\qopname\newmcodes@#1{#3}}%
}
\@onlypreamble\RedeclareMathOperator
\makeatother

\RedeclareMathOperator{\Pr}{{\mathbb{P}}}

\usepackage{hyperref}
\usepackage[utf8]{inputenc} %
\usepackage[T1]{fontenc}    %
\usepackage{url}            %
\usepackage{booktabs}       %
\usepackage{amsfonts}       %
\usepackage{nicefrac}       %
\usepackage{microtype}      %
\usepackage{xcolor}         %
\usepackage{graphicx}
\usepackage{cleveref}
\usepackage{enumitem}
\usepackage{makecell}
\usepackage{tablefootnote}
\usepackage{titletoc} 

\begin{document}
\title{\cface: A Synthetic Face Dataset for Fine-Grained Counterfactual Evaluation of Face Recognition Systems}

\author{Guruprasad Viswanathan Ramesh}
\email{viswanathanr@wisc.edu}
\affiliation{%
  \institution{University of Wisconsin-Madison}
  \city{Madison}
  \state{WI}
  \country{USA}
}
\author{Ashish Hooda}
\authornote{Now at Google DeepMind}
\email{ahooda@wisc.edu}
\affiliation{%
  \institution{University of Wisconsin-Madison}
  \city{Madison}
  \state{WI}
  \country{USA}
}

\author{Shimaa Ahmed}
\email{shiahmed@visa.com}
\affiliation{%
  \institution{Visa Research}
  \city{Foster City}
  \state{CA}
  \country{USA}
}

\author{Harrison J Rosenberg}
\email{hrosenberg@ece.wisc.edu}
\authornote{Now at ALL3D, Inc.}

\author{Ramya Korlakai Vinayak}
\email{ramya@ece.wisc.edu}

\author{Kassem Fawaz}
\email{kfawaz@wisc.edu}
\affiliation{%
  \institution{University of Wisconsin-Madison}
  \city{Madison}
  \state{WI}
  \country{USA}
}

\renewcommand{\shortauthors}{Ramesh et al.}
\renewcommand{\shorttitle}{\cface: Synthetic Counterfactual Dataset for Face Recognition Evaluation}

\begin{abstract}
Face recognition (FR) systems are widely deployed in critical applications, making their reliability and robustness across diverse populations and conditions essential. Standard evaluation of FR systems typically relies on datasets such as LFW to estimate average recognition accuracy. Some benchmarks also capture coarse-grained intra-identity variations such as aging, pose, and lighting. However, human faces undergo more fine-grained changes, including appearance changes such as hairstyles and makeup, that are underrepresented in existing benchmarks. Counterfactual evaluation provides a method to assess FR robustness under such fine-grained variations. Existing counterfactual face datasets synthesized with image generators, however, are limited in attribute coverage due to the use of humans for verification in the pipeline. 

We propose \cface, a new counterfactual evaluation dataset comprising 20 facial attributes and 8 demographic factors, exceeding prior synthetic face datasets by 14 attributes and 2 demographics. The dataset is generated using a fully automated pipeline based on off-the-shelf image generators with custom verifiers, removing human need for verification. \cface contains 11,821 counterfactual face pairs, and a post-hoc user study confirms the faithfulness of the generated counterfactuals. We evaluate two commercial and four open-source FR systems~(AWS Rekognition, Face++, AdaFace, MagFace, ArcFace, FaceNet) across 160 attribute–demographic combinations. Our dataset helps in the isolation of precise failure modes for individual systems unlike standard evaluation benchmarks. Results indicate that the performance degradation varies across attributes and demographics for all six systems and occluding attributes~(e.g., facemask and facial hair) universally degrade performance.

\end{abstract}



\begin{abstract}
Face recognition (FR) systems are widely deployed in critical applications, making their reliability and robustness across diverse populations and conditions essential. Standard evaluation of FR systems typically relies on datasets such as LFW to estimate average recognition accuracy. Some benchmarks also capture coarse-grained intra-identity variations such as aging, pose, and lighting. However, human faces undergo more fine-grained changes, including appearance changes such as hairstyles and makeup, that are underrepresented in existing benchmarks. Counterfactual evaluation provides a method to assess FR robustness under such fine-grained variations. Existing counterfactual face datasets synthesized with image generators, however, are limited in attribute coverage due to the use of humans for verification in the pipeline. 

We propose \cface, a new counterfactual evaluation dataset comprising 20 facial attributes and 8 demographic factors, exceeding prior synthetic face datasets by 14 attributes and 2 demographics. The dataset is generated using a fully automated pipeline based on off-the-shelf image generators with custom verifiers, removing human need for verification. \cface contains 11,821 counterfactual face pairs, and a post-hoc user study confirms the faithfulness of the generated counterfactuals. We evaluate two commercial and four open-source FR systems~(AWS Rekognition, Face++, AdaFace, MagFace, ArcFace, FaceNet) across 160 attribute–demographic combinations. Our dataset helps in the isolation of precise failure modes for individual systems unlike standard evaluation benchmarks. Results indicate that the performance degradation varies across attributes and demographics for all six systems and occluding attributes~(e.g., facemask and facial hair) universally degrade performance.

\end{abstract}

\maketitle

\section{Introduction}

\begin{figure*}
    \centering
    \includegraphics[width=\textwidth]{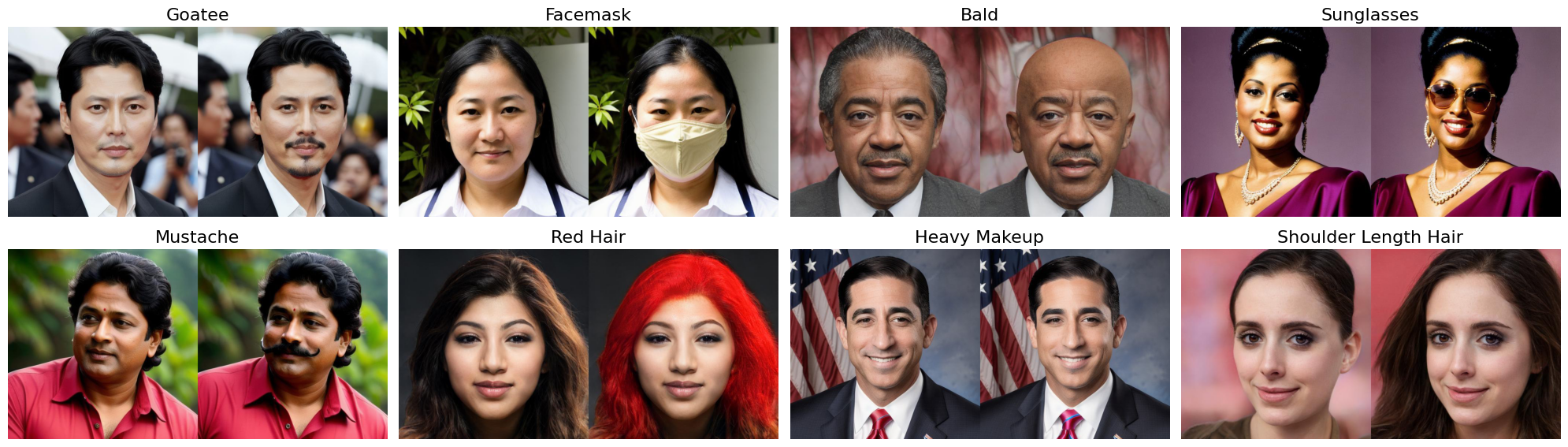}
    \caption{Examples of counterfactual face pairs in our dataset \cface. The text appearing above is the attribute applied or removed from the source face. The first face in each pair is without the attribute, and the second face is with the attribute.}
    \label{fig:transformed_faces}
\end{figure*}

Commercial and Government organizations are rapidly and widely adopting Face Recognition (FR) systems for retail, surveillance, social media, and airport check-ins~\cite{jacobson2023future,tsa2024facial,riordan2024rise,lerner2024retailers}. However, past studies show that these systems often exhibit bias against  underrepresented groups~\cite{grother2017bias,grother2019face,cook2019demographic} and sensitivity to factors such as lighting, pose, and skin-tone~\cite{balakrishnan2021towards,terhorst2021maad,dhar2021pass}. Such findings have led industry, ISO~\cite{ISO19795-10}, and NIST~\cite{temoshok2024digital} to recommend testing FR \emph{performance} across a wide range of demographic, semantic, and environmental factors. Therefore, there is a need for \emph{fine-grained diagnostics} of FR systems to identify and explain their failure modes.

Counterfactual evaluation enables fine-grained analysis of trained models for assessing their robustness, biases, and explainability. It has been used to evaluate ML models across several domains~\cite{le2023coco, pmlr-v235-hooda24a, howard2024socialcounterfactuals}. A counterfactual example is obtained by making targeted
but human-understandable changes to input data to characterize the model’s unintended correlation to irrelevant features and its failure modes~\cite{kusner2017counterfactual}. In the context of FR, counterfactual examples can be obtained by applying or removing facial attributes on a given face, as evident in~\Cref{fig:transformed_faces}. As such, these examples can help answer the question “Would a light-skinned male be identified by a face recognition system as the same person when wearing heavy makeup?” Compared to standard benchmarks, which do not represent such fine-grained contexts~\cite{ethayarajh2020utility}, counterfactual examples may be constructed to meet targeted requirements. However, collecting real and natural counterfactual face pairs differing only in a targeted facial attribute is impractical at scale~\cite{liang2023benchmarking} due to the prohibitive cost of modifying targeted attributes across a large population while controlling for confounding factors~(e.g., change in pose or lighting). In contrast, synthetic counterfactual face data offers a viable alternative for performing fine-grained evaluation of face recognition and analysis systems~\cite{denton2019image,balakrishnan2021towards, liang2023benchmarking}. 

Synthetic counterfactual examples can be obtained directly using generative models, such as GANs~\cite{goodfellow2014generative} and diffusion models~\cite{rombach2021highresolution}, or by augmenting generative models with editing techniques~\cite{brack2023sega, patashnik2021styleclip, shen2020interfacegan}. These models, however, may fail to apply the requested edit, introduce unrelated changes, change the identity, or degrade the quality of the original face~(see~\Cref{fig:ip2p_fail}). Moreover, they struggle with counterfactuals that are naturally uncommon yet possible, such as facial hair on female faces. Previous attempts to curate synthetic datasets either ignore these failures by not verifying the edits~\cite{denton2019image} or rely on human annotations to select compliant samples~\cite{balakrishnan2021towards, liang2023benchmarking}. The former yields noisy data, while the latter is costly and limits the attribute and demographic coverage of the resulting datasets. In this paper, we address these shortcomings by constructing an automatically verified synthetic dataset and use it to perform counterfactual evaluation of six popular commercial and open-source FR systems.

We construct \textbf{\cface}~(\Cref{fig:transformed_faces}), a new counterfactual dataset for faces using an \emph{automated pipeline} based on Rejection Sampling~(RS). Our automated pipeline consists of 1) off-the-shelf diffusion and GAN models to generate \emph{candidates} for faces 2) custom artifact and attribute verifiers (detectors) to select counterfactual face pair datasets from the candidates based on three strict counterfactual requirements~(\Cref{sec:requirements}). It addresses the shortcomings of previous synthetic counterfactual face datasets~\cite{balakrishnan2021towards, liang2023benchmarking} by (1) removing the need for manual verification, and (2) introducing artifact detectors to reject candidate faces with unnatural artifacts and distortions. 

We use the automated pipeline to generate a dataset of 11,821 face pairs, spanning 20 attributes and 8 demographics~(75 face pairs for 153 of the 160 attribute-demographic combinations), which is 14 more attributes and 2 more demographics compared to prior counterfactual face datasets, which relied on human verification~\cite{balakrishnan2021towards,liang2023benchmarking}. We perform a post-hoc user study~(n=900) to assess the quality of a subset of images generated by our automated framework. Across the 1583 image pairs surveyed: 96.9\% retained identity and 84.04\% met the counterfactual requirements. Unlike previous work focusing on counterfactual face datasets that focused on just open-source FR models~\cite{liang2023benchmarking} or facial attribute classifiers~\cite {balakrishnan2021towards}, we evaluate both commercial and open-source FR systems: AWS Rekognition, Face++, MagFace~\cite{meng2021magface}, AdaFace~\cite{kim2022adaface}, FaceNet~\cite{schroff2015facenet}, and ArcFace~\cite{deng2019arcface}~(\Cref{sec:fr_results}). 

Our dataset enables the precise identification of 160 attribute-demographic combinations~(largest number of fine-grained combinations to the best of our knowledge) where FR systems struggle. We pinpoint these failure modes across six different systems. We also understand the role of model architecture and training data size in generalization to the counterfactual changes applied to original faces. Our analyses show that occluding attributes, such as facemask and sunglasses, consistently degrade performance across all six systems, though the extent of this drop varies across demographics. Finally, our ablation studies demonstrate the necessity of minimizing confounding factors when generating counterfactual data for reliable FR evaluation.

\section{Background and Related Work}
In the following section, we present related work on face generation architectures, face recognition fairness and failure, and the applications of rejection sampling in data generation.

\subsection{Synthetic Face Generation Architectures}
GANs~\cite{goodfellow2014generative} and latent diffusion models~\cite{rombach2021highresolution} are prominent architectures for image synthesis, including facial images. StyleGAN~\cite{karras2019style, Karras2019stylegan2}, in particular, has been extensively used to synthesize faces~\cite{shen2018faceid, shen2020interfacegan, qiu2021synface,kammoun2022generative}. Methods like InterfaceGAN~\cite{shen2020interfacegan} and StyleCLIP~\cite{patashnik2021styleclip} modify the latent space of GANs to manipulate facial semantics, such as pose~\cite{siarohin2018deformable, 8373839}, age~\cite{antipov2017face, song2018dual}, and expressions~\cite{yi2018data}. However, GAN-generated faces often exhibit limited variability~\cite{colbois2021use}.

Text conditioning in diffusion models enables users to describe desired images via textual prompts, allowing for the generation of faces with varying styles, poses, accessories, expressions, and ages~\cite{kim2023dcface,melzi2023gandiffface,banerjee2023identity}. Moreover, diffusion editing techniques~\cite{brack2023sega, avrahami2022blended, brooks2022instructpix2pix} enable semantic modifications of existing images. However, such methods may introduce unintended edits, alter facial identities, and are less robust for fine-grained editing scenarios~\cite{huang2024diffusion}.

\subsection{Face Recognition Fairness and Failures}
Face recognition (FR) systems have largely been studied for demographic performance disparities ~\cite{albiero2020analysis, buolamwini2018gender, klare2012face, phillips2003face, drozdowski2020demographic, abdurrahim2018review}. Failure studies along other semantic attributes are also common like age~\cite{albiero2020does,klare2012face,lu2019experimental}, pose~\cite{schyns2019viewpoint,kortylewski2018empirically}, and hair~\cite{lu2019experimental,bhatta2023gender,wu2024facial}. However, such evaluations are rarely conducted in a \emph{counterfactual} setting, where the impact of a specific attribute is isolated through controlled and meaningful changes, but rather through sensitive attribute group evaluation.

Denton et al.\cite{denton2019image}, Balakrishnan et al.\cite{balakrishnan2021towards}, and Liang et al.\cite{liang2023benchmarking} conduct GAN-based counterfactual evaluation of attribute classifiers~\cite{denton2019image,balakrishnan2021towards} and FR systems~\cite{liang2023benchmarking}. However, their counterfactual pipelines lack edit verification~\cite{denton2019image} or rely on manual annotation~\cite{balakrishnan2021towards, liang2023benchmarking}, which is costly and unscalable. Thus, their evaluation is limited to a small number of attributes, while also accounting for only those attributes for verifying the edits for counterfactual correctness. We include more details of the two datasets and their differences to ours in \Cref{sec:comparison_with_other_datasets}.

\begin{table}[t]
  \centering
  \small
  \resizebox{\textwidth}{!}{%
    \begin{tabular}{@{} l l c c c c c c r @{}}
      \toprule
      Dataset & Type & Size & \# Attributes & Control & Verification & Demographics & Open Sourced & \# Identities \\
      \midrule
      LFW~\cite{huang2008labeled}             & Real      & 13,000                 & $\times$  & $\times$  & NA      & –   & $\checkmark$ & 5,700   \\
      CelebA~\cite{liu2015faceattributes}      & Real      & 200\,k                 & 40        & $\times$  & NA      & –   & $\checkmark$ & 10,000  \\
      VGGFace2~\cite{cao2018vggface2}          & Real      & 3.3\,M                 & $\times$  & $\times$  & NA      & –   & Retracted    & 9,000   \\
      Transect~\cite{balakrishnan2021towards}  & Synthetic & 8,000                  & 6         & $\checkmark$ & Human   & 4     & $\times$ & 1000  \\
      CausalFaces~\cite{liang2023benchmarking} & Synthetic & $19,200^{*}$ & 4         & $\checkmark$ & Human   & 6      & $\checkmark$ & $400^{*}$     \\
      \textbf{\cface}                           & Synthetic & 11,821                 & 20        & $\checkmark$ & Automated & 8 & $\checkmark$ (for research) & 1,166   \\
      \bottomrule
    \end{tabular}%
  }
  \caption{Comparison of existing real and synthetic faces datasets. Real datasets lack controlled edits and do not support counterfactual evaluation. Other synthetic datasets~(Transect and CausalFaces) for counterfactual evaluation have limited attribute and demographics. Our dataset \cface spans 20 attributes, 8 demographics enabling evaluation for a lot more attribute-demographic combinations. Refer to \Cref{sec:comparison_with_other_datasets} for more details on dataset differences~(* - exact size of dataset not clear as explained in \Cref{sec:comparison_with_other_datasets}).}
    \label{tab:dataset_comparison}
\end{table}

\subsection{Rejection Sampling}
Rejection Sampling~(RS) is a technique for drawing samples from a hard-to-sample distribution by first generating candidates from an easier proposal distribution and then keeping only those candidates that meet an acceptance criterion~\cite{rs-gilks-and-wild}. RS has been used to improve target distribution learning with image generative models~\cite{grover2018variational, azadi2018discriminator,na2024diffusion}, fine-tuning Large Language Models with human preference~\cite{touvron2023llama, nakano2021webgpt,cobbe2021training}, and mitigating disparities in ML models~\cite{erfanian2024chameleon,ramachandranpillai2024bt}. It has also been implemented in the automated pipelines of counterfactual dataset generation in other domains~\cite{howard2024socialcounterfactuals, le2023coco}, where CLIP~\cite{radford2021learning} based verifiers are used to reject samples not satisfying requirements such as detecting the presence of an object in an image, a global attribute. However, to understand local attributes, such as facial attributes, more sophisticated verifiers are needed due to CLIP's drawback in understanding low-level details in an image~\cite{xie2025fg, li2022grounded, zhong2022regionclip, zhang2024long}, including facial attributes~\cite{rosenberg2023unbiased}. We also present an empirical analysis highlighting this limitation~\Cref{sec:clip_failure}.

\section{\cface Dataset Generation}
\label{sec:framework}
We utilize an RS-based automated pipeline to construct \cface based on off-the-shelf GAN and diffusion models and customized artifact and attribute verifiers. Below, we introduce the notation used in the paper. Next, we outline the requirements for generating counterfactual data. Subsequently, we describe the image generators and verifiers used in our data generation pipeline. Finally, we discuss the details of our dataset. Due to space constraints, we defer the low-level details to ~\Cref{sec:design_choices} in the appendix.

\subsection{Notation}
\label{sec:notation}

Let $\sample \in \realNumbers^{H\times W\times 3}$ denote an RGB image. Typically, $\sample$ depicts a face containing identity $\labelElement$ within the set of identities $\labelSpace$. Each face image is associated with a set of semantic attributes $\attributes$ within the space of semantic attributes $\attributeSpace$. For the scope of this paper, we assume all attributes are binary.  Attribute vector $\attributes_{\sample} \in \{0,1\}^{\attributeSpace}$ denotes the presence or absence of attributes $\attribute_1,\dotsc,\attribute_{\abs{\attributeSpace}}$ in image $\sample$. We use attribute transformation function $\attributeInducement_{\attribute_i}: \realNumbers^{H\times W\times 3} \rightarrow \realNumbers^{H\times W\times 3}$ to modify an attribute $\attribute_i$ on a face $\sample$. The modification $\attributeInducement_{{\attribute_i}}$ will be the application of attribute $\attribute_i$ if $\attribute_i$ is not present in $\sample$, i.e., $\attributes_{\sample}[i] = 0$. And it will be the removal of attribute $\attribute_i$ if $\attribute_i$ is already present in $\sample$, i.e., $\attributes_{\sample}[i] = 1$.

\subsection{Counterfactual Requirements}
\label{sec:requirements}

Prior work has established a set of requirements (\textit{validity}, \textit{correctness}, and \textit{specificity}) for counterfactual examples~\cite{verma2022counterfactual, abid2022meaningfully}. We adopt these requirements for generating a counterfactual example of a face. These requirements together are essential for selecting and rejecting a candidate sample in our pipeline. The requirements are: 

\noindent
\textbf{Validity:} 
Counterfactual examples need to be \textbf{valid} by satisfying real-world constraints. For faces, modifying attribute $\attribute_i$ of a  face $\sample$ should keep $\attributeInducement_{\attribute_i}(\sample)$ semantically correct by not introducing unnatural artifacts, retaining identity $\labelElement$ of $\sample$.

\noindent
\textbf{Correctness:}
A counterfactual example is \textbf{correct} if it satisfies an intended result. In our case, the counterfactual face should correctly reflect the modified semantic attribute $\attribute_i$. While applying $\attribute_i$ to $\sample$, the $i^{\text{th}}$ element of the attribute vector of $\attributeInducement_{\attribute_i}(\sample)$ should be 1, i.e., $\attributes_{\attributeInducement_{\attribute_i}(\sample)}[i] = 1$. Similarly, while removing $\attribute_i$ from $\sample$, the $i^{\text{th}}$ element of the attribute vector of $\attributeInducement_{\attribute_i}(\sample)$ should be 0, i.e., $\attributes_{\attributeInducement_{\attribute_i}(\sample)}[i] = 0$.

\noindent
\textbf{Specificity:}
The counterfactual example should have only the intended attribute being altered, while all other attributes remain unchanged. That is to say, $\attributeInducement_{\attribute}(\sample)$ should only change attribute $\attribute$, but no other enumerated semantic attributes; i.e., $\attributes_{\attributeInducement_{\attribute_i}(\sample)}[j] = \attributes_{\sample}[j] \quad \forall \; j \neq i.$

\begin{figure*}[t]
 \centering
 \includegraphics[width=\textwidth]{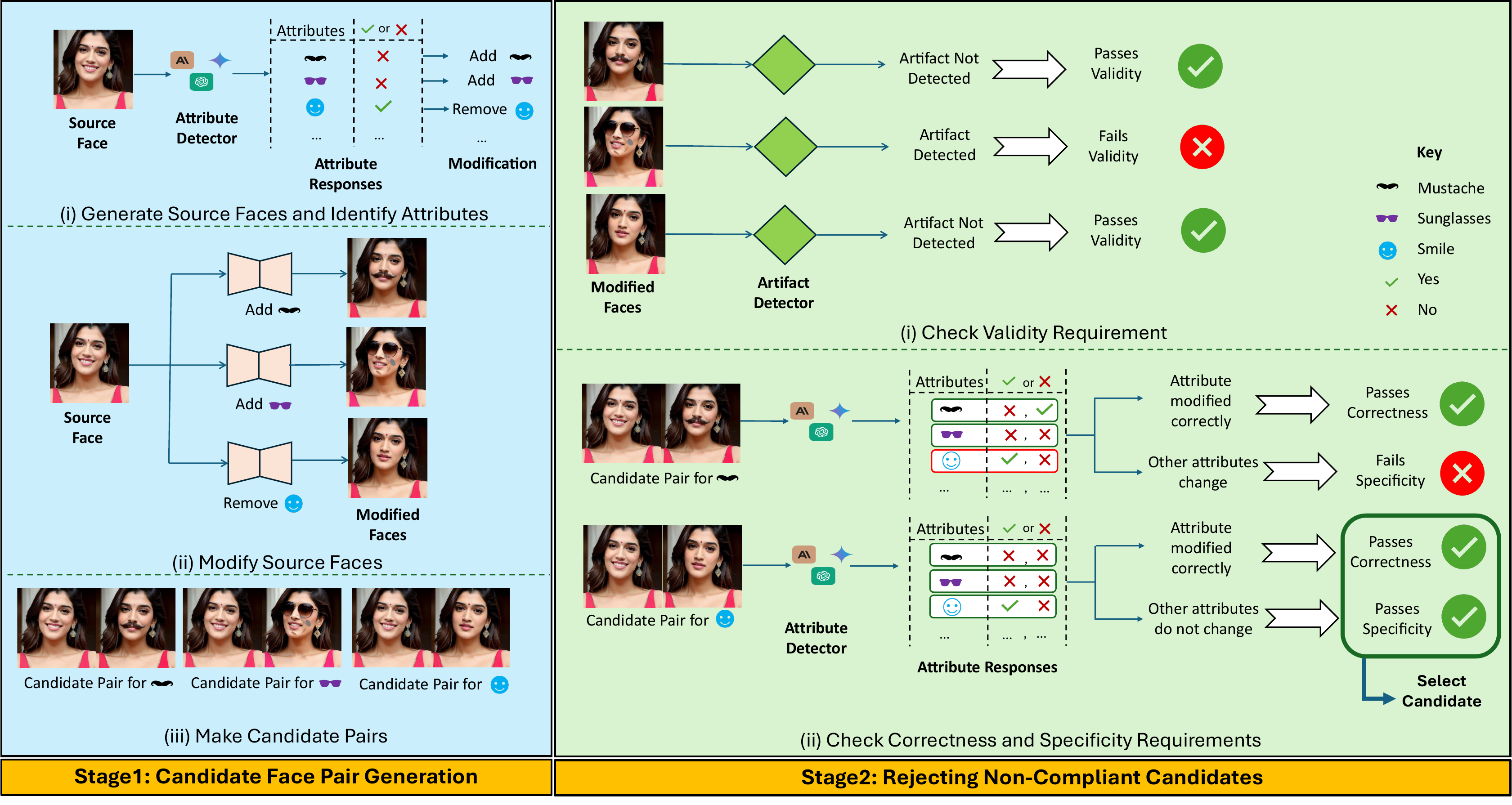}
 \caption{\cface Generation Pipeline: In the first stage, candidate face pairs are generated using off-the-shelf generative models. In the second stage, candidates are checked for three counterfactual requirements~(Validity, Correctness, and Specificity) using an artifact and an attribute verifiers. Only candidates meeting all three requirements are retained for the dataset.}
 \label{fig:pipeline}
\end{figure*}

\subsection{Rejection Sampling Pipeline for Generating Counterfactual Faces}

Our dataset pipeline utilizes off-the-shelf generative models and editing techniques to generate counterfactual face pairs by modifying a single attribute. As described earlier, existing image generation and editing techniques face challenges in producing faces for uncommon attribute-demographic combinations. Off-the-shelf generation models frequently fail to apply desired attributes correctly, inadvertently alter other facial features, or generate unnatural artifacts. Our key insight is that these failures are nondeterministic and do not occur in every instance.

Thus, we develop a rejection sampling pipeline~(\Cref{fig:pipeline}) with automated verifiers to ensure that only modified faces meeting the aforementioned counterfactual requirements are accepted. The pipeline comprises two stages: 1) Generating candidate face pairs, and 2) Rejecting noncompliant candidates. In the first stage, we use off-the-shelf generative models to generate a source face and subsequently modify it to add or remove an attribute using an editing technique. The resulting pairs serve as candidates for the second stage, where we employ artifact and attribute verifiers for each attribute-demographic combination by combining different machine learning models and use these verifiers to check if the modified face meets the three counterfactual requirements. If it passes the requirements, we add it to \cface. The main improvement in our dataset pipeline comes from the use of automated verifiers that remove human intervention, enabling better scalability and reducing costs compared to previous counterfactual face datasets.

\subsubsection{Stage 1: Generating candidate face pairs}
The output of the first stage is a large set of \emph{candidate face pairs} across all attribute-demographic combinations, where each pair consists of a source face and its modification, i.e., a face with and without a target attribute. Below, we describe the steps we perform to generate the candidate dataset.

While source faces can come from natural datasets or generative models (e.g., GANs or diffusion), we use synthesized faces to remain consistent with existing counterfactual datasets~\cite{balakrishnan2021towards,liang2023benchmarking}. If the source face contains the target attribute, the modification function (an image editing technique) removes it; otherwise, the function adds the attribute. For instance, if the target attribute is `shoulder length hair’, the function lengthens or shortens the hair based on the source state, without altering identity or introducing distortions. We implement this stage by (1) creating source faces of multiple identities across different demographics, (2) generating multiple source faces for each identity~(referred to as variations), and (3) modifying semantic attributes of the source faces. The final output is a set of candidate face pairs that become input to the verifiers in the second stage.

To generate source faces, we follow the methodology of Rosenberg et al.~\cite{rosenberg2023unbiased}. As they demonstrate, naive prompts with demographic give faces with limited diversity and identity consistency. To overcome this, we compile 150 celebrity names for each of the 8 demographics using GPT-4o~\cite{openai2024gpt4o}. Using the prompt ``A photo of the face of \(\langle \text{Name} \rangle\)'', we generate faces with Realistic Vision~\cite{rombach2021highresolution, Realistic_Vision_V4_0_noVAE}. This helps in generating identity-consistent source faces with high inter-class variance. Varying the random seed produces six variations per identity, totaling 900 source faces per demographic.

We manually validate the variations for identity consistency and remove a small subset of identities where different variations did not adhere to one source identity or demographic. This ensures that only variations maintaining a consistent identity are chosen for obtaining counterfactuals. This validation step is performed by two of the authors and takes about an hour to check all the identities across all demographics. (see \Cref{sec:design_choices} for further details)

Next, we modify source faces using either SEGA~\cite{brack2023sega}
(a diffusion-based approach) or StyleCLIP~\cite{patashnik2021styleclip} (a GAN-based approach). This method of using one of the two approaches was due to our observation that some attributes were supported better by one method than the other. Selecting an appropriate method for each attribute ensures that the candidate faces are of suitable quality to get the intended samples in the rejection sampling stage.

We manually test both methods on all 20 attributes and select SEGA for modifying 7 attributes, StyleCLIP for 4 attributes, and either of them for the remaining 9 attributes~(\Cref{tab:attr_and_editing_tech}). When using StyleCLIP, we invert the source face generated by the diffusion model to the GAN latent space using e4e~\cite{tov2021designing}; in this case, the source face becomes the reconstructed image. We carefully tune the hyperparameters for each method on a small validation set to ensure the intended attribute is applied with no or little change to other attributes while maintaining identity consistency~(\Cref{sec:design_choices}). This calibration of hyperparameters, performed by the authors, requires approximately 5 to 10 minutes per attribute. Finally, we end up with a candidate face pair dataset containing 900 pairs of source and modified faces for each of the 8 demographics and  20 attributes; a total of 144,000 pairs.

\subsubsection{Stage 2: Rejecting Noncompliant Candidates}
We evaluate each candidate face pair against the three counterfactual requirements using the verifiers with rejection sampling. First, an artifact verifier examines the pair for unnatural distortions introduced during editing, ensuring \emph{Validity} requirement. Subsequently, attribute verifiers analyze the presence of all attributes in both faces. For \emph{Correctness}, we verify that the target attribute is successfully modified (added or removed) relative to the source face. For,  \emph{Specificity}, we compare the detector responses for all non-target attributes; we require these attributes to remain consistent between the source and modified faces, allowing changes only for any related attributes as defined in the transition matrix~(\Cref{fig:transition-matrix-conditioning}). We pass candidate face pairs to sequence of these steps for each attribute-demographic combination until we achieve the desired target of 75 counterfactual pairs for that combination. Below, we describe the verifiers used in the stage to filter the candidate dataset obtained in the first stage and our methodology to develop these verifiers. Due to space constraints, we only cover the high-level details here and defer further details to \Cref{sec:attribute_detector_design}, and  \Cref{sec:artifact_detector_design} in the supplementary material. 

We develop artifact and attribute detectors as verifiers to reject candidate face pairs that do not satisfy required counterfactual requirements. We observe that artifact detection is a comparatively simpler task and thus choose open-source variants.  However, in preliminary tests for attribute detection, we observe that open-source models perform poorly. Consequently, we opt to use more powerful proprietary models for the more challenging task of attribute detection to ensure higher accuracy. Each verifier used in our pipeline is validated for performance using separate user-studies~(\Cref{sec:user_survey_appendix}). Below, we discuss the high-level details of the artifact and attribute detector. 

\textbf{Artifact Detector:} We develop two types of artifact detectors. The first is a Linear SVM model trained on embeddings from a ViT-L-14 CLIP model. We utilize this detector for modifications generated via SEGA and diffusion models. To train this detector, we obtain annotations through a user study (details in \Cref{sec:user_survey_efficacy} and \Cref{sec:user_survey_appendix}) on a dataset of modified faces with and without artifacts. We calibrate the SVM to achieve a True Positive Rate of 0.97 for each attribute-demographic combination. The second artifact detector is an open-source Vision-Language Model (MOLMO-72B~\cite{deitke2024molmopixmoopenweights}) that takes in the concatenated source and modified face pair, and predicts the presence of any unnatural artifacts. We apply this detector to all GAN and StyleCLIP-based modifications. Three authors independently validate a subset of images for each attribute-demographic combination, and the detector demonstrates high agreement with the authors' annotations.

\textbf{Attribute Detectors:} We use online Vision-Language Model~(VLM)-based attribute detectors and select one of three models for each attribute-demographic combination: OpenAI GPT-4o~\cite{openai2024gpt4o}, Anthropic's Claude 3.5 Sonnet~\cite{anthropic2024claude35}, or Google's Gemini-Pro~\cite{reid2024gemini}. For all attributes except darker and lighter skin tone, we prompt the models to respond with a binary ``Yes'' or ``No'' classification regarding the attribute's presence. For the skin tone attributes, we measure relative skin tone difference between source and modified face, by making the models identify the face with the lighter skin tone. We use this response to assess if the skin tone is darker or lighter or has remained the same in the modified face.

To choose the best detector for each combination, we use a human annotated validation dataset obtained from our user surveys~(\Cref{sec:user_survey_efficacy} and \Cref{sec:user_survey_appendix}). This dataset includes five faces with the attribute and five faces without the attribute for each attribute-demographic combination. We prompt the models to identify the presence of the attribute with a Yes/No response and select the model with the highest accuracy, ensuring the overall agreement with human responses is at least 80\%. If no model achieves this, we use few-shot prompting with GPT-4o or Claude 3.5 to improve performance. Even with few-shot prompting, we observe performance of skin-tone detection on \emph{East-Asian Female} demographic is only 46.7\%, being the only combination with less than 80\% accuracy.~(further details in~\Cref{sec:attribute_detector_design})

\textbf{Ensuring Identity Consistency in candidate pairs:}
As outlined in \Cref{sec:requirements}, the faces in each candidate pair must share the same identity. We make a conscious choice of not using an FR model to verify identity preservation as such a component would go against the objective of our research, which is to enable fine-grained evaluation of FR systems. Employing an FR system in the pipeline to filter candidates inevitably biases the dataset toward samples recognizable by that specific model, regardless of its state-of-the-art status. Consequently, we rely on the intrinsic quality of the editing methods, controlled by rigorous hyperparameter calibration. This calibration, performed under author supervision, ensures high identity consistency, as corroborated by the results of our user survey~(\Cref{tab:survey--identity-consistency}).

\subsection{\cface: Dataset Details}
\label{sec:cf_dataset}
Using the above pipeline, we generate a high-quality and diverse dataset of counterfactual face pairs. \Cref{sec:rej_samplingn_pass_rates} contains the detailed performance analysis of our rejection sampling pipeline performance across the 160 attribute-demographic combinations and \Cref{tab:sample_rate_vcs} shows the pass rate, the percentage of candidate pairs passing counterfactual verifiers for each attribute-demographic combination. Our rejection sampling objective is to reach achieve 75 pairs for each combination from a maximum of 900 initial candidate pairs. We successfully achieve this target for 153 combinations. The 7 combinations that fall short are explicitly marked in \Cref{tab:sample_rate_vcs}. Therefore, our dataset is mostly balanced containing exactly 75 pairs per combination for a majority of the combinations. Attributes like glasses' and blond hair' have higher average pass rates (86.14\% and 81.31\%) than `heavy makeup' (38.30\%). These results confirm that image generators perform better for some attributes than others, possibly a reflection of their training data. Yet, with rigorous rejection sampling, we were able to extract face pairs even for challenging combinations. Overall, a total of 26733 face pairs passed the Validity and Correctness requirements, out of which 11821 also passed the Specificity requirement.

Thus, \cface consists of 11821 counterfactual face pairs, which is comparable to existing face recognition evaluation datasets like LFW~\cite{kumar2009lfwattributes}. The dataset spans 8 demographics~(2 sexes and 4 ethnicities) and 20 facial attributes. We base our demographic categories on existing FR datasets~\cite{balakrishnan2021towards, liang2023benchmarking, fournier2025fairer} and face fairness research~\cite{shen2023finetuning, yu2025data, rosenberg2023unbiased}, which use binary gender and four ethnicities~(some omit one or more of the four). The facial attributes were adopted from the popular face dataset CelebA~\cite{liu2015faceattributes} and Almudhaka et al. \cite{almudhahka2017semantic}, where facial attribute descriptions are used for face retrieval. They cover different regions of the face and can be categorized into 8 different groups: hair color, hairstyle, facial hair, makeup, skin complexion, expression, and accessories. The lists of demographics and attributes are as follows:
\begin{itemize}[leftmargin=*]
    \item \textbf{Demographics:} East Asian Male (AM), East Asian Female (AF), Indian Male (IM), Indian Female (IF), Black Male (BM), Black Female (BF), White Female (WF), White Male (WM).
    
    \item \textbf{Attributes:} Glasses, Sunglasses, Head Band, Facemask, Scarf, Smile, Lighter Skin Tone, Darker Skin Tone, Red Lipstick, Heavy Makeup, Blue Hair, Red Hair, Blond Hair, Bald, Shoulder Length Hair, Pigtails, Buzz Cut, Goatee, Mustache, Thick Beard.
\end{itemize}

Due to the sensitive nature of our dataset, we plan to keep it gated and share it only with researchers working on face recognition systems. The dataset will be available under a non-commercial license allowing only use to evaluate face recognition and related models, strictly prohibiting any use in training. The details of data and code release are available in \Cref{sec:data_release}.

\section{Post-Hoc User-Survey to Assess Dataset Quality} 
\label{sec:user_survey_efficacy}

We measure faithfulness of our pipeline in meeting counterfactual requirements using a \emph{post-hoc user survey} on a subset of \cface. The survey measures: Validity (no visible artifacts and identity is preserved), Correctness (target attribute is correctly changed), and Specificity (other attributes remain unchanged) of the counterfactual pairs. We include 10 face pairs for each attribute-demographic combination, balanced between pairs selected and rejected by our pipeline. We ask survey participants to evaluate: (i) whether the identity is preserved across the pair, and (ii) the absence or presence of a set of attributes in each image. These questions correspond to the three counterfactual requirements: Validity (identity preservation), Correctness (target attribute presence), and Specificity (non-target attribute preservation). On average, three respondents annotate each pair, following similar prior face attribute annotation studies~\cite{terhorst2021maad}. In~\Cref{tab:combined_survey}, we report results of two criteria: majority vote of respondents, and agreement from at least one respondent. 

\textbf{Identity Preservation.} \Cref{tab:survey--identity-consistency} shows that the framework preserves identity in at least 95\% of the counterfactual pairs. Remember that we enforce identity preservation by rigorous hyperparameter calibration of the editors prior to dataset generation, rather than relying on an FR model for verification. We adopt this strategy to avoid introducing bias from any other FR system. This survey attests to the efficacy of our hyperparameter tuning strategy.

\textbf{Attribute Detector.} As shown in \Cref{tab:annotator_agreement}, the attribute detectors agree with human annotators, accepting at least 62.63\% for positive samples (True Positive Rate) and rejecting 74\% of negative samples (True Negative Rate).

\textbf{Artifact Detector.} We manually inspect the dataset for distortions. Out of 1,583 face pairs, we find that the artifact detectors failed to detect distortions in only 27 pairs (False Negatives), representing an error rate of less than 2\%.

The survey highlights two key findings. First, the high TNR \textbf{confirms our hypothesis} that off-the-shelf image generators are unreliable for scalable counterfactual test set generation; human annotators reject up to 88\% of the samples that our framework also rejects. Second, our pipeline successfully generates a counterfactual dataset that satisfies the counterfactual requirements with a faithfulness rate of up to \textbf{84.08\%}.

We would like to highlight that this post-hoc survey is entirely distinct from the three other user surveys conducted in this work. We utilize the other surveys to collect testing data for validating artifact detectors and for selecting the optimal proprietary VLM for each attribute-demographic combination. In total, we conduct four user surveys, all approved by our Institutional Research Board and hosted on the Prolific platform. Participants were compensated at a rate of 15 USD per hour. We provide detailed descriptions of all four user studies in \Cref{sec:user_survey_appendix}.

\begin{table*}[ht]
    \centering
\resizebox{0.75\textwidth}{!}{
\begin{tabular}{lllllllll}
\toprule
 & AM & AF & BM & BF & WM & WF & IM & IF \\
\midrule
blond hair & 72.82 & 78.12 & 77.32 & 96.15 & 82.42 & 96.15 & 47.47 & 100.00 \\
bald & 90.36 & 44.38 & 97.40 & 60.98 & 94.94 & 37.69 & 94.94 & 64.66 \\
mustache & 60.98 & 82.42 & 50.34 & 84.27 & 62.50 & 60.48 & 22.80 & 86.21 \\
buzz cut & 50.34 & 39.68 & 30.49 & 38.27 & 45.45 & 33.04 & 64.10 & 52.45 \\
thick beard & 60.00 & 36.41 & 48.70 & 33.19 & 40.54 & 37.69 & 58.14 & 45.18 \\
red hair & 86.21 & 94.94 & 43.86 & 61.98 & 84.27 & 76.53 & 81.52 & 97.40 \\
goatee & 56.39 & 47.17 & 23.58 & 32.33 & 33.33 & 13.20 & 27.27 & 25.17 \\
pigtails & 16.74\textsuperscript{*} & 58.14 & 10.22\textsuperscript{*} & 33.19 & 17.77 & 32.75 & 17.20\textsuperscript{*} & 33.04 \\
heavy makeup & 51.02 & 54.35 & 40.76 & 32.33 & 33.33 & 22.73 & 47.17 & 23.89 \\
shoulder length hair & 30.86 & 33.78 & 36.95 & 14.73 & 50.00 & 19.18 & 63.56 & 18.25 \\
blue hair & 53.57 & 50.68 & 28.63 & 39.89 & 61.48 & 32.75 & 51.72 & 66.37 \\
head band & 83.33 & 86.21 & 70.75 & 87.21 & 89.29 & 18.84 & 5.15\textsuperscript{*} & 92.59 \\
smile & 88.24 & 76.53 & 77.32 & 56.82 & 77.32 & 52.82 & 93.75 & 89.29 \\
red lipstick & 84.27 & 92.59 & 32.33 & 91.46 & 44.64 & 90.36 & 36.59 & 94.94 \\
scarf & 57.69 & 78.95 & 55.97 & 83.33 & 62.50 & 65.22 & 55.97 & 76.53 \\
glasses & 86.90\textsuperscript{*} & 90.36 & 100.00 & 67.57 & 100.00 & 98.68 & 94.94 & 50.68 \\
sunglasses & 61.98 & 85.23 & 60.48 & 63.56 & 65.79 & 67.57 & 75.00 & 91.46 \\
facemask & 92.59 & 93.75 & 64.66 & 81.52 & 53.19 & 24.43 & 6.25\textsuperscript{*} & 100.00 \\
darker skin tone & 39.47 & 10.02\textsuperscript{*} & 64.66 & 71.43 & 38.27 & 26.41 & 53.19 & 75.76 \\
lighter skin tone & 32.47 & 27.47 & 56.82 & 67.57 & 32.89 & 22.32 & 53.57 & 57.69 \\
\bottomrule
\end{tabular}}

\caption{Pass rate (percentage of faces successfully selected by our RS-based dataset pipeline). We aim for 75 face pairs per attribute-demographic combination. Cells with an asterisk superscript indicate cases where fewer than 75 face pairs were obtained.}
\label{tab:sample_rate_vcs}
\end{table*}

\begin{table}[t]
  \centering
  \begin{subtable}[t]{0.45\columnwidth}
    \centering
    \begin{tabular}{@{}lrr@{}}
      \toprule
      Pair Category                  & \# Pairs & Percentage (\%) \\ 
      \midrule
      Majority of respondents        & 1503     & 95.0            \\
      At least one respondent        & 1534     & 96.9            \\
      \hline
      Total pairs in survey          & 1583     & 100.0           \\
      \bottomrule
    \end{tabular}
    \caption{Identity preservation results.}
    \label{tab:survey--identity-consistency}
  \end{subtable}
  \begin{subtable}[t]{0.45\columnwidth}
    \centering
    \begin{tabular}{@{}llrr@{}}
      \toprule
      Sample Type & \begin{tabular}{@{}c@{}}At least one \\ annotator\end{tabular} 
                              & \begin{tabular}{@{}c@{}}Majority \\ annotators\end{tabular}  \\ 
      \midrule
      Selected   & 84.08\% & 62.63\% \\
      Rejected   & 88.00\% & 74.00\% \\
      \bottomrule
    \end{tabular}
    \caption{Annotator agreement with attribute detectors.}
    \label{tab:annotator_agreement}
  \end{subtable}

  \caption{User survey results: (a) number of face pairs preserving facial identity, and (b) annotator agreement with verifiers}
  \vspace{-2em}
  \label{tab:combined_survey}
\end{table}

\section{Face Recognition Results} \label{sec:fr_results}
In this section, we present FR system counterfactual evaluation results using \cface. We first outline the evaluation setup, detailing the models and metrics employed. Next, we report fine-grained system performance across all attribute-demographic combinations. Finally, we present ablation studies analyzing the impact of model and dataset size, decision threshold and the importance of minimizing the confounding factors in preparing counterfactual data.

\subsection{Evaluation Setup}
\label{sec:evaluation_setup}
\textbf{Utilized Systems:} We use \cface and perform counterfactual evaluation of two commercial FR systems: AWS Rekognition~\cite{aws_rekognition_comparefaces} and Face++~\cite{faceplusplus_compare_api}, and four open-source models: AdaFace~(2022)~\cite{kim2022adaface}, MagFace~(2021)~\cite{meng2021magface}, FaceNet~(2015)~\cite{schroff2015facenet}, and ArcFace~(2019)~\cite{deng2019arcface}. We choose these representative systems to ensure variety in the model architectures and the datasets they were trained on.

Face detection is an important pre-processing step of modern FR systems, and we use a public implementation of MTCNN~\cite{sandberg_facenet} for face detection with all the open-source models. Subsequently, we extract embeddings from the detected faces and compute the similarity score between the source and modified faces using cosine similarity. For AdaFace and MagFace, we use the ResNet100 model trained on the MS1MV2~\cite{guo2016ms} dataset available from the official repositories. For FaceNet, we use a PyTorch implementation of an InceptionNet model trained on VGGFace2~\cite{timesler_facenet_pytorch}. Lastly, for ArcFace we use a ResNet34 model trained on MS1MV2 dataset available on the DeepFace face recognition library~\cite{serengil2024lightface}. Due to the black-box nature of AWS and Face++, we directly use their face comparison APIs to get a similarity score for the counterfactual pairs. Majority of the 27 unnaturally distorted faces ~(\Cref{sec:user_survey_efficacy}) and still remained in the dataset do not pass the face detection step and were excluded from the evaluation.

\textbf{Metrics and Threshold tuning:}
We use False Non-Match Rate~(FNMR) and False Match Rate~(FMR) as the metrics for evaluating our system, adhering to NIST recommendations~\cite{grother2019face} and prior literature~\cite{liang2023benchmarking}. FNMR is the proportion of wrongly rejected pairs of the same identity~(false negatives), and FMR is the proportion of wrongly matched pairs of different identities~(false positives). The face pairs in our dataset for each attribute-demographic combination belong to the same identity. Thus, we calculate the FNMR for each combination at an FMR of 0.1\%~(similar to prior work by Conti et al.~\cite{biaspaper}). Our choice of FMR of 0.1\% for main experiments follows NIST recommendation~\cite{grother2019face} reflecting a security-sensitive deployment environment, where the cost of a false acceptance is high. We also evaluate a utility-focused setting with FMR of 1\% as an ablation in \Cref{sec:ablation_3}.

We calibrate the decision threshold of each system at an FMR=0.1\% and 1\% using a separate source face dataset. We curate this dataset by generating five variations of the identities with the Realistic Vision diffusion model, and obtaining their inversions into StyleGAN2 latent space to reconstruct the GAN-based source face images. We use these synthesized diffusion and GAN faces as positive samples. For negative samples, we randomly sample 12,000 faces of different identities without any demographic restrictions. The tuned decision thresholds are $96.59$ when FMR=0.1\%~(64.98 when FMR=1\%), $87.78~(80.40)$, $0.467~(0.32)$, $0.53~(0.37)$, $0.76~(0.60)$, and $0.55~(0.34)$
for AWS, Face++, AdaFace, MagFace, FaceNet, and ArcFace, respectively. Note that commercial systems output a similarity score between 0 and 100, whereas open-source models utilize cosine similarity.

\subsection{Assessing Model Performance with \cface}
\label{sec:main_face_reg_results}

We perform fine-grained evaluation of these six systems on our dataset spanning 20 attributes and 8 demographics~(160 combinations). Each combination contains up to 75 face pairs differing in the target facial attribute. Due to space constraints, we only present a portion of the face recognition evaluation on our dataset in \Cref{tab:fr_specificity} and \Cref{tab:adaface_magface_specificity} for 12 of the 20 attributes. We request the reader to refer to \Cref{sec:additional_tabels} for the full results. We infer the following:

\smallskip \noindent \textbf{1. Performance across Models:} All six FR systems exhibit non-zero FNMR on the applied counterfactual changes with varying levels of robustness across different attributes and demographics. The two commercial systems~(AWS and Face++) and the newer open-source models~(AdaFace and MagFace) are overall more robust to the counterfactual changes than the older open-source models~(FaceNet and ArcFace). AWS and AdaFace, in particular, is most robutst to the counterfactual changes, showing an FNMR of 0\% across in excess of 100 attribute-demographic combinations.

\begin{table*}[!h]
    \centering
\resizebox{\textwidth}{!}{
\begin{tabular}{lrrrrrrrr|rrrrrrrr}
    \toprule
 & \multicolumn{8}{c}{\textbf{AWS}} & \multicolumn{8}{c}{\textbf{FaceNet}} \\
 & AM & AF & BM & BF & WM & WF & IM & IF & AM & AF & BM & BF & WM & WF & IM & IF \\
\midrule
bald & \cellcolor{green!60}0.00 & \cellcolor{lightgreen!60}2.67 & \cellcolor{green!60}0.00 & \cellcolor{green!60}0.00 & \cellcolor{green!60}0.00 & \cellcolor{green!60}0.00 & \cellcolor{green!60}0.00 & \cellcolor{lightgreen!60}4.00 & \cellcolor{red!60}53.33 & \cellcolor{red!60}56.00 & \cellcolor{yellow!60}32.00 & \cellcolor{yellow!60}41.33 & \cellcolor{red!60}73.33 & \cellcolor{red!60}78.67 & \cellcolor{red!60}58.67 & \cellcolor{red!60}76.00\\
mustache & \cellcolor{lightgreen!60}1.33 & \cellcolor{lightgreen!60}6.67 & \cellcolor{green!60}0.00 & \cellcolor{green!60}0.00 & \cellcolor{green!60}0.00 & \cellcolor{green!60}0.00 & \cellcolor{green!60}0.00 & \cellcolor{lightgreen!60}1.33 & \cellcolor{lightgreen!60}8.00 & \cellcolor{yellow!60}22.67 & \cellcolor{green!60}0.00 & \cellcolor{yellow!60}14.67 & \cellcolor{lightgreen!60}6.67 & \cellcolor{yellow!60}24.00 & \cellcolor{lightgreen!60}5.33 & \cellcolor{yellow!60}24.00 \\
buzz cut & \cellcolor{green!60}0.00 & \cellcolor{lightgreen!60}3.64 & \cellcolor{green!60}0.00 & \cellcolor{green!60}0.00 & \cellcolor{green!60}0.00 & \cellcolor{green!60}0.00 & \cellcolor{green!60}0.00 & \cellcolor{green!60}0.00 & \cellcolor{lightgreen!60}2.67 & \cellcolor{lightgreen!60}9.09 & \cellcolor{green!60}0.00 & \cellcolor{green!60}0.00 & \cellcolor{green!60}0.00 & \cellcolor{lightgreen!60}3.77 & \cellcolor{green!60}0.00 & \cellcolor{lightgreen!60}5.48\\
thick beard & \cellcolor{yellow!60}24.00 & \cellcolor{red!60}55.77 & \cellcolor{green!60}0.00 & \cellcolor{yellow!60}28.00 & \cellcolor{lightgreen!60}2.67 & \cellcolor{yellow!60}40.00 & \cellcolor{lightgreen!60}4.00 & \cellcolor{yellow!60}18.67 & \cellcolor{red!60}48.00 & \cellcolor{red!60}61.54 & \cellcolor{yellow!60}22.67 & \cellcolor{red!60}61.33 & \cellcolor{yellow!60}21.33 & \cellcolor{red!60}74.67 & \cellcolor{yellow!60}29.33 & \cellcolor{red!60}50.67\\
goatee & \cellcolor{green!60}0.00 & \cellcolor{red!60}45.33 & \cellcolor{green!60}0.00 & \cellcolor{yellow!60}10.67 & \cellcolor{green!60}0.00 & \cellcolor{yellow!60}13.33 & \cellcolor{green!60}0.00 & \cellcolor{lightgreen!60}5.33 & \cellcolor{yellow!60}13.33 & \cellcolor{red!60}49.33 & \cellcolor{lightgreen!60}2.74 & \cellcolor{yellow!60}14.67 & \cellcolor{lightgreen!60}2.67 & \cellcolor{red!60}44.00 & \cellcolor{green!60}0.00 & \cellcolor{lightgreen!60}9.33 \\
heavy mkup & \cellcolor{green!60}0.00 & \cellcolor{green!60}0.00 & \cellcolor{green!60}0.00 & \cellcolor{green!60}0.00 & \cellcolor{lightgreen!60}2.67 & \cellcolor{green!60}0.00 & \cellcolor{green!60}0.00 & \cellcolor{green!60}0.00 & \cellcolor{lightgreen!60}9.33 & \cellcolor{lightgreen!60}2.67 & \cellcolor{lightgreen!60}1.33 & \cellcolor{green!60}0.00 & \cellcolor{yellow!60}14.67 & \cellcolor{green!60}0.00 & \cellcolor{lightgreen!60}2.67 & \cellcolor{green!60}0.00 \\
shldr len hair & \cellcolor{yellow!60}28.00 & \cellcolor{yellow!60}12.00 & \cellcolor{lightgreen!60}5.41 & \cellcolor{green!60}0.00 & \cellcolor{green!60}0.00 & \cellcolor{green!60}0.00 & \cellcolor{lightgreen!60}1.75 & \cellcolor{green!60}0.00 & \cellcolor{red!60}36.00 & \cellcolor{yellow!60}12.00 & \cellcolor{yellow!60}16.00 & \cellcolor{lightgreen!60}1.33 & \cellcolor{lightgreen!60}7.94 & \cellcolor{lightgreen!60}5.33 & \cellcolor{lightgreen!60}4.84 & \cellcolor{lightgreen!60}4.05 \\
head band & \cellcolor{green!60}0.00 & \cellcolor{green!60}0.00 & \cellcolor{green!60}0.00 & \cellcolor{green!60}0.00 & \cellcolor{green!60}0.00 & \cellcolor{green!60}0.00 & \cellcolor{green!60}0.00 & \cellcolor{green!60}0.00 & \cellcolor{lightgreen!60}4.00 & \cellcolor{lightgreen!60}4.00 & \cellcolor{green!60}0.00 & \cellcolor{lightgreen!60}2.67 & \cellcolor{lightgreen!60}4.00 & \cellcolor{lightgreen!60}1.33 & \cellcolor{green!60}0.00 & \cellcolor{lightgreen!60}5.33 \\
scarf & \cellcolor{green!60}0.00 & \cellcolor{green!60}0.00 & \cellcolor{green!60}0.00 & \cellcolor{green!60}0.00 & \cellcolor{green!60}0.00 & \cellcolor{green!60}0.00 & \cellcolor{green!60}0.00 & \cellcolor{green!60}0.00 & \cellcolor{green!60}0.00 & \cellcolor{lightgreen!60}1.33 & \cellcolor{green!60}0.00 & \cellcolor{green!60}0.00 & \cellcolor{green!60}0.00 & \cellcolor{green!60}0.00 & \cellcolor{lightgreen!60}2.67 & \cellcolor{green!60}0.00\\
sunglasses & \cellcolor{yellow!60}24.00 & \cellcolor{yellow!60}29.33 & \cellcolor{green!60}0.00 & \cellcolor{lightgreen!60}4.00 & \cellcolor{lightgreen!60}2.67 & \cellcolor{lightgreen!60}2.67 & \cellcolor{lightgreen!60}2.67 & \cellcolor{lightgreen!60}2.67 & \cellcolor{red!60}76.00 & \cellcolor{red!60}70.67 & \cellcolor{yellow!60}27.03 & \cellcolor{yellow!60}41.33 & \cellcolor{red!60}62.67 & \cellcolor{red!60}73.33 & \cellcolor{yellow!60}18.67 & \cellcolor{red!60}50.67 \\
facemask & \cellcolor{yellow!60}13.33 & \cellcolor{yellow!60}18.67 & \cellcolor{green!60}0.00 & \cellcolor{lightgreen!60}8.00 & \cellcolor{lightgreen!60}2.67 & \cellcolor{yellow!60}14.67 & \cellcolor{yellow!60}15.79 & \cellcolor{yellow!60}10.67 & \cellcolor{red!60}74.67 & \cellcolor{red!60}77.33 & \cellcolor{yellow!60}30.67 & \cellcolor{red!60}77.33 & \cellcolor{red!60}62.67 & \cellcolor{red!60}70.67 & \cellcolor{red!60}57.89 & \cellcolor{red!60}80.00 \\
\bottomrule
\end{tabular}}
\caption{AWS and FaceNet performance (FNMR @ FMR=0.1\%) using our counterfactual data a subset of the attribute-demographic combinations.}

    \label{tab:fr_specificity}
\end{table*}

\begin{table*}[!h]
    \centering
\resizebox{\textwidth}{!}{
\begin{tabular}{lrrrrrrrr|rrrrrrrr}
    \toprule
 & \multicolumn{8}{c}{\textbf{AdaFace}} & \multicolumn{8}{c}{\textbf{MagFace}} \\
 & AM & AF & BM & BF & WM & WF & IM & IF & AM & AF & BM & BF & WM & WF & IM & IF \\
\midrule
bald & \cellcolor{lightgreen!60}1.33 & \cellcolor{lightgreen!60}4.00 & \cellcolor{green!60}0.00 & \cellcolor{lightgreen!60}2.67 & \cellcolor{green!60}0.00 & \cellcolor{green!60}0.00 & \cellcolor{green!60}0.00 & \cellcolor{lightgreen!60}2.67  & \cellcolor{lightgreen!60}4.00 & \cellcolor{lightgreen!60}6.67 & \cellcolor{lightgreen!60}1.33 & \cellcolor{yellow!60}10.67 & \cellcolor{lightgreen!60}1.33 & \cellcolor{lightgreen!60}6.67 & \cellcolor{lightgreen!60}2.67 & \cellcolor{yellow!60}33.33  \\
mustache & \cellcolor{lightgreen!60}2.67 & \cellcolor{yellow!60}13.70 & \cellcolor{green!60}0.00 & \cellcolor{lightgreen!60}2.67 & \cellcolor{green!60}0.00 & \cellcolor{lightgreen!60}5.33 & \cellcolor{green!60}0.00 & \cellcolor{lightgreen!60}4.05  & \cellcolor{lightgreen!60}6.67 & \cellcolor{yellow!60}32.00 & \cellcolor{green!60}0.00 & \cellcolor{lightgreen!60}2.67 & \cellcolor{lightgreen!60}1.33 & \cellcolor{lightgreen!60}5.33 & \cellcolor{green!60}0.00 & \cellcolor{lightgreen!60}6.67  \\
buzz cut & \cellcolor{green!60}0.00 & \cellcolor{lightgreen!60}9.09 & \cellcolor{green!60}0.00 & \cellcolor{green!60}0.00 & \cellcolor{green!60}0.00 & \cellcolor{lightgreen!60}1.89 & \cellcolor{green!60}0.00 & \cellcolor{green!60}0.00  & \cellcolor{green!60}0.00 & \cellcolor{lightgreen!60}9.09 & \cellcolor{green!60}0.00 & \cellcolor{green!60}0.00 & \cellcolor{green!60}0.00 & \cellcolor{lightgreen!60}3.77 & \cellcolor{green!60}0.00 & \cellcolor{green!60}0.00  \\
thick beard & \cellcolor{yellow!60}32.43 & \cellcolor{red!60}54.00 & \cellcolor{lightgreen!60}5.33 & \cellcolor{yellow!60}28.38 & \cellcolor{lightgreen!60}2.67 & \cellcolor{yellow!60}39.19 & \cellcolor{lightgreen!60}1.33 & \cellcolor{yellow!60}17.33  & \cellcolor{yellow!60}38.67 & \cellcolor{red!60}71.15 & \cellcolor{lightgreen!60}9.33 & \cellcolor{red!60}45.33 & \cellcolor{lightgreen!60}5.33 & \cellcolor{red!60}52.00 & \cellcolor{lightgreen!60}5.33 & \cellcolor{yellow!60}22.67  \\
goatee & \cellcolor{green!60}0.00 & \cellcolor{red!60}48.00 & \cellcolor{lightgreen!60}1.37 & \cellcolor{lightgreen!60}5.33 & \cellcolor{green!60}0.00 & \cellcolor{yellow!60}14.86 & \cellcolor{green!60}0.00 & \cellcolor{lightgreen!60}5.33  & \cellcolor{lightgreen!60}1.33 & \cellcolor{red!60}57.33 & \cellcolor{green!60}0.00 & \cellcolor{yellow!60}17.33 & \cellcolor{green!60}0.00 & \cellcolor{yellow!60}21.33 & \cellcolor{green!60}0.00 & \cellcolor{yellow!60}13.33  \\
heavy makeup & \cellcolor{green!60}0.00 & \cellcolor{green!60}0.00 & \cellcolor{green!60}0.00 & \cellcolor{green!60}0.00 & \cellcolor{lightgreen!60}4.00 & \cellcolor{green!60}0.00 & \cellcolor{green!60}0.00 & \cellcolor{green!60}0.00  & \cellcolor{lightgreen!60}4.00 & \cellcolor{green!60}0.00 & \cellcolor{lightgreen!60}1.33 & \cellcolor{lightgreen!60}1.33 & \cellcolor{lightgreen!60}5.33 & \cellcolor{green!60}0.00 & \cellcolor{green!60}0.00 & \cellcolor{green!60}0.00  \\
shldr len hair & \cellcolor{yellow!60}28.00 & \cellcolor{yellow!60}14.67 & \cellcolor{lightgreen!60}9.33 & \cellcolor{green!60}0.00 & \cellcolor{green!60}0.00 & \cellcolor{lightgreen!60}1.39 & \cellcolor{lightgreen!60}1.61 & \cellcolor{green!60}0.00  & \cellcolor{yellow!60}37.33 & \cellcolor{yellow!60}14.67 & \cellcolor{yellow!60}10.67 & \cellcolor{green!60}0.00 & \cellcolor{lightgreen!60}1.59 & \cellcolor{green!60}0.00 & \cellcolor{lightgreen!60}3.23 & \cellcolor{green!60}0.00  \\
head band & \cellcolor{lightgreen!60}1.33 & \cellcolor{lightgreen!60}1.33 & \cellcolor{green!60}0.00 & \cellcolor{green!60}0.00 & \cellcolor{lightgreen!60}2.67 & \cellcolor{lightgreen!60}2.67 & \cellcolor{green!60}0.00 & \cellcolor{green!60}0.00  & \cellcolor{lightgreen!60}2.67 & \cellcolor{lightgreen!60}4.00 & \cellcolor{lightgreen!60}1.33 & \cellcolor{green!60}0.00 & \cellcolor{lightgreen!60}1.33 & \cellcolor{green!60}0.00 & \cellcolor{green!60}0.00 & \cellcolor{green!60}0.00  \\
scarf & \cellcolor{green!60}0.00 & \cellcolor{lightgreen!60}1.33 & \cellcolor{green!60}0.00 & \cellcolor{green!60}0.00 & \cellcolor{green!60}0.00 & \cellcolor{green!60}0.00 & \cellcolor{green!60}0.00 & \cellcolor{green!60}0.00  & \cellcolor{green!60}0.00 & \cellcolor{green!60}0.00 & \cellcolor{green!60}0.00 & \cellcolor{green!60}0.00 & \cellcolor{green!60}0.00 & \cellcolor{green!60}0.00 & \cellcolor{green!60}0.00 & \cellcolor{green!60}0.00  \\
sunglasses & \cellcolor{yellow!60}39.19 & \cellcolor{red!60}48.00 & \cellcolor{lightgreen!60}1.35 & \cellcolor{yellow!60}15.07 & \cellcolor{lightgreen!60}9.33 & \cellcolor{yellow!60}21.33 & \cellcolor{lightgreen!60}2.67 & \cellcolor{yellow!60}16.00  & \cellcolor{red!60}46.67 & \cellcolor{red!60}52.00 & \cellcolor{lightgreen!60}2.70 & \cellcolor{yellow!60}17.33 & \cellcolor{yellow!60}13.33 & \cellcolor{yellow!60}28.00 & \cellcolor{lightgreen!60}8.00 & \cellcolor{yellow!60}25.33  \\
facemask & \cellcolor{yellow!60}32.88 & \cellcolor{red!60}52.78 & \cellcolor{lightgreen!60}4.05 & \cellcolor{yellow!60}26.03 & \cellcolor{yellow!60}10.67 & \cellcolor{yellow!60}25.71 & \cellcolor{yellow!60}21.05 & \cellcolor{yellow!60}25.33  & \cellcolor{red!60}58.67 & \cellcolor{red!60}78.67 & \cellcolor{yellow!60}12.00 & \cellcolor{red!60}65.33 & \cellcolor{yellow!60}20.00 & \cellcolor{red!60}51.35 & \cellcolor{yellow!60}31.58 & \cellcolor{red!60}56.00  \\
\bottomrule
\end{tabular}}
\caption{AdaFace and MagFace performance (FNMR @ FMR=0.1\%) using our counterfactual data a subset of the attribute-demographic combinations.}
    \label{tab:adaface_magface_specificity}
\end{table*}

We attribute the difference to the architectural improvements and bigger training corpora in the newer models. We use AdaFace and MagFace variants that are both a ResNet100 backbone, trained with MS1MV2 dataset, differing mainly in the loss functions. The increased model capacity allows better feature learning and is evident in the results, when compared to the ResNet34 backbone in ArcFace. Furthermore, MS1MV2 corpus is larger than VGGFace2~(used in the FaceNet variant) and helps the model to generalize to the counterfactual changes in our dataset. These recent improvements to face recognition models and corpora are apparent in our benchmark, consistent with results on standard benchmarks like LFW~\cite{LFWTech}, IJB~\cite{whitelam2017iarpa}, and AgeDB-30~\cite{moschoglou2017agedb}.   

\smallskip \noindent \textbf{2. Fine-grained Analysis by Attribute and Demographic:} We observe that different models exhibit distinct failure modes. A clear example is the results to \textit{sunglasses}: AWS almost always correctly matches identities for all demographics except East Asian males and females, whereas other systems fail on the application of sunglasses almost always. This pattern can be extended across the various attribute-demographic combinations in our evaluation. This highlights the unique advantage of \cface over standard benchmarks: it helps in isolating specific failure modes of an FR system, enabling a model owner to adjust their system for such scenarios in-the-wild.

Our analyses also reveal a common subset of attributes that consistently degrade performance across all models. Specifically, occluding attributes such as \textit{sunglasses}, \textit{facemask} pose significant challenges for all six FR systems, including the top-performing AdaFace and AWS. Conversely, attributes that appear on the periphery of the face have negligible effect on all the systems and combinations, except for a handful of exceptions such as \textit{shoulder length hair} on \textit{East Asian Males}. Even the older FaceNet model is fairly robust to these peripheral attributes, with a 0\% or a very small FNMR for attributes such as \textit{head band}, \textit{scarf}, \textit{hair colors}, \textit{pigtails}. The ArcFace variant is the most sensitive model, exhibiting performance degradation even on these peripheral attributes.

The  commercial and newer open-source systems are either fully or mostly robust to the application of \textit{heavy makeup}, \textit{red lipstick}, \textit{smile}, \textit{hairstyle} related attributes, and \textit{glasses}, with few exceptions such as \textit{bald} causing moderately higher FNMR for Face++ and \textit{East Asian Female}, and MagFace and \textit{Indian Female}. For AdaFace and AWS, we observe that \textit{sunglasses} and \textit{facemask} show relatively higher FNMR than other attributes, across all demographics. The two models also drop in performance when \textit{thick beard} and \textit{goatee} are applied on female faces. Although facial hair on female faces is rare, this failure mode offers insight into the explainability of the features learned by these models. It suggests that while the systems can effectively distinguish males with and without facial hair, they struggle to generalize similarly for female faces, pointing to a difference in the facial regions these models use to distinguish male and female faces.

The application of \textit{thick beard} on \textit{East Asian Males} is a unique case. All six systems exhibit higher FNMRs, with even AdaFace and AWS showing relatively higher numbers~(32.43\% and 24.00\%, respectively). We believe this disparity could be due to facial hair being less common in the \textit{East Asian Male} demographic compared to other male demographics in our dataset, therefore occurring less in the training corpora. One more unique combination is the application of \textit{darker skin tone} on \textit{White Males}. Except AWS, which shows no performance drop for any skin tone related combination, all other FR systems show a slightly higher FNMR for this combination when compared to \textit{darker skin tone} and any other demographic. Overall, the performance of all six systems on the skin tone changes is either entirely robust~(AWS) or only has a small relative drop in performance when compared to other facial attributes (we urge the reader to treat skin tone performance on \emph{East Asian Females} with caution due to limited verifier performance on this combination). 

Our dataset and evaluation also provides the ability to analyze performance for a specific ethnicity, sex, or ethnicity-sex combination. For example, observing the results of AWS in \Cref{tab:aws_specificity_fr2}, we can clearly discern that the \textit{East Asian Female} demographic performance is almost always the worst across most attributes. Similarly, from \Cref{tab:magface_specificity}, we can identify that MagFace fails to match identities for female demographics more than male demographics even for non facial hair features like \textit{sunglasses} and performs the opposite for \textit{shoulder length hair}. 

\subsection{Ablations}
\label{sec:face_rec_ablations}
We perform three ablations: 1) Role of model size and training dataset in counterfactual performance 2) Importance of minimizing confounding factors in counterfactual data, and 3) Counterfactual performance of models when FMR=1\%. 

\subsubsection{Role of FR Model Size and Training Dataset}
\label{sec:ablation_1}
To isolate the impact of model capacity and training data size, we examine two additional variants of AdaFace. The first is a ResNet-100 trained on the WebFace-4M dataset~\cite{zhu2021webface260m}, referred to as \emph{Ada-100-Web}. The second is a ResNet-50 trained on the MS1MV2 dataset, referred to as \emph{Ada-50-MS}. We treat the original model~(ResNet-100 trained on MS1MV2) as the baseline, \emph{Ada-100-MS}. Ada-100-MS differs from Ada-100-Web in the dataset and from Ada-50-MS differs in the model size. This single difference between the two model pairs helps in isolating the effects of model size or the dataset in generalizing to counterfactual changes when all the other parameters are kept the same. We follow the exact same evaluation setup as the baseline model for both variants. The partial face recognition evaluation results on \cface are in \Cref{tab:adaface_model_dataset_ablation} and the full results are available in \Cref{sec:ablation_tables_1}. 

First, we compare \emph{Ada-100-MS} and \emph{Ada-50-MS}. It is clear that the smaller model performs worse, even when other factors remain constant. This is expected, as larger models possess greater expressive power, enabling them to learn more robust features and generalize better to unseen variations. We observe this performance drop even for attributes like \textit{bald}, where \emph{Ada-100-MS} performs robustly; \emph{Ada-50-MS} shows a notable performance drop, specifically for the \textit{Indian Female} demographic. Furthermore, the effect of occluding attributes \textit{sunglasses} and \textit{facemask}~(where Ada-100-MS struggles with) on Ada-50-MS is more pronounced and the larger FNMR values across all demographics.

Next, we compare \emph{Ada-100-MS} and \emph{Ada-100-Web}. The two datasets, WebFace-4M and MS1MV2, are comparable in size~(4M vs 5.8M), are both scraped from the web, and are used by the latest face recognition models as training data. We can observe from \Cref{tab:adaface_model_dataset_ablation} and \Cref{tab:adaface_magface_specificity} that performance is not too different across most attribute-demographic combinations. Majority of the attributes like \textit{mustache}, \textit{thick beard}, \textit{mustache}, \textit{goatee} yield similar results. The minimal difference in performance appears for a handful of the attributes. For \textit{bald}, \emph{Ada-100-Web} is less robust for female demographics. For \textit{sunglasses} and \textit{facemask}, \emph{Ada-100-Web} proves relatively more robust across demographics, achieving perfect classification (0\% FNMR) for \textit{Black Male}. We believe this similarity in performance could be an artifact of the distributions of the two datasets being very similar.

\subsubsection{Importance of Minimizing Confounding Factors in the Counterfactual Dataset}
\label{sec:ablation_2}
The main challenge in the curation of a counterfactual dataset lies in minimizing the role of confounding factors to ensure that model failures can be attributed solely to the targeted attribute change. However, in practice, and as we show in \Cref{sec:framework} and \Cref{sec:attribute_detector_design}, achieving this is feasible but hard. To quantify the impact of this constraint, we analyze how FR performance changes when the specificity criterion is not considered, effectively allowing confounding factors during the modification of a source face. We generate a similar dataset to \cface using our automated pipeline, by retaining all face pairs that satisfy the validity and correctness requirements, regardless of whether they satisfy the specificity criterion. We evaluate all six FR systems on this relaxed dataset. The full results are provided in \Cref{sec:ablation_tables_2}.

We observe from the paired results for all six FR systems (e.g., \Cref{tab:aws_specificity_fr2} and \Cref{tab:aws_correctness_fr2} for AWS) that not enforcing the specificity criterion gives higher FNMR values with the relaxed dataset compared to \cface across most attribute-demographic combinations. This increase ranges from a small change for attribute groups like hairstyle and hair color to a significant change for attribute group like facial hair and occluding attributes. The large difference in results across many combinations reinforces the necessity of the specificity criterion~(minimizing confounding factors) for ensuring the source only changes along the target attributes and any other related attributes.

\begin{table*}[ht]
    \centering
    \resizebox{0.9\textwidth}{!}{
\begin{tabular}{lrrrrrrrr|rrrrrrrr}
\toprule
 & \multicolumn{8}{c}{\textbf{AdaFace-50-MS}} & \multicolumn{8}{c}{\textbf{AdaFace-100-Web4m}} \\
 & AM & AF & BM & BF & WM & WF & IM & IF & AM & AF & BM & BF & WM & WF & IM & IF \\
\midrule
bald & \cellcolor{lightgreen!60}6.67 & \cellcolor{lightgreen!60}6.67 & \cellcolor{green!60}0.00 & \cellcolor{lightgreen!60}4.00 & \cellcolor{green!60}0.00 & \cellcolor{lightgreen!60}5.33 & \cellcolor{green!60}0.00 & \cellcolor{yellow!60}22.67 & \cellcolor{lightgreen!60}2.67 & \cellcolor{lightgreen!60}6.67 & \cellcolor{green!60}0.00 & \cellcolor{yellow!60}10.67 & \cellcolor{green!60}0.00 & \cellcolor{lightgreen!60}5.33 & \cellcolor{green!60}0.00 & \cellcolor{yellow!60}13.33 \\
mustache & \cellcolor{lightgreen!60}2.67 & \cellcolor{yellow!60}16.44 & \cellcolor{green!60}0.00 & \cellcolor{lightgreen!60}2.67 & \cellcolor{green!60}0.00 & \cellcolor{lightgreen!60}6.67 & \cellcolor{green!60}0.00 & \cellcolor{lightgreen!60}5.41 & \cellcolor{lightgreen!60}2.67 & \cellcolor{lightgreen!60}9.59 & \cellcolor{green!60}0.00 & \cellcolor{green!60}0.00 & \cellcolor{green!60}0.00 & \cellcolor{lightgreen!60}2.67 & \cellcolor{green!60}0.00 & \cellcolor{lightgreen!60}4.05 \\
buzz cut & \cellcolor{green!60}0.00 & \cellcolor{lightgreen!60}9.09 & \cellcolor{green!60}0.00 & \cellcolor{green!60}0.00 & \cellcolor{green!60}0.00 & \cellcolor{lightgreen!60}1.89 & \cellcolor{green!60}0.00 & \cellcolor{green!60}0.00 & \cellcolor{green!60}0.00 & \cellcolor{lightgreen!60}7.27 & \cellcolor{green!60}0.00 & \cellcolor{green!60}0.00 & \cellcolor{green!60}0.00 & \cellcolor{lightgreen!60}1.89 & \cellcolor{green!60}0.00 & \cellcolor{green!60}0.00 \\
thick beard & \cellcolor{yellow!60}28.38 & \cellcolor{red!60}58.00 & \cellcolor{lightgreen!60}2.67 & \cellcolor{yellow!60}31.08 & \cellcolor{lightgreen!60}2.67 & \cellcolor{red!60}50.00 & \cellcolor{lightgreen!60}4.00 & \cellcolor{yellow!60}17.33 & \cellcolor{yellow!60}31.08 & \cellcolor{red!60}58.00 & \cellcolor{lightgreen!60}2.67 & \cellcolor{yellow!60}29.73 & \cellcolor{lightgreen!60}2.67 & \cellcolor{red!60}41.89 & \cellcolor{lightgreen!60}1.33 & \cellcolor{yellow!60}16.00 \\
goatee & \cellcolor{lightgreen!60}1.33 & \cellcolor{red!60}49.33 & \cellcolor{lightgreen!60}1.37 & \cellcolor{yellow!60}10.67 & \cellcolor{green!60}0.00 & \cellcolor{yellow!60}13.51 & \cellcolor{green!60}0.00 & \cellcolor{lightgreen!60}5.33 & \cellcolor{green!60}0.00 & \cellcolor{yellow!60}37.33 & \cellcolor{lightgreen!60}1.37 & \cellcolor{lightgreen!60}4.00 & \cellcolor{green!60}0.00 & \cellcolor{yellow!60}13.51 & \cellcolor{green!60}0.00 & \cellcolor{lightgreen!60}2.67 \\
heavy makeup & \cellcolor{lightgreen!60}1.33 & \cellcolor{green!60}0.00 & \cellcolor{green!60}0.00 & \cellcolor{lightgreen!60}1.33 & \cellcolor{lightgreen!60}4.00 & \cellcolor{green!60}0.00 & \cellcolor{green!60}0.00 & \cellcolor{green!60}0.00 & \cellcolor{green!60}0.00 & \cellcolor{green!60}0.00 & \cellcolor{green!60}0.00 & \cellcolor{lightgreen!60}1.33 & \cellcolor{lightgreen!60}2.67 & \cellcolor{green!60}0.00 & \cellcolor{green!60}0.00 & \cellcolor{green!60}0.00 \\
shldr len hair & \cellcolor{yellow!60}36.00 & \cellcolor{yellow!60}16.00 & \cellcolor{yellow!60}12.00 & \cellcolor{green!60}0.00 & \cellcolor{green!60}0.00 & \cellcolor{lightgreen!60}1.39 & \cellcolor{lightgreen!60}3.23 & \cellcolor{green!60}0.00 & \cellcolor{yellow!60}29.33 & \cellcolor{yellow!60}12.00 & \cellcolor{lightgreen!60}9.33 & \cellcolor{green!60}0.00 & \cellcolor{green!60}0.00 & \cellcolor{lightgreen!60}1.39 & \cellcolor{lightgreen!60}1.61 & \cellcolor{green!60}0.00 \\
head band & \cellcolor{lightgreen!60}2.67 & \cellcolor{lightgreen!60}2.67 & \cellcolor{green!60}0.00 & \cellcolor{green!60}0.00 & \cellcolor{lightgreen!60}1.33 & \cellcolor{lightgreen!60}2.67 & \cellcolor{green!60}0.00 & \cellcolor{lightgreen!60}1.33 & \cellcolor{lightgreen!60}1.33 & \cellcolor{lightgreen!60}1.33 & \cellcolor{green!60}0.00 & \cellcolor{green!60}0.00 & \cellcolor{lightgreen!60}2.67 & \cellcolor{lightgreen!60}2.67 & \cellcolor{green!60}0.00 & \cellcolor{green!60}0.00 \\
scarf & \cellcolor{green!60}0.00 & \cellcolor{lightgreen!60}1.33 & \cellcolor{green!60}0.00 & \cellcolor{green!60}0.00 & \cellcolor{green!60}0.00 & \cellcolor{green!60}0.00 & \cellcolor{green!60}0.00 & \cellcolor{green!60}0.00 & \cellcolor{green!60}0.00 & \cellcolor{lightgreen!60}1.33 & \cellcolor{green!60}0.00 & \cellcolor{green!60}0.00 & \cellcolor{green!60}0.00 & \cellcolor{green!60}0.00 & \cellcolor{green!60}0.00 & \cellcolor{green!60}0.00 \\
sunglasses & \cellcolor{red!60}50.00 & \cellcolor{red!60}52.00 & \cellcolor{lightgreen!60}5.41 & \cellcolor{yellow!60}16.44 & \cellcolor{yellow!60}13.33 & \cellcolor{yellow!60}24.00 & \cellcolor{lightgreen!60}9.33 & \cellcolor{yellow!60}18.67 & \cellcolor{yellow!60}29.73 & \cellcolor{yellow!60}40.00 & \cellcolor{green!60}0.00 & \cellcolor{yellow!60}13.70 & \cellcolor{lightgreen!60}9.33 & \cellcolor{yellow!60}14.67 & \cellcolor{lightgreen!60}4.00 & \cellcolor{lightgreen!60}6.67 \\
facemask & \cellcolor{red!60}42.47 & \cellcolor{red!60}58.33 & \cellcolor{lightgreen!60}6.76 & \cellcolor{yellow!60}35.62 & \cellcolor{yellow!60}13.33 & \cellcolor{yellow!60}31.43 & \cellcolor{yellow!60}26.32 & \cellcolor{yellow!60}29.33 & \cellcolor{yellow!60}31.51 & \cellcolor{yellow!60}38.89 & \cellcolor{lightgreen!60}4.05 & \cellcolor{yellow!60}36.99 & \cellcolor{lightgreen!60}8.00 & \cellcolor{yellow!60}25.71 & \cellcolor{yellow!60}21.05 & \cellcolor{yellow!60}28.00 \\
\bottomrule
\end{tabular}
}
\caption{\textbf{Model and Dataset Size Ablation:} Performance (FNMR @ FMR=0.1\%) using \cface for AdaFace-50-MS and AdaFace-100-Web.}
\label{tab:adaface_model_dataset_ablation}
\end{table*}

\subsubsection{Counterfactual Performance when FMR=1.0\%}\label{sec:ablation_3}

We relax the threshold to an FMR to 1\% to represent a utility-focused settings and evaluate FR performance. At this relaxed threshold, as one would expect, we observe that all six systems show improved robustness across most non-occluding attributes~(such as hair color and makeup), yielding FNMR close to zero. Even the open-source variants FaceNet and ArcFace, that deteriorate on the application of any attribute when FMR=0.1\%, show significant deterioration for less than 10 attributes when FMR=1\%. However, all systems still suffer with occluding attributes~(such as \emph{sunglasses} and \emph{facemask}) and out-of-distribution attributes-demographic combinations~(such as facial hair on female demographics and \emph{shoulder length hair} on male demographics). AdaFace and AWS show the least performance deterioration across all attribute-demographic combinations, and AdaFace is particularly robust across all but 5 combinations~(FNMR>10\%). Overall, in comparison to the more stringent FMR=0.1\%, FR systems are counterfactually more robust in a utility-focused setting. \Cref{sec:ablation_tables_3} consists of the FMR=1.0\% results.

\section{Discussion}
Our work is a proof-of-concept for automated counterfactual dataset generation and performing subsequent fine-grained analysis on FR systems. We present challenges and limitations of this work, and planned future work in this section.

\textbf{Demographic and Attribute Coverage:} Although our dataset significantly improves upon existing counterfactual face datasets~\cite{balakrishnan2021towards, liang2023benchmarking}, we acknowledge that there is further scope for extending the attribute and demographic coverage. Our current demographic choices align with established face recognition benchmarks, utilizing binary gender and four primary ethnicities. However, it does not contain non-binary gender, broader ethnic categories~(e.g., Middle Eastern, Hispanic), and the fine-grained phenotypic variations that exist within macro-groups~(e.g., distinguishing between Chinese and Vietnamese faces). Similarly, while it contains 20 facial attributes covering different regions of the face, it currently lacks low-level physiological attributes like specific nose types that are often utilized by FR systems for identification~\cite{almudhahka2017semantic}. Addition of these attributes will help in further minimization of confounding factors. Also, currently we treat each attribute modification, including skin-tone changes, as a binary “Yes/No” decision, determined by the presence of the edited attribute rather than its strength. Accurately measuring intensity with automated verifiers for attributes like facial hair and age is a challenge. We plan to address these limitations in future work.

\textbf{Verifier Bias Analyses:} Our pipeline utilizes VLMs for attribute verification. As detailed in \Cref{sec:attribute_detector_design}, our analyses reveal performance disparities and biases within these VLMs across specific combinations. There are a handful of cases where the detector performance is less than 90\%  and just one less than 80\% agreement even after few-shot prompting~(skin tone detection for East Asian Females). While our conservative filtering mechanism effectively mitigates these isolated verifier errors to ensure the final dataset's integrity~(validated by our post-hoc dataset analysis), we acknowledge it is possible that such verifier errors propagate into FR counterfactual performance analysis. Additionally, we present attribute verifier bias analyses of VLMs used in the paper in \Cref{sec:attribute_detector_design}. A comprehensive fairness audit of the VLMs, in the context of facial attributes, falls outside the scope and main goal of this paper. Understanding the biases in VLMs for facial attribute detection requires more dedicated focus involving multiple models and larger datasets.

\textbf{Image Generator Choices:} Given the computational constraints of an academic setting, we prioritized SEGA and StyleCLIP, as these two methods showed promising results to modify faces using different editing techniques for both diffusion and GAN models. We show
examples of other techniques we tried in \Cref{sec:design_choices}.

\textbf{Out-of-Distribution Attribute-Demographic Combinations:} We consciously retain rare attribute-demographic combinations~(e.g., female demographic with facial hair) to evaluate how FR systems handle out-of-distribution combinations. Although such combinations are rare, they do appear in real-world settings.

\textbf{Counterfactual Training for Face Recognition:} Another possible extension of our work is counterfactual training of FR systems. In particular, by incorporating the counterfactuals from the automated pipeline, FR systems can be made more robust towards attribute-demographic combinations that performing poorly.

\textbf{Comparison with Existing Counterfactual Datasets:} Transect~\cite{balakrishnan2021towards} and CausalFaces~\cite{liang2023benchmarking} are the two comparable datasets to ours. The former is not open-sourced to the best of our knowledge. We report \emph{ArcFace} performance on CausalFaces in \Cref{sec:comparison_with_other_datasets} and the performance on the only overlapping attribute, \emph{smile}, is different for some demographics compared to our dataset. However, an explicit comparison is very hard due to 1) totally different counterfactual pairs in the two datasets, 2) limited minimization of confounding factors in CausalFaces. Our recommendation to FR practitioners is to ensure their dataset has minimal confounding factors to perform a faithful counterfactual evaluation.

\section{Conclusion}

We present a new counterfactual face dataset, \cface, for evaluating face recognition systems in a fine-grained manner. \cface was generated using an automated pipeline developed using modern image generative models and custom machine learning verifiers as rejection samplers. Our dataset spans 8 demographics and 20 attributes~(160 combinations) and is a significant improvement on previous counterfactual face datasets~\cite{liu2023improved, balakrishnan2021towards}, enabled by the automated rejection sampling pipeline developed in this paper.  We use the dataset to present a unique fine-grained evaluation of six face recognition systems— two proprietary systems: AWS, Face++, and four open-source face recognition models: MagFace, AdaFace, ArcFace, and Facenet. Our analyses help in identifying specific failure causing attribute-demographic combinations for all six systems, especially occurring with rare attribute demographic combinations like the application of facial hair on East Asian Males, and occluding attributes like sunglasses and facemask across all demographics.

\clearpage
\section{Impact Statement}
\label{sec:impact_statement}
While our data generation pipeline is designed to multi-dimensionally characterize face recognition effectiveness, we acknowledge its potential for misuse. Our work should not be employed to misappropriate individual likenesses. We also recognize that generated faces may, in rare instances, resemble real individuals due to memorization effects in the generative model. Although this phenomenon cannot be fully quantified given the inaccessible nature of the generative model's training data, we mitigate this risk through rigorous dataset governance protocols, including controlled access, explicit usage restrictions, and removal of any identifying information from released data. Please refer to \Cref{sec:data_release} for more details on our plan to release data with strict regulation protocols. 

In implementing our data pipeline, we designed prompts for source face generation to achieve identity consistency across multiple generations while preserving demographic diversity. Specifically, we employed human names in our prompts as a strategy to generate faces that exhibit both diversity across identities and consistency within each given identity. This approach builds upon published prior work~\cite{rosenberg2023unbiased}, which demonstrates that diffusion models can implicitly capture demographic and identity-like characteristics from commonly used names. Alternative strategies, such as directly specifying demographics in prompts, fail to guarantee sufficient identity diversity. Crucially, our dataset contains no real images of any individual; names were utilized solely to enhance the consistency and diversity of generated faces.

Given that commercial applications of generative AI technology remain relatively nascent, predicting downstream implications presents significant challenges. As with any tool, our software may be employed for beneficial or harmful purposes. It is therefore imperative that our data pipeline and resulting datasets be utilized in accordance with applicable legal frameworks. Despite these legitimate concerns, we identify several constructive applications for our pipeline, including enhanced understanding and mitigation of performance limitations in commercial face recognition systems, as well as increased transparency regarding the constraints of face-based biometric system deployments.

\section{Generative AI Usage Statement}
We confirm that a large language model was only used to assist in improving grammar, checking spelling, and polishing the text written by the authors. No large language model text was directly pasted into the document.

\section{Acknowledgment}
This work was supported by National Science Foundation awards CNS-1942014 and 2247381, the NSF I-Corps award 2431502, and the Wisconsin Alumni Research Foundation~(WARF) Accelerator Program.

\clearpage

\bibliographystyle{ACM-Reference-Format}
\bibliography{main}

\clearpage
\appendix

\titlecontents{lsection}
  [2.6em]                                   
  {\addvspace{0.9em}\bfseries}              
  {\contentslabel{2.6em}}                   
  {\hspace*{-2.6em}}                        
  {\hfill\contentspage}                     
  [\addvspace{0.2em}]
\titlecontents{lsubsection}
  [5.4em]                                   
  {\addvspace{0.25em}}
  {\contentslabel{2.8em}}                   
  {}
  {\hspace{0.4em}\titlerule*[0.6pc]{.}\contentspage}  
  [\addvspace{0.1em}]

\startcontents[appendices]
\begingroup
  \setcounter{tocdepth}{2}
  \phantomsection
  \section*{\centering Appendix Contents}
  \vspace{-0.5em}
  {\color{black!45}\hrule height 0.7pt}
  \vspace{1.1em}
  \printcontents[appendices]{l}{1}{}
\endgroup
\clearpage

\section{Pipeline Implementation Design Choices}
\label{sec:design_choices}
 
\subsection{Generating source faces}
Rosenberg et al.~(\cite{rosenberg2023unbiased}) propose and demonstrate that diffusion models can interpret the demographics of human names provided in prompts and successfully retain identity and demographic consistency across various variations of the same identity. We use their approach to procure source faces for different demographics. This approach helps us to mitigate the lack of diversity in faces generated with a naive prompt like `A photo of the face of an Indian Male'. Using names in the prompt helps us get variability and generate the source faces for our candidate dataset. Specifically, we utilize 150 celebrity names obtained from GPT-4o per demographic in the prompt of the form `A photo of the face of \textlangle{Name}\textrangle' and generate multiple variations of the same identity using different random seeds. 

However, we observe that the approach fails to preserve identity across multiple variations if the seeds are randomly sampled for certain less common unisex names. To ensure better identity consistency across variations, we adopt the following strategy: for each identity, we first generate 20 variations using randomly sampled seeds and then select six with the highest average cosine similarity of Facenet embeddings~\cite{schroff2015facenet}. Despite this approach, there are a handful of cases for East Asian and Black demographics with unfamiliar and unisex names where the six variations do not belong to the same identity. We manually remove such cases to ensure that the source faces are from the same identity.

\begin{figure*}
    \centering
    \begin{subfigure}[b]{0.8\textwidth}
        \centering
        \includegraphics[width=\textwidth]{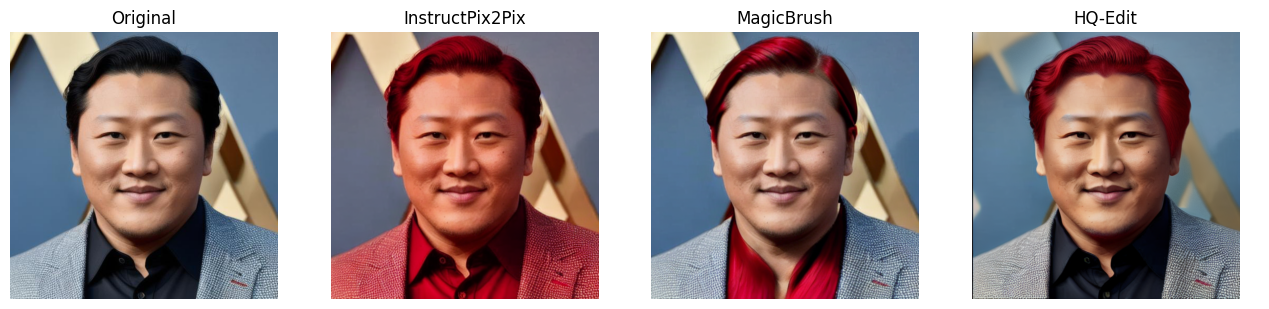}
        \caption{Edit Prompt: `Make hair color red'}
        \label{fig:sub1}
    \end{subfigure}

    \begin{subfigure}{0.8\textwidth}
        \centering
        \includegraphics[width=\textwidth]{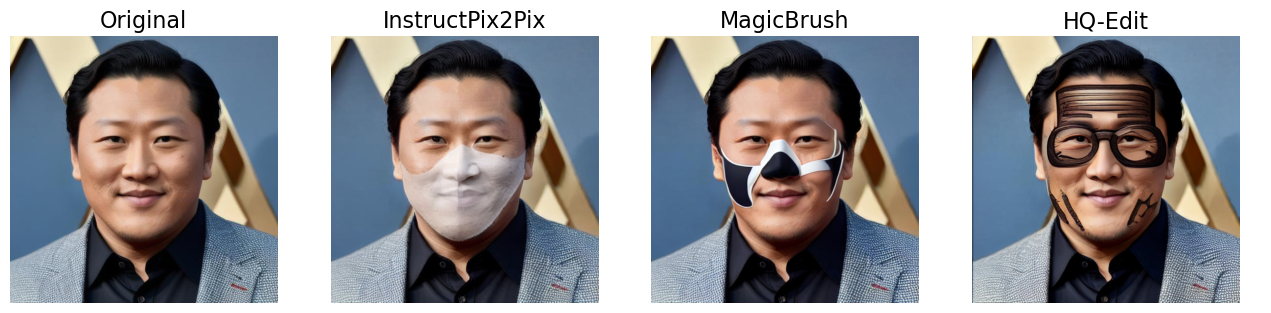}
        \caption{Edit Prompt: `Add facemask'}
        \label{fig:sub2}
    \end{subfigure}

    \begin{subfigure}{0.8\textwidth}
        \centering
        \includegraphics[width=\textwidth]{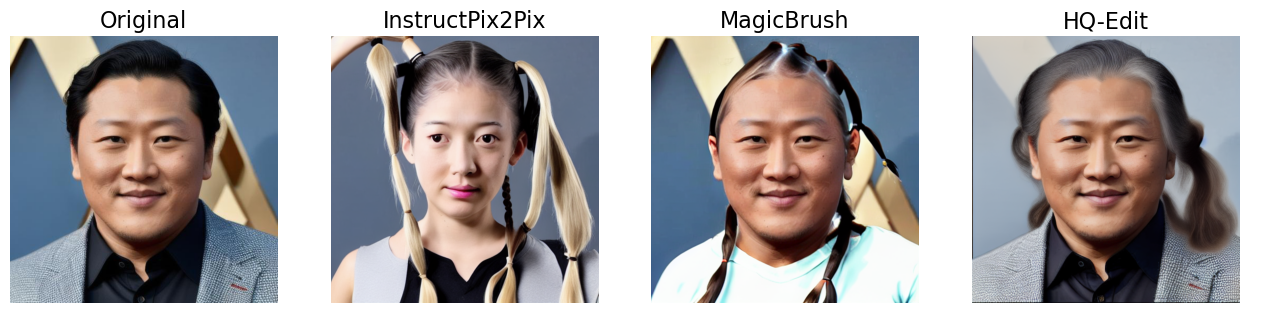}
        \caption{Edit Prompt: `Change to pigtails hairstyle'}
        \label{fig:sub3}
    \end{subfigure}

    \caption{Modification of source faces (Original) using instruction-guided editing techniques: InstructPix2Pix, MagiBrush, and HQ-Edit for different prompts. Different failures can be observed: incorrect edits, unintended edits, and identity shifts.}
    \label{fig:ip2p_fail}
\end{figure*}

\subsection{Applying modifications to source faces}
The next stage in our pipeline to generate the candidate dataset is to modify source faces by applying or removing the attribute based on whether the attribute was already present or not in the source face. Different editing approaches are prevalent in the literature using diffusion models~\cite{melzi2023gandiffface, brooks2022instructpix2pix, brack2023sega, huang2024diffusion}. Among these, we first test variants of instruction-guided editing techniques InstructPix2Pix, MagicBrush, and HQ-Edit. As the training datasets of these techniques do not cover editing images with fine details, none of these methods is able to either apply or remove all our attributes of interest, while also changing the identity often. Some examples of this can be found in \Cref{fig:ip2p_fail}.  

SEGA~\cite{brack2023sega} is a technique based on attention guidance that enables implicit editing of an image using an edit prompt, a collection of hyperparameters, the original prompt, and the random seed used to generate the original image. Our testing of SEGA shows promise to edit faces with facial attributes in our list

We tune its hyperparameters and edit prompts for modifying source faces for 16 of our 20 attributes~(refer to \Cref{tab:attr_and_editing_tech}) on a dataset containing one source face per demographic group. The tuned hyperparameters are then applied to all other source faces. The hyperparameters are also varied depending on whether attributes are being added or removed. For a small set of attributes, such as facial hair-related attributes like `thick beard' and makeup attribute `heavy makeup', we use separate hyperparameters for male and female source faces.

We observe that SEGA is not able to modify the other 4 attributes in our list. To address this, we test GAN-based editing approaches and choose StyleCLIP because it offers similar implict like SEGA but in the GAN latent space. We tune StyleCLIP's hyperparameters using the same approach as SEGA and apply it to modify all source images. Due to the ease of generating edits with StyleCLIP, we also generate StyleCLIP-based modifications for 9 of the 16 attributes we originally obtained from the diffusion model. 

For the 9 overlapping attributes, we generate modifications using both methods. To prevent duplicate pairs for the same identity, we select only one successful modification per source face. We prioritize SEGA-based modifications; however, if a SEGA modification was rejected by our artifact verifier, we utilize the StyleCLIP modification for that face instead.

\Cref{tab:attr_and_editing_tech} contains the list of attributes and the method we use to apply and remove attributes. Example images resulting from our pipeline can be found in~\reffigure{fig:transformed_faces}.

\begin{table}[h]
    \centering

    \begin{tabular}{l|l}
        \textbf{Method} & \textbf{Attributes} \\ 
        \toprule
        Diffusion + SEGA & facemask, sunglasses, mustache, \\
        & pigtails, scarf, blue hair, head band \\
        GAN + StyleCLIP & bald, darker skin tone,\\
        & lighter skin tone, blond hair \\  
        Both & buzz cut, glasses, heavy makeup,\\
        & red hair, shoulder length hair, goatee,\\
        &  red lipstick, smile, thick beard \\  
        \bottomrule
    \end{tabular}
    \caption{List of attributes and the corresponding edit method that performs best for each, based on our initial quality assessment.}
    \label{tab:attr_and_editing_tech}
\end{table}

\section{Empirical Analysis of CLIP's Limitations for Fine-Grained Facial Attributes} \label{sec:clip_failure}
We opt for VLM-based attribute verifiers instead of CLIP~\cite{radford2021learning} due to its well-documented limitations with fine-grained features~\cite{li2022grounded, zhong2022regionclip, zhang2024long}. Even~\emph{CLIP embedding-based metrics like CLIP-directional~\cite{gal2022stylegan}, commonly used to verify edit correctness, fail to capture localized semantic facial edits} as shown in prior work~\cite{rosenberg2023unbiased}. The CLIP text encoder token limit~(77 tokens) prevents using complex prompts with all attributes at once, so we instead prompt CLIP with a single attribute. We show below that even this version fails to reliably detect whether the target attribute was applied.

To evaluate CLIP as an automated attribute verifier, we treat it as a binary facial attribute classifier, similar to the verifiers used in our work. We use a balanced set of 75 faces with and 75 without the target attribute per demographic, covering four attributes: \emph{red lipstick}, \emph{red hair}, \emph{facemask}, and \emph{smile}. Although drawn from our generated dataset, these labels are reliable, as the underlying VLM verifiers were validated against human annotations and dataset is validated through a post-hoc survey. We compute cosine similarities between CLIP image and text features using targeted prompts (e.g., \textit{``face with red lipstick''}) and report the Area Under the Receiver Operating Characteristic Curve (AUC-ROC). Because AUC reflects class separation across all thresholds, it provides a clear indication of whether CLIP’s latent space can reliably distinguish the attribute.

\begin{figure*}[t]
\centering
\begin{subfigure}{0.48\textwidth}
    \centering
    \includegraphics[width=\linewidth]{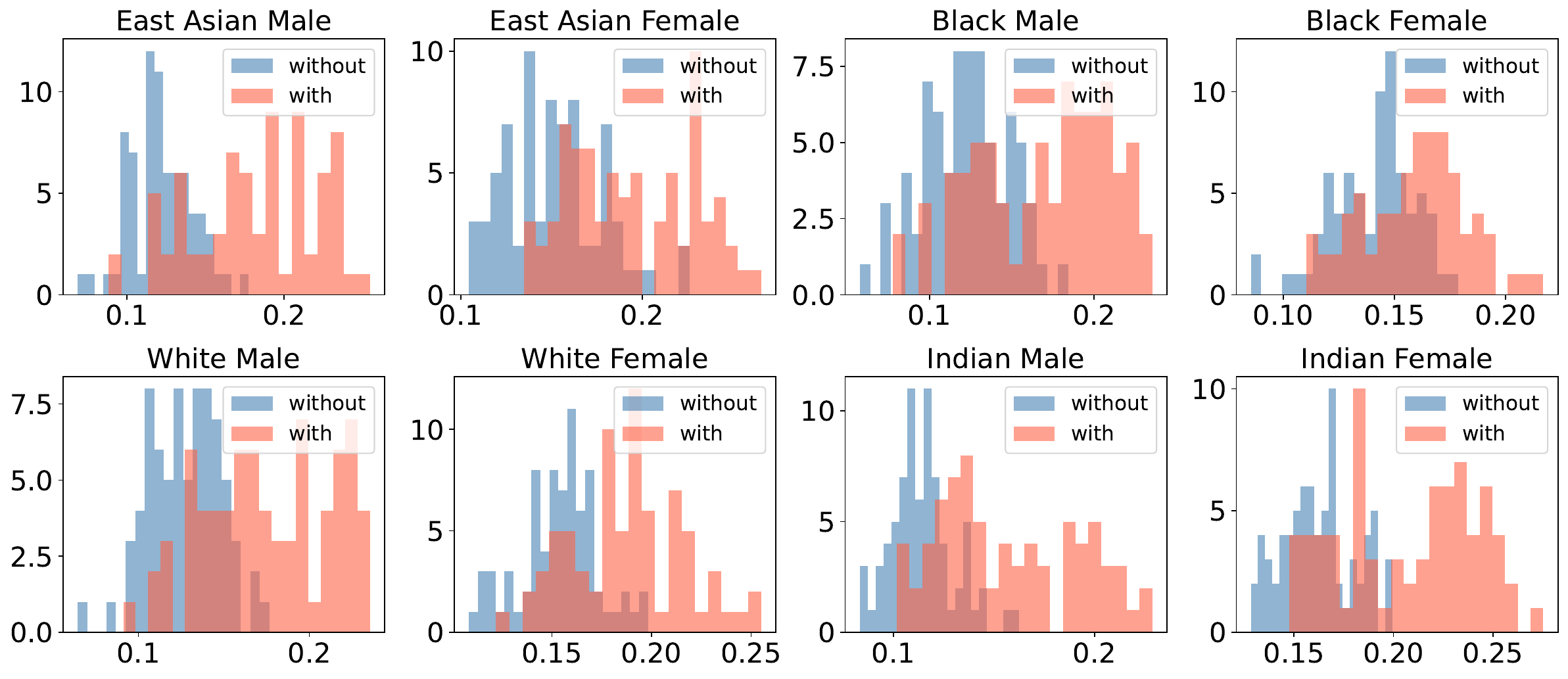}
    \caption{Red Lipstick}
    \label{fig:clip_red_lipstick}
\end{subfigure}\hfill
\begin{subfigure}{0.48\textwidth}
    \centering
    \includegraphics[width=\linewidth]{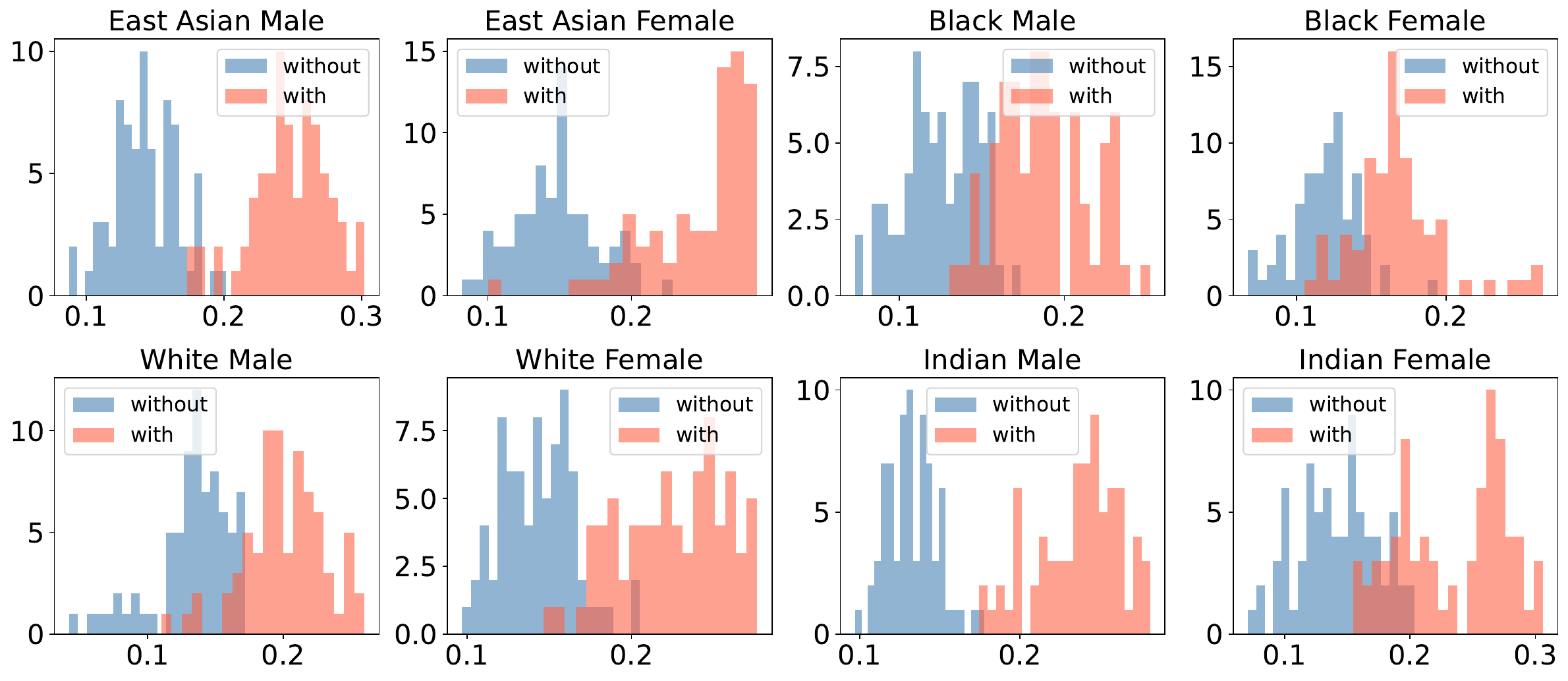}
    \caption{Red Hair}
    \label{fig:clip_red_hair}
\end{subfigure}

\vspace{0.4cm} 

\begin{subfigure}{0.48\textwidth}
    \centering
    \includegraphics[width=\linewidth]{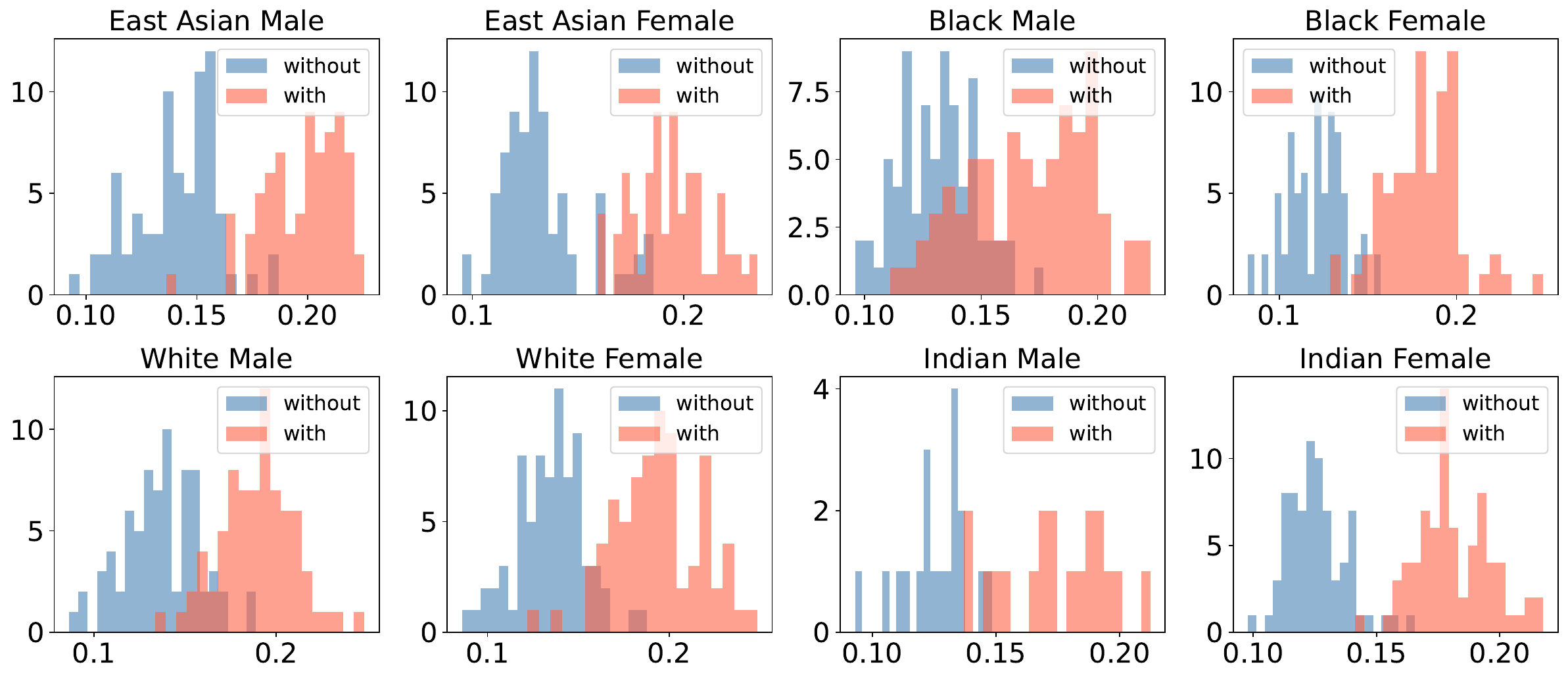}
    \caption{Facemask}
    \label{fig:clip_facemask}
\end{subfigure}\hfill
\begin{subfigure}{0.48\textwidth}
    \centering
    \includegraphics[width=\linewidth]{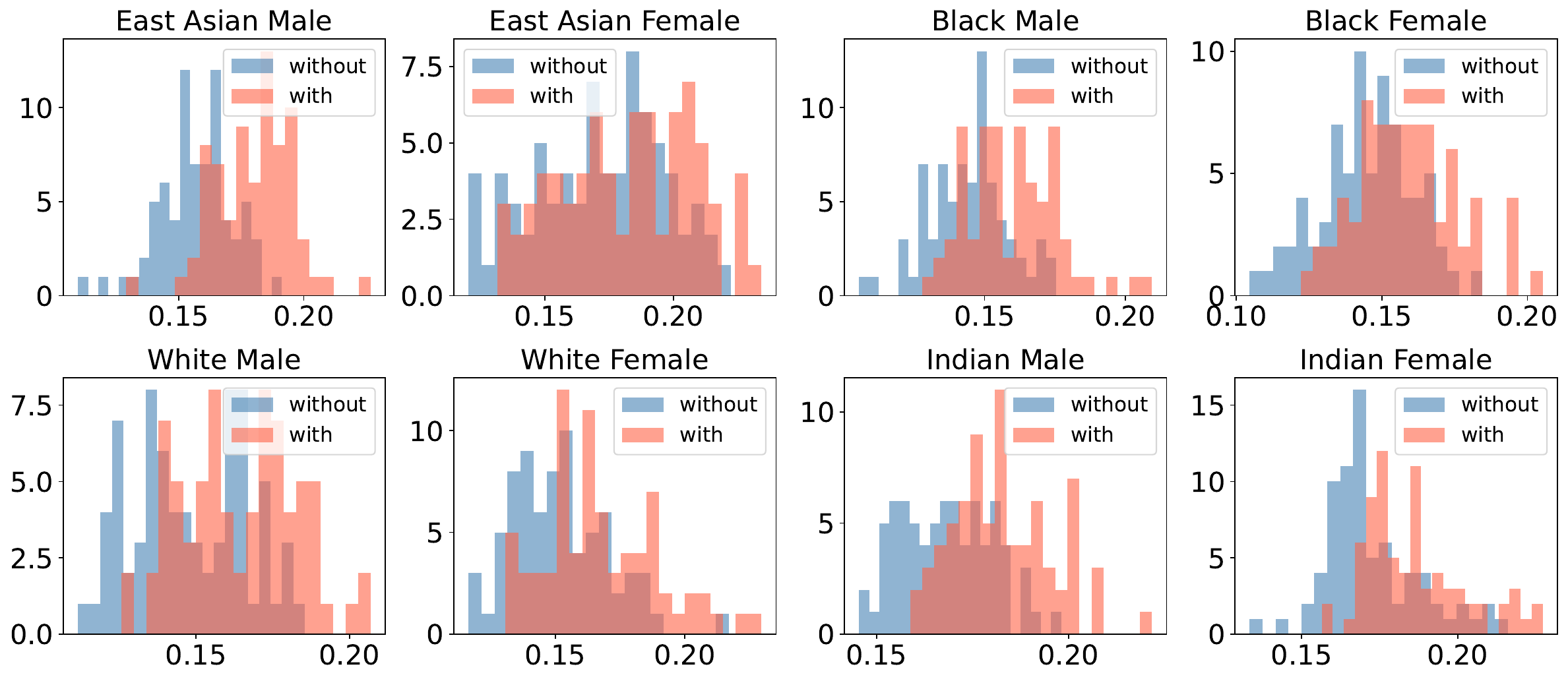}
    \caption{Smile}
    \label{fig:clip_smile}
\end{subfigure}
\caption{Distribution of vanilla CLIP cosine similarities across eight demographics for four target attributes using balanced dataset of faces with~(red) and without~(blue) the target attribute.}
\label{fig:clip_distributions_grid}
\end{figure*}

As shown in \Cref{tab:clip_auc_all} and \Cref{fig:clip_distributions_grid}, the results reveal clear limitations. CLIP works well for visually dominant attributes \emph{red hair} and \emph{facemask} (often with AUC $>0.95$ across all demographics), but struggles with the subtle and more local attributes \emph{red lipstick} and \emph{smile}. For example, detecting a \emph{smile} on \emph{East Asian Females} yields an AUC of $0.615$, which is only slightly above chance. We also see noticeable demographic disparities. The AUC for \emph{red lipstick} is $0.898$ for Asian Males but  $0.740$ for \emph{Black Females}. We also observe that for such categories the score distributions show high overlap between classes~(see \Cref{fig:clip_distributions_grid}), making it difficult to classify faces with and without the attribute. As a result, even demographic-specific thresholds are unreliable, and errors would compound when using CLIP-based verifiers for the \emph{Specificity} criterion, where confounding factors must be minimized.

A potential workaround is fine-tuning vanilla CLIP or manually tuning decision thresholds for every attribute-demographic combination. However, this is infeasible due to the lack of a large manually annotated dataset comprising all our target attributes. Constructing such a dataset is difficult without manual annotation, unlike artifact verifier training where we construct a dataset with a smart workaround~(see \Cref{sec:artifact_detector_design}). We therefore turn to proprietary VLMs that are trained on substantially larger, more diverse datasets than LAION~\cite{schuhmann2022laion}. These models yield a robust baseline, and we further optimize the pipeline by empirically selecting the best-performing VLM for each specific attribute-demographic combination. Our attribute verifiers pass stringent validation using manually annotated data from user surveys conducted in this work~(\Cref{sec:attribute_survey_appendix}) and the final dataset is also validated using a post-hoc user-survey~(\Cref{sec:dataset_validation_survey}).

\begin{table*}[t]
\centering
\resizebox{\textwidth}{!}{
\begin{tabular}{lcccc}
\toprule
\textbf{Demographic} & \textbf{Red Lipstick} & \textbf{Red Hair} & \textbf{Facemask} & \textbf{Smile} \\
\midrule
Asian Male & 0.898 [0.844, 0.945] & 0.996 [0.990, 1.000] & 0.983 [0.959, 0.999] & 0.864 [0.801, 0.915] \\
Asian Female & 0.835 [0.769, 0.894] & 0.972 [0.940, 0.994] & 0.976 [0.955, 0.992] & \textbf{0.615} [0.522, 0.700] \\
Black Male & 0.817 [0.740, 0.878] & 0.971 [0.946, 0.989] & \textbf{0.890} [0.838, 0.939] & 0.756 [0.682, 0.828] \\
Black Female & \textbf{0.740} [0.658, 0.813] & \textbf{0.921} [0.871, 0.961] & 0.988 [0.975, 0.998] & 0.710 [0.618, 0.788] \\
White Male & 0.858 [0.797, 0.914] & 0.956 [0.916, 0.986] & 0.964 [0.935, 0.987] & 0.726 [0.645, 0.799] \\
White Female & 0.834 [0.764, 0.896] & 0.980 [0.960, 0.994] & 0.970 [0.937, 0.993] & 0.711 [0.625, 0.791] \\
Indian Male & 0.880 [0.826, 0.932] & 1.000 [0.999, 1.000] & 0.983 [0.943, 1.000] & 0.792 [0.722, 0.861] \\
Indian Female & 0.851 [0.793, 0.910] & 0.961 [0.934, 0.982] & 0.997 [0.992, 1.000] & 0.762 [0.677, 0.836] \\
\bottomrule
\end{tabular}}
\caption{AUC-ROC scores and 95\% confidence intervals~(in brackets) for four facial attributes using CLIP as a facial attribute classifier. Demographic with the lowest AUC for each attribute is in bold.}
\label{tab:clip_auc_all}
\end{table*}

\section{Compute Resources} \label{sec:compute}
We ran image generation on NVIDIA RTX A6000 GPUs, using diffusion models and GANs. For GAN images, we verify attributes with MoLMO-72B, distributed across four NVIDIA RTX A6000 GPUs. For diffusion model generated images, we use the OpenAI, Anthropic, and Google Vertex APIs to check attributes with GPT-4o, Claude 3.5 Sonnet, and Gemini 1.5 Pro. For face-recognition evaluation, we use Amazon Rekognition’s Compare\_Faces function and Face++’s Compare API.We generated up to 15,000 faces in a single GPU per day using diffusion. We use 50 steps for all diffusion generations.

\section{Comparison with Existing Datasets}
\label{sec:comparison_with_other_datasets}

We present a comparison of our dataset \cface with existing real datasets~\cite{kumar2009lfwattributes, liu2015faceattributes, cao2018vggface2} and synthetic datasets~\cite{balakrishnan2021towards, liang2023benchmarking} that have a counterfactual/causal aspect in \Cref{tab:dataset_comparison}. Existing real datasets, irrespective of their target use in face recognition, evaluation~\cite{kumar2009lfwattributes} or training~\cite{liu2015faceattributes, cao2018vggface2}, have the same issue: they do not contain controlled edits to support counterfactual evaluation. Despite containing annotations for 40 different attributes, CelebA does not have controlled edits of faces. LFW is the most prominent dataset for evaluating FR systems. However, it is skewed towards white-skinned people and neither contains facial attribute annotations nor controlled edits. 

The only two datasets comparable to ours are Transect~\cite{balakrishnan2021towards} and CausalFaces~\cite{liang2023benchmarking}. Despite the obvious similarity of being synthetic and counterfactual, our dataset differs in several aspects. Firstly, both Transect and CausalFaces use only GAN-generated and modified synthetic data. Transect produces causal changes to existing faces to validate face gender classifiers across six attributes. CausalFaces, on the other hand, contains edits for four facial attributes across six demographics and validates open-source face recognition models.

As highlighted earlier, both these datasets rely on humans for face attribute annotations. Based on the cost disclosure in Transect, i.e., 0.52M annotations at the rate of 1.2 cents per annotation, the annotation amounts to around \$6000. In comparison, our automated pipeline cost us around \$650 as overall API cost to annotate in excess of 12,000 face pairs or 24,000 faces across 20 attributes. This is significantly cheaper compared to human annotation cost reported in Transect.  Our dataset contains 14 more attributes and 4 more demographics than Transect. This shows that our approach is both cost-efficient and scalable.

Lastly,  because neither Transect nor CausalFaces explicitly report the number of identities and true face pairs, we infer these numbers with based on available documentation and summarize them in \Cref{tab:dataset_comparison}. For Transect, \textit{where the data is unavailable}, we extract the counts from the description in Section 5.2 of the paper~\cite{balakrishnan2021towards}. For CausalFaces, we noticed differences in the reported numbers in the paper and the released dataset~\footnote{\url{https://rice.app.box.com/s/0t7dtfurh8jf80mhq3f7s8nbya2g58w9/folder/222955221321}}. The paper describes an initial "600 prototype images" that are later transformed for four attributes~(\emph{expression} (smile), \emph{pose}, \emph{age}, and \emph{lighting}) with differing strengths. The prototype images are also modified to obtain demographic variations~(starting from the same latent). We conservatively treat every demographic variant as a distinct identity in their dataset. The paper also mentions 48,0000 synthetic face pairs and 10,200 unique synthetic faces. However, the full dataset is not present in the released version. It contains a total of 400 identities~(100 in each demographic), and a total of 19,200 unique synthetic faces~(6000 in \emph{smile}, 5400 in \emph{age}, 3000 in \emph{lighting}, and 4800 in \emph{pose}). The reported pairs in the paper includes both pairs belonging to same and different identities within the same demographic~(we confirm this from the released identity annotation sheet). This is again different to how \textit{we characterize our dataset \cface based on the number of pairs of faces} that passed three strict counterfactual requirements. Lastly, a subset of the modified faces of CausalFaces contain attribute annotations for the target attribute in the released version. The lack of annotations for other attributes makes it hard to understand the role of confounding factors in the modified face.

\Cref{tab:arcface_iccv_causal_faces} contains the counterfactual performance of \emph{ArcFace}~(also evaluated in~\cite{liang2023benchmarking}) on the CausalFaces~\cite{liang2023benchmarking} dataset. We calibrate the threshold using our tuning dataset (see \Cref{sec:evaluation_setup}). Since CausalFaces applies attribute edits at multiple strengths, to report the overall performance, we consider the base strength version for each attribute as the source face equivalent and all other strength variants as the modified counterfactual face equivalent. The dataset spans four attributes—\emph{age}, \emph{lighting}, \emph{pose}, and \emph{smile}~(expression)—across East Asian, White, and Black demographics~(both male and female). Crucially, \emph{smile} is the only overlapping attribute with our dataset, and CausalFaces does not contain the Indian demographic.

Based on the results in \Cref{tab:arcface_iccv_causal_faces}, we observe that \emph{ArcFace} exhibits the highest robustness across most demographics for the attributes \emph{smile} and \emph{age}, whereas significant performance deterioration occurs for \emph{pose} and \emph{lighting}. Notably, the model's performance on the overlapping attribute \emph{smile} is poorest for the Black Female demographic (FNMR of 50.29\%). This contrasts with results on our dataset, where the Black Female demographic achieves much lower error rates for this attribute compared to other groups.

\textbf{We highlight} that a direct comparison between CausalFaces and \cface is difficult due to the \emph{totally different counterfactual pairs} in the two datasets and the methodologies used to generate them. CausalFaces controls for confounding factors across fewer attributes than our dataset. Thus, isolating the exact cause of these performance discrepancies is challenging. As demonstrated in \Cref{sec:ablation_2}, strictly minimizing confounding factors is essential to understand the true causal impact of an attribute edit. Therefore, we recommend face recognition practitioners the use of datasets with strict confounding factor minimization for faithful and reliable counterfactual evaluations.

\begin{table*}[ht]
    \centering
    \resizebox{0.6\textwidth}{!}{
\begin{tabular}{lrrrrrr}
\toprule
Attribute & AF & AM & BF & BM & WF & WM \\
\midrule
Age & 22.01 & 28.65 & 26.97 & 32.00 & 21.67 & 34.64 \\
Lighting & 78.85 & 64.82 & 51.76 & 72.04 & 61.22 & 54.25 \\
Pose & 73.16 & 95.03 & 93.91 & 73.88 & 77.96 & 82.13 \\
Smile & 37.66 & 30.01 & 50.29 & 43.98 & 40.61 & 23.00 \\
\bottomrule
\end{tabular}
}
\caption{\emph{ArcFace} performance (FNMR @ FMR=0.1\%) using CausalFaces.}
\label{tab:arcface_iccv_causal_faces}
\end{table*}

\section{Dataset and Code Release}
\label{sec:data_release}

Our novel face dataset consists of facial edits to synthetic faces, some of which resemble real individuals. Thus, to ensure responsible and secure use, we will keep the dataset gated and will only share it with researchers developing non-commercial face recognition or analysis systems for research purposes. Interested researchers can reach out to the corresponding author. In addition, our dataset is explicitly prohibited for training machine learning models, regardless of whether the research is commercial or non-commercial, following the recommendations in Benjamin et al.~\cite{benjamin2019towards}. The dataset will be devoid of any references to the names used during prompt generation. 

We will continuously monitor evolving legal frameworks related to AI-generated faces and update our dataset usage policy accordingly, following best practices for dataset governance as suggested in Peng et al.~\cite{peng2021mitigating}. 

Despite the limitations in fully open-sourcing the dataset, we emphasize that the dataset and the pipeline developed in this paper is novel. The fine-grained benchmarking is also a first of its kind, considering the large number of attribute-demographic combinations. The evaluation highlights discrepancies in face recognition systems along dimensions not previously studied. A future implementation of the pipeline can easily replace the generators and verifiers with then latest models, improving capabilities of the dataset.

The code and other relevant material like hyperparameters and captions used in modifying faces are available in \url{https://github.com/Guruprasad68/counterface_facct2026}.

\section{Designing the Attribute Detector}
\label{sec:attribute_detector_design}
We use a combination of three vision-language models to detect attributes in our RS pipeline. Specifically, we combine the responses from different models for various attributes and use them to verify the counterfactual requirements, correctness, and specificity. In this section, we first describe how we select one model for each attribute across different demographics. Then, we explain how we improve model alignment for a few attributes with human responses using few-shot prompting.

\subsection{Choosing the best vision model for each attribute-demographic combination}
As mentioned in \Cref{sec:framework}, we use online vision-language models to detect attributes in our candidate pairs. This is a crucial step in our RS pipeline, as it requires robust models capable of identifying various facial features. To the best of our knowledge, there is no open-source detector that can work with all attributes in our dataset. While there are multi-attribute detection models~\cite{jimaging8040105, zhuang2018multi}, they are typically limited to attributes found in open-source datasets such as CelebA~\cite{liu2015faceattributes}. 

Recent advancements in vision-language models, such as GPT-4o~\cite{openai2024gpt4o}, Claude-Sonnet3.5~\cite{anthropic2024claude35}, and Gemini~\cite{reid2024gemini}, have demonstrated capability in various image understanding tasks, ranging from optical character recognition (OCR) to aiding users in website automation. Hassanpour et al.~\cite{hassanpour2024chatgpt} and Deandreas et al.~\cite{Deandres_Tame_2024} have shown that GPT-4o can identify facial attributes in the CelebA dataset. We extend this evidence to our problem of generating counterfactual face pairs at scale. We observe that all three models, GPT-4o, Claude, and Gemini, are capable of understanding facial attributes and describing faces with low-level details, albeit with varying degrees of accuracy. 

Our pipeline requires these models to determine whether a facial attribute is present or absent in a given face. Since no existing benchmark evaluates these models for a comprehensive list of facial attributes, we curate a \textit{balanced validation dataset} and use it to select the best-performing model. For all attributes except `darker skin tone' and `lighter skin tone', the detectors respond with `Yes' or `No' to indicate the presence or absence of the attribute. For these attributes, we select a total of 10 faces, with 5 containing the attribute and 5 missing the attribute.

For skin tone difference between the faces, the detector identifies the face with a lighter skin tone from three possible responses: "right face," "left face," or "no significant difference". In this case, we curate five face pairs for each response category. We select all face images and pairs from the "Attribute Survey" described in \Cref{sec:user_survey_appendix}, and the ground truth responses using a majority vote were considered for the labels. 

Next, we input these faces to the three models to identify the presence of the 18 attributes~(excluding darker and lighter skin tone) with a `Yes' or `No', and the face with a lighter skin tone. We collect the responses from each model and compare them with the ground truth to identify the best-performing model for each attribute-demographic combination. 

\Cref{tab:det_agreement_claude}, \Cref{tab:det_agreement_gemini}, \Cref{tab:det_agreement_gpt} contain the individual performance of Claude, Gemini, and GPT-4o on our dataset for each non-skin tone attribute-demographic combinations. Using the performances of the three models, we pick the best-performing model for each attribute as the model in our RS pipeline for a given attribute and demographic. For example, the performance of GPT-4o is highest for the attribute `sunglasses' and demographic `WF~(white female)'. Thus, when we pass candidates to our rejection sampler, the response used to assess the attribute `sunglasses' for all faces belonging to `WF' is obtained from GPT-4o. In case of a tie between models, we maintain the preference order GPT4o, Claude, and then Gemini. The aggregate detector performance can be found in \Cref{tab:overall_detector_table_with_few_shot}

Our validation indicates that the three models show varying levels of alignment with human responses across attributes and demographics. All three models achieve high accuracy for attributes that are more objective in nature, such as hair color, mustache, glasses, etc. However, the lack of alignment with human responses becomes evident for attributes that are more subjective, including shoulder-length hair, heavy makeup, smile, and others. This behavior mirrors the human annotations for these attributes, as observed in our user surveys.

\Cref{tab:skin tone agreement} contains the skin tone detection performance by the three models. We observe how all three models struggle to identify skin tone change for demographics that are predominantly represented with a fairer skin tone, i.e., Asian and White ethnicities. Even though Claude performs better than the other two models across all but one demographic~(\emph{East Asian Females}), its performance is also not optimal for the fairer skin demographics. GPT-4o's prediction of skin tone change in the face pair is considerably worse than the other two models across almost all demographics.

\subsection{Using Few-Shot Prompting to Improve Performance}
The preceding results were made using a zero-shot manner to determine the presence of an attribute. As a result, some attribute-demographic combinations did not reach 80\% accuracy for any of the three models. To improve alignment with human responses for these cases, we use few-shot prompting with GPT-4o or Claude. We select two examples for each response category (Yes/No) for non-relative attributes and added them to the prompt. This allows the model to use human-selected references to refine its predictions.  

We apply few-shot prompting to attributes where the best model correctly predicted had less than 80\% accuracy, i.e., fewer than 8 correct predictions for all attributes except face with lighter skin tone prediction and fewer than 12 out of 15 predictions for the relative lighter skin tone prediction. We also consider combinations that had a false positive rate above 20\% (i.e., more than one out of five negative examples was misclassified as positive). The few-shot examples did not overlap with the validation dataset used to assess detector performance.  

\Cref{tab:overall_detector_table_with_few_shot} shows the attribute-demographic combinations where we use few-shot prompting to identify attributes during their application or removal on a source face. These examples demonstrate how few-shot prompting helps models to adjust their responses. For instance, the detector’s performance for shoulder-length hair in the Indian Female demographic improved from 60\% to 100\% accuracy. In the zero-shot setting, subjective attributes like straight hair and shoulder-length hair underperformed, but few-shot examples significantly improve their accuracy. This approach also reduces demographic disparities in performance.

Due to the increased cost of processing images in Claude and GPT-4o APIs, we use this approach selectively. We apply it only when determining the target attribute in the modified face. In all other cases, we rely on the best-performing model’s zero-shot predictions.   

We also calculate a lower and upper confidence bound for the detector performance for each attribute using KL-based Chernoff Hoeffding bounds at a 95\% confidence interval~(\Cref{tab:det bounds}), applying a union across all the attributes. The bounds once again reveal a disparity in performance between the attributes, where subjective attributes have significantly smaller lower confidence bounds. 

Despite optimizing our pipeline by aggregating the best-performing models and applying few-shot prompting, our analysis of the aggregate attribute detector~(\Cref{tab:overall_detector_table_with_few_shot}) reveals systemic demographic disparities. As mentioned earlier, the aggregate detector exhibits severe performance gaps in skin tone evaluation across all demographics; it achieves near-perfect accuracy~(14 to 15 out of 15) for Black and Indian demographics, but drops to near-chance levels for \emph{East Asian Females}~(6/15) and \emph{East Asian Males}~(8/15). Similarly, performance on subjective attributes such as \emph{shoulder length hair} is 6/10 for \emph{East Asian Females} while maintaining perfect scores for \emph{White and Indian Males}. These results highlight the demographic disparities in the VLMs used in the paper.

\begin{table*}[h]
    \centering
\centering
\resizebox{0.6\columnwidth}{!}{
\begin{tabular}{lrrrrrrrr}
\toprule
 & AM & AF & BM & BF & WM & WF & IM & IF \\
\midrule
blond hair & 9 & 10 & 9 & 10 & 10 & 10 & 10 & 10 \\
bald & 10 & 7 & 10 & 8 & 10 & 9 & 10 & 9 \\
mustache & 10 & 10 & 8 & 10 & 10 & 9 & 8 & 10 \\
buzz cut & 9 & 9 & 5 & 8 & 7 & 10 & 8 & 10 \\
thick beard & 9 & 10 & 10 & 9 & 10 & 9 & 10 & 10 \\
red hair & 10 & 9 & 10 & 10 & 8 & 9 & 10 & 10 \\
goatee & 9 & 10 & 6 & 10 & 9 & 10 & 7 & 10 \\
pigtails & 9 & 9 & 9 & 9 & 9 & 7 & 9 & 6 \\
heavy makeup & 9 & 7 & 9 & 7 & 10 & 6 & 8 & 9 \\
shoulder length hair & 10 & 7 & 10 & 4 & 8 & 6 & 9 & 5 \\
blue hair & 10 & 9 & 8 & 10 & 8 & 9 & 9 & 10 \\
head band & 10 & 10 & 10 & 10 & 9 & 10 & 9 & 10 \\
smile & 9 & 7 & 10 & 8 & 9 & 9 & 8 & 8 \\
red lipstick & 10 & 9 & 9 & 10 & 8 & 10 & 10 & 9 \\
scarf & 10 & 9 & 10 & 10 & 9 & 10 & 10 & 9 \\
glasses & 10 & 9 & 9 & 10 & 10 & 10 & 10 & 10 \\
sunglasses & 10 & 10 & 10 & 9 & 10 & 8 & 10 & 10 \\
facemask & 10 & 9 & 10 & 10 & 10 & 10 & 10 & 10 \\
\bottomrule
\end{tabular}}
\caption{Claude Sonnet 3.5 performance in detecting non-skin attributes. The number in each cell represents the model’s accuracy for the corresponding attribute-demographic combination over 10 samples , with 5 with and 5 without the attribute}
\label{tab:det_agreement_claude}
\end{table*}

\begin{table*}[h]
    \centering

\resizebox{0.6\textwidth}{!}{
\begin{tabular}{lrrrrrrrr}
\toprule
 & AM & AF & BM & BF & WM & WF & IM & IF \\
\midrule
blond hair & 8 & 10 & 6 & 9 & 10 & 10 & 8 & 10 \\
bald & 10 & 10 & 10 & 9 & 10 & 10 & 10 & 10 \\
mustache & 10 & 10 & 9 & 10 & 10 & 9 & 8 & 10 \\
buzz cut & 8 & 10 & 6 & 9 & 7 & 10 & 9 & 10 \\
thick beard & 10 & 10 & 10 & 10 & 10 & 9 & 9 & 10 \\
red hair & 10 & 10 & 10 & 10 & 8 & 8 & 10 & 10 \\
goatee & 10 & 9 & 7 & 9 & 10 & 10 & 10 & 10 \\
pigtails & 10 & 9 & 8 & 10 & 9 & 9 & 9 & 8 \\
heavy makeup & 10 & 7 & 10 & 8 & 10 & 5 & 10 & 9 \\
shoulder length hair & 9 & 7 & 9 & 7 & 8 & 5 & 9 & 5 \\
blue hair & 10 & 10 & 9 & 10 & 8 & 10 & 10 & 10 \\
head band & 10 & 10 & 10 & 10 & 9 & 10 & 10 & 10 \\
smile & 8 & 6 & 9 & 8 & 8 & 7 & 7 & 7 \\
red lipstick & 10 & 9 & 10 & 9 & 9 & 10 & 10 & 9 \\
scarf & 10 & 8 & 9 & 10 & 9 & 9 & 10 & 10 \\
glasses & 10 & 8 & 9 & 10 & 10 & 10 & 10 & 10 \\
sunglasses & 10 & 10 & 10 & 10 & 10 & 10 & 10 & 10 \\
facemask & 10 & 10 & 10 & 10 & 10 & 10 & 10 & 10 \\
\bottomrule
\end{tabular}}
\caption{Gemini 1.5 Pro performance in detecting non-skin attributes. The number in each cell represents the model’s accuracy for the corresponding attribute-demographic combination over 10 samples , with 5 with and 5 without the attribute}
\label{tab:det_agreement_gemini}
\end{table*}

\begin{table*}[h]
    \centering

\resizebox{0.6\columnwidth}{!}{
\begin{tabular}{lrrrrrrrr}
\toprule
 & AM & AF & BM & BF & WM & WF & IM & IF \\
\midrule
blond hair & 9 & 10 & 8 & 10 & 10 & 10 & 10 & 10 \\
bald & 10 & 10 & 10 & 10 & 10 & 10 & 10 & 10 \\
mustache & 9 & 10 & 8 & 10 & 10 & 9 & 9 & 10 \\
buzz cut & 10 & 10 & 8 & 9 & 8 & 10 & 10 & 10 \\
thick beard & 10 & 10 & 10 & 10 & 10 & 10 & 10 & 10 \\
red hair & 10 & 9 & 10 & 10 & 9 & 10 & 10 & 10 \\
goatee & 10 & 10 & 10 & 10 & 10 & 10 & 9 & 10 \\
pigtails & 10 & 10 & 10 & 10 & 10 & 10 & 10 & 10 \\
heavy makeup & 10 & 9 & 10 & 9 & 10 & 10 & 10 & 10 \\
shoulder length hair & 9 & 7 & 10 & 5 & 8 & 6 & 10 & 6 \\
blue hair & 10 & 10 & 10 & 10 & 10 & 10 & 10 & 10 \\
head band & 10 & 10 & 10 & 10 & 10 & 10 & 10 & 10 \\
smile & 8 & 6 & 8 & 7 & 9 & 8 & 8 & 6 \\
red lipstick & 10 & 10 & 10 & 10 & 10 & 10 & 10 & 10 \\
scarf & 10 & 10 & 10 & 10 & 10 & 9 & 9 & 10 \\
glasses & 10 & 8 & 9 & 10 & 10 & 10 & 10 & 10 \\
sunglasses & 10 & 10 & 10 & 10 & 10 & 10 & 10 & 10 \\
facemask & 10 & 10 & 10 & 10 & 10 & 10 & 10 & 10 \\
\bottomrule
\end{tabular}}
\caption{GPT-4o performance in detecting non-skin attributes. The number in each cell represents the model’s accuracy over 10 samples for the corresponding attribute-demographic combination, with 5 labeled as 'Yes' and 5 labeled as 'No'}
\label{tab:det_agreement_gpt}
\end{table*}

\begin{table*}[h]
    \centering
    \resizebox{0.8\textwidth}{!}{\begin{tabular}{lllllllll}
\toprule
 & AM & AF & BM & BF & WM & WF & IM & IF \\
\midrule
blond hair & 9\textsuperscript{{4o}} & 10\textsuperscript{{4o}} & 9\textsuperscript{{C}} & 10\textsuperscript{{4o}} & 10\textsuperscript{{4o}} & 10\textsuperscript{{4o}} & 10\textsuperscript{{4o}} & 10\textsuperscript{{4o}} \\
bald & 10\textsuperscript{{4o}} & 10\textsuperscript{{4o}} & 10\textsuperscript{{4o}} & 10\textsuperscript{{4o}} & 10\textsuperscript{{4o}} & 10\textsuperscript{{4o}} & 10\textsuperscript{{4o}} & 10\textsuperscript{{4o}} \\
mustache & 10\textsuperscript{{C}} & 10\textsuperscript{{4o}} & 9\textsuperscript{{G}} & 10\textsuperscript{{4o}} & 10\textsuperscript{{4o}} & 9\textsuperscript{{4o}} & 9\textsuperscript{{4o}} & 10\textsuperscript{{4o}} \\
buzz cut & 10\textsuperscript{{4o}} & 10\textsuperscript{{4o}} & 8\textsuperscript{{4o}} & 9\textsuperscript{{4o}} & 8\textsuperscript{{4o}} & 10\textsuperscript{{4o}} & 10\textsuperscript{{4o}} & 10\textsuperscript{{4o}} \\
thick beard & 10\textsuperscript{{4o}} & 10\textsuperscript{{4o}} & 10\textsuperscript{{4o}} & 10\textsuperscript{{4o}} & 10\textsuperscript{{4o}} & 10\textsuperscript{{4o}} & 10\textsuperscript{{4o}} & 10\textsuperscript{{4o}} \\
red hair & 10\textsuperscript{{4o}} & 10\textsuperscript{{G}} & 10\textsuperscript{{4o}} & 10\textsuperscript{{4o}} & 9\textsuperscript{{4o}} & 10\textsuperscript{{4o}} & 10\textsuperscript{{4o}} & 10\textsuperscript{{4o}} \\
goatee & 10\textsuperscript{{4o}} & 10\textsuperscript{{4o}} & 10\textsuperscript{{4o}} & 10\textsuperscript{{4o}} & 10\textsuperscript{{4o}} & 10\textsuperscript{{4o}} & 10\textsuperscript{{G}} & 10\textsuperscript{{4o}} \\
pigtails & 10\textsuperscript{{4o}} & 10\textsuperscript{{4o}} & 10\textsuperscript{{4o}} & 10\textsuperscript{{4o}} & 10\textsuperscript{{4o}} & 10\textsuperscript{{4o}} & 10\textsuperscript{{4o}} & 10\textsuperscript{{4o}} \\
heavy makeup & 10\textsuperscript{{4o}} & 9\textsuperscript{{4o}} & 10\textsuperscript{{4o}} & 9\textsuperscript{{4o}} & 10\textsuperscript{{4o}} & 10\textsuperscript{{4o}} & 10\textsuperscript{{4o}} & 10\textsuperscript{{4o}} \\
shoulder length hair & 10\textsuperscript{{C}} & 7\textsuperscript{{4o}}~(8)\textsuperscript{4o} & 10\textsuperscript{{4o}} & 7\textsuperscript{{G}}~(9)\textsuperscript{4o} & 8\textsuperscript{{4o}} & 6\textsuperscript{{4o}}~(9)\textsuperscript{4o} & 10\textsuperscript{{4o}} & 6\textsuperscript{{4o}}~(10)\textsuperscript{4o} \\
blue hair & 10\textsuperscript{{4o}} & 10\textsuperscript{{4o}} & 10\textsuperscript{{4o}} & 10\textsuperscript{{4o}} & 10\textsuperscript{{4o}} & 10\textsuperscript{{4o}} & 10\textsuperscript{{4o}} & 10\textsuperscript{{4o}} \\
head band & 10\textsuperscript{{4o}} & 10\textsuperscript{{4o}} & 10\textsuperscript{{4o}} & 10\textsuperscript{{4o}} & 10\textsuperscript{{4o}} & 10\textsuperscript{{4o}} & 10\textsuperscript{{4o}} & 10\textsuperscript{{4o}} \\
smile & 9\textsuperscript{{C}} & 7\textsuperscript{{C}}~(9)\textsuperscript{4o} & 10\textsuperscript{{C}} & 8\textsuperscript{{C}}~(10)\textsuperscript{4o} & 9\textsuperscript{{4o}} & 9\textsuperscript{{C}} & 8\textsuperscript{{4o}}~(10)\textsuperscript{4o} & 8\textsuperscript{{C}}~(10)\textsuperscript{4o} \\
red lipstick & 10\textsuperscript{{4o}} & 10\textsuperscript{{4o}} & 10\textsuperscript{{4o}} & 10\textsuperscript{{4o}} & 10\textsuperscript{{4o}} & 10\textsuperscript{{4o}} & 10\textsuperscript{{4o}} & 10\textsuperscript{{4o}} \\
scarf & 10\textsuperscript{{4o}} & 10\textsuperscript{{4o}} & 10\textsuperscript{{4o}} & 10\textsuperscript{{4o}} & 10\textsuperscript{{4o}} & 10\textsuperscript{{C}} & 10\textsuperscript{{C}} & 10\textsuperscript{{4o}} \\
glasses & 10\textsuperscript{{4o}} & 9\textsuperscript{{C}} & 9\textsuperscript{{4o}} & 10\textsuperscript{{4o}} & 10\textsuperscript{{4o}} & 10\textsuperscript{{4o}} & 10\textsuperscript{{4o}} & 10\textsuperscript{{4o}} \\
sunglasses & 10\textsuperscript{{4o}} & 10\textsuperscript{{4o}} & 10\textsuperscript{{4o}} & 10\textsuperscript{{4o}} & 10\textsuperscript{{4o}} & 10\textsuperscript{{4o}} & 10\textsuperscript{{4o}} & 10\textsuperscript{{4o}} \\
facemask & 10\textsuperscript{{4o}} & 10\textsuperscript{{4o}} & 10\textsuperscript{{4o}} & 10\textsuperscript{{4o}} & 10\textsuperscript{{4o}} & 10\textsuperscript{{4o}} & 10\textsuperscript{{4o}} & 10\textsuperscript{{4o}} \\
\bottomrule
\end{tabular}}

\caption{Aggregate Detector performance by using the best model for each attribute-demographic combination. The number in each cell represents the model’s accuracy over 10 samples for the corresponding attribute-demographic combination, with 5 labeled as 'Yes' and 5 labeled as 'No'. The superscript text over the number indicates the model used. `4o' indicates GPT-4o, `C' indicates Claude, and `G' indicates Gemini. And information within parentheses, if present indicates the few shot performance.}
\label{tab:overall_detector_table_with_few_shot}
\end{table*}

\begin{table}[h]
    \centering
\resizebox{0.3\columnwidth}{!}{
\begin{tabular}{lrrr}
\toprule
Attribute & LCB & Mean & UCB \\
\midrule
blond hair & 0.86 & 0.97 & 1.00 \\
bald & 0.92 & 1.00 & 1.00 \\
mustache & 0.83 & 0.96 & 1.00 \\
buzz cut & 0.79 & 0.94 & 0.99 \\
thick beard & 0.92 & 1.00 & 1.00 \\
red hair & 0.88 & 0.99 & 1.00 \\
goatee & 0.92 & 1.00 & 1.00 \\
pigtails & 0.92 & 1.00 & 1.00 \\
heavy makeup & 0.86 & 0.97 & 1.00 \\
shoulder length hair & 0.77 & 0.93 & 0.99 \\
blue hair & 0.92 & 1.00 & 1.00 \\
head band & 0.92 & 1.00 & 1.00 \\
smile & 0.81 & 0.95 & 1.00 \\
red lipstick & 0.92 & 1.00 & 1.00 \\
scarf & 0.92 & 1.00 & 1.00 \\
glasses & 0.86 & 0.97 & 1.00 \\
sunglasses & 0.92 & 1.00 & 1.00 \\
facemask & 0.92 & 1.00 & 1.00 \\
\bottomrule
\end{tabular}}
\caption{Lower and Upper Confidence Bounds~(LCB and UCB) of combined detector performance (including few-shot) for each attribute at a 95\% Confidence Interval, using KL-based Chernoff-Hoeffding bounds.}
\label{tab:det bounds}
\end{table}

\clearpage
\subsection{Example prompt used in the detector}
\begin{tcolorbox}[colback=blue!5!white, colframe=blue!75!black, title=Attribute Detector Prompt]
\label{box:gpt4_prompt}
\begin{center}
\includegraphics[width=0.5\textwidth]{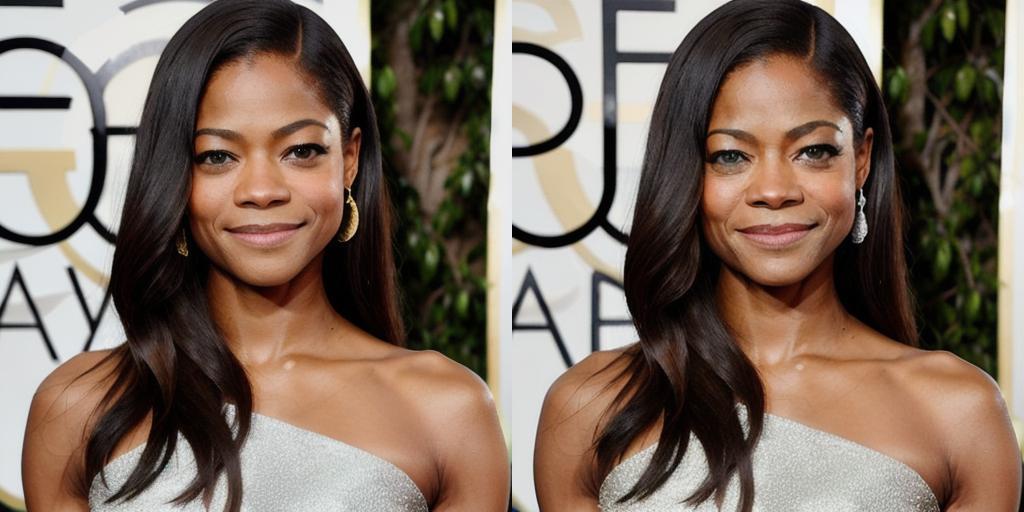}
\end{center}
Assess the two faces in the image for the following attributes:\\
red\_hair\\
mustache\\
...\\
face\_with\_lighter\_skin\_tone\\

Your output should only consist of a JSON that contains the attributes. For all attributes except 'face\_with\_lighter\_skin\_tone', the output should be a list of two Yes/No responses. The first Yes/No of the list should correspond to the left face and the second Yes/No should correspond to the right face. For face\_with\_lighter\_skin\_tone, the output should be either 'Right face', 'No significant difference', or 'Left face' 

An example output would be:\\
\{\\
    "attribute1":["Yes","No"],\\
    "attribute2":["Yes","Yes"],\\
    ...\\
    "face\_with\_lighter\_skin\_tone":"Right face"\\
\}
\end{tcolorbox}
\begin{table}[ht]
    \centering
    \begin{tabular}{lrrrrrrrr}
    \toprule
     & AM & AF & BM & BF & WM & WF & IM & IF \\
    \midrule
    Claude & 11 (12) & 6 (7) & 14 & 13 & 14 & 11 (12) & 15 & 13 \\
    Gemini & 7 & 5 & 11 & 13 & 10 & 10 & 11 & 9 \\
    Gpt4o & 9 & 7 & 13 & 11 & 9 & 9 & 8 & 8 \\
    \bottomrule
    \end{tabular}
    \caption{Skin tone performance by the three models Claude, Gemini, GPT-4o split across different demographics. Each model chooses which of the two faces in a candidate face pair has a lighter skin tone and can choose one of three options: `Right Face', `Left Face', and `No Significant Difference'. Each cell indicates the accuracy on a balanced set of 15 samples annotated by humans. The cells that contain numbers within parentheses indicate the few-shot performance of the detector.}
    \label{tab:skin tone agreement}
\end{table}

\section{Rejection Sampling Performance across different Attribute-Demographic Combinations}

\label{sec:rej_samplingn_pass_rates}
\Cref{tab:sample_rate_vcs} presents the generator's performance in producing counterfactual face pairs for each attribute-demographic combination. We report \textbf{pass rates}, defined as the percentage of generated face pairs that successfully passed our verification pipeline before reaching the target of 75 pairs per combination. For example, a pass rate of 72.82\% indicates that$(75/72.82)*100=103$ candidate pairs were passed to the pipeline to obtain 75 face pairs.

We report pass rates rather than absolute counts as they provide more interpretable performance insights. The total number of evaluated pairs occasionally deviates from the maximum of 950 per demographic due to manual removal of identities that failed to match intended demographics or maintain identity consistency~\Cref{sec:framework}. The dataset remains balanced across all combinations except those marked with asterisks in \Cref{tab:sample_rate_vcs}, where the target of 75 pairs could not be achieved.
Pass rates indicate the generator's effectiveness in modifying target attributes while preserving other facial characteristics. High pass rates suggest easier attribute modification, while low pass rates indicate greater difficulty for specific attribute-demographic combinations. For instance, generating pigtails for Asian males exhibits particularly low pass rates in \Cref{tab:sample_rate_vcs}. These disparities reflect varying generation difficulty across objective and subjective attributes, as discussed in \Cref{sec:cf_dataset}

While these performance disparities raise important questions about fairness in generative models, a more comprehensive analysis is needed for understanding this disparity in detail as the comparison needs to be performed across different models, prompt types etc. We defer such discussion to future work as the goal of this work is to just generate a dataset that facilitates face recognition evaluation in a fine-grained manner. We achieve that using our automated rejection sampling framework.

\section{Attribute Transition Matrix for Specificity Requirement}
\label{sec:candidate_filtering-transition-matrix}
\begin{figure*}[t]
    \centering
    \resizebox{\textwidth}{!}{\includegraphics{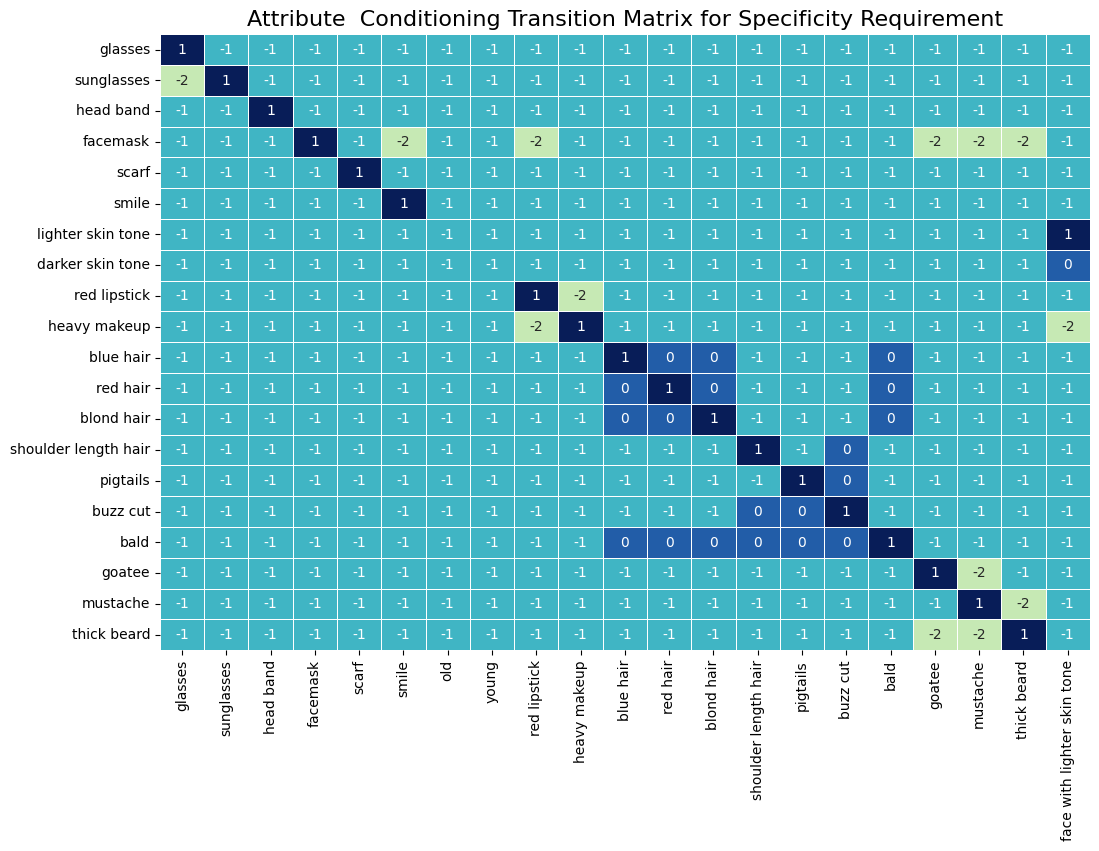}}
    \caption{Attribute Change Transition Matrix for the Candidate Filtering Step: This matrix contains our requirements for what should happen to all the attributes on a modified face obtained from a source face while applying an attribute. The columns indicate all the attributes and each row is the attribute being applied or removed. `1' indicates the attribute should be present in the modification. `0' indicates the attribute should not be present in the modification. `-1' indicates that the attribute should be present in the modification only if it is there in the source face. `-2' indicates the attribute is not considered for specificity requirement for the induced attribute.}
    \label{fig:transition-matrix-conditioning}
\end{figure*}

Our RS pipeline ensures that only the attribute applied and its `coinciding' or `contradicting' attributes change in the transformed face (\refsection{sec:framework}{Candidate Filtering}). In fact, a practitioner can loosen the requirements based on whether they want tighter or looser specificity criteria based on the attribute. For our dataset and the pipeline, we define the requirements for each attribute in a transition matrix, as shown in \reffigure{fig:transition-matrix-conditioning}. Each row in the matrix corresponds to attribute $\attribute_i$ being applied on a source face $\sample$. The vector describes the requirements of $\attributes$ in $\attributeInducement_{\attribute_i}(\sample)$, i.e., transformed face. The values in the row vector can be `1', `-1', `-2', and `0'. `1' indicates that an attribute $\attribute_j$ should be present in the transformed face. `-1' indicates that an attribute $\attribute_j$ should be present in the transformed face only if it is present in the source face. `-2' indicates that $\attribute_j$ is not considered in the filtering step. `0' indicates that $\attribute_j$ should not be present in the transformed face.

For example, when the attribute `facemask' is applied on $\sample$, the attributes `smile', `mustache', `goatee', and `red lipstick' are covered by a `facemask'. Similarly, when the attribute `sunglasses' is applied, we do not remove faces for which the `glasses' attribute also changes, as measured by detectors, because `sunglasses' are a subset of `glasses'. And, for all changes in hair color attributes, we set the other hair color attributes to zero to ensure the color of the hair changes in the modification. 

We use the column \textit{face with lighter skin tone} to indicate skin tone changes due to the application or removal of an attribute. In this column, a value of 1 means the right face in a candidate pair has a lighter tone, 0 means the left face has a lighter tone, and -1 indicates no significant difference. We apply this condition only to the attributes \textit{darker skin tone} and \textit{lighter skin tone}, as we observed that other attributes, such as \textit{heavy makeup}, can cause subtle changes in skin tone. The same degree of variation in skin tone could be perceived differently as significant or insignificant based on one's perspective.

While we receive responses from the model, we form the candidate pair by concatenating the face without the attribute and then the face with the attribute; the same version of the Attribute Transition Matrix is used irrespective of whether the attribute is applied or removed from a source face. 
 
\section{Designing Artifact Detector}
\label{sec:artifact_detector_design}
\subsection{Artifact Detector for Diffusion and SEGA-based modifications}
As described in \refsection{sec:framework}, to train the artifact detector, we curate a separate training dataset to train the artifact detector. This comprises sets of \emph{clean faces} and \emph{distorted faces}. We do not want an overlap between the source face modifications of our candidate dataset and this training set. So, we obtain a set of 10 names for each demographic. We use the source face prompt template~(refer \refsection{sec:framework} Stage1) and obtain clean faces for 25 variations of these 80 names. The seeds are randomly generated for this dataset as the clean faces are only needed for training the detector and not generating counterfactuals. We observe that as the hyperparameters of SEGA are increased to larger values, the modifications from SEGA contains artifacts that occlude the identity and\/or semantics of a face. Thus, we generate a set of distorted faces for 3 variations of the non-celebrity names for all the attributes. 

\begin{figure}
    \centering
    \includegraphics[width=\columnwidth]{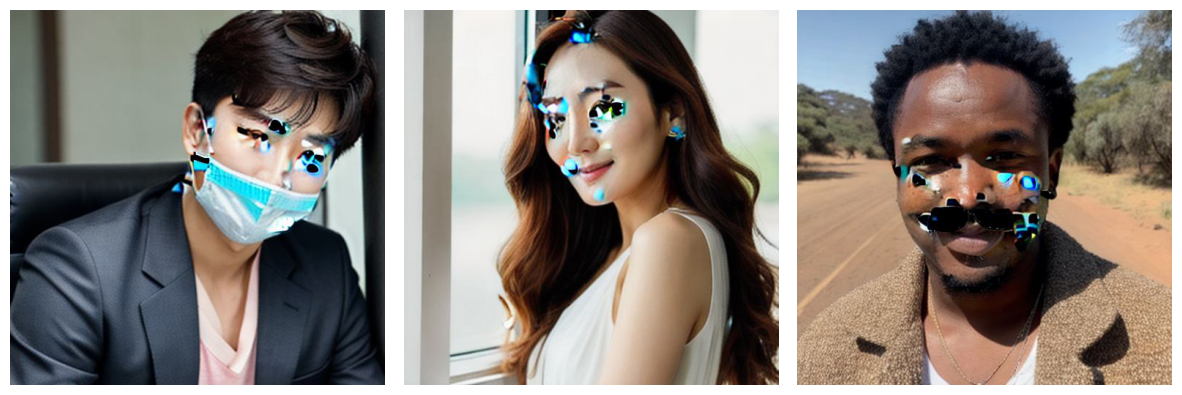}
    \caption{Subset of faces with artifacts used in training artifact detector for SEGA modification of source faces}
    \label{fig:distortion_training}
\end{figure}

\begin{figure}
    \centering
    \includegraphics[width=\columnwidth]{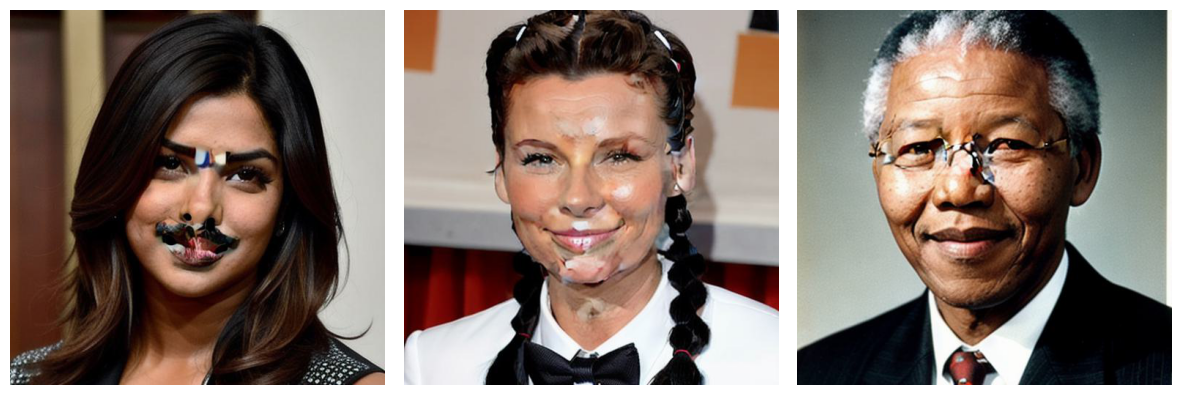}
    \caption{Example of images labeled as distorted in the Distortion Survey}
    \label{fig:distortion_survey_images}
\end{figure}

We train a Linear SVM model with the training set consisting of CLIP embeddings of these clean and distorted faces. A subset of the modified faces with artifacts used in the training set can be found in \Cref{fig:distortion_training}. As the amount of distortion in these images can be higher compared to the candidate transformed faces~(see \reffigure{fig:distortion_survey_images}), we tune the distortion detector using annotated data from a separate user-survey~(refer to \Cref{sec:distortion_survey}), to have a recall of 0.97 for detecting distortion. We use this artifact detector for all source image modifications obtained using SEGA.

\subsection{Artifact Detector for GAN and StyleCLIP-based modifications}

Unlike the modifications obtained with Diffusion and SEGA, GAN-StyleCLIP modifications produced minimal artifacts. To identify artifacts in StyleCLIP modifications, we use the open-source Vision-Language Model (MOLMO-7B). We concatenate the source and modified image pairs and prompted the model to detect any unnatural artifacts, modifications, or facial distortions. The GAN-generated images exhibit significantly less distortion compared to the diffusion-based modifications.

To validate this model, we curate a dataset consisting of five randomly selected pairs per attribute-demographic combination modified with StyleCLIP. Three authors of this paper independently reviewed these images for artifacts and observed that modifications related to the attribute bald contained the most artifacts, while those related to hair color also exhibited unnatural distortions. We finetune the prompt for the MOLMO-7B model to ensure accurate artifact detection in StyleCLIP edits. We find that providing the model with both the source and modified images helped it better understand the task.  

\section{User-Survey Details}
\label{sec:user_survey_appendix}
We conduct a total of four user surveys in this work. The first three are designed to validate and tune the verifiers of our pipeline, while the fourth is conducted to estimate the quality of our dataset \cface, curated using the automated pipeline.

The first survey, referred to as the \emph{Artifact Survey}, analyzes artifacts produced by the SEGA modification of diffusion model–generated source faces. This survey is used to tune the threshold of the distortion detector.

Next, we conduct two separate surveys to collect annotation data for the 20 attributes of interest, divided into two parts: one for diffusion model–generated source faces and modifications, and the other for GAN-generated source faces and modifications. We collectively refer to these as the Attribute Surveys: Diffusion and GAN.

Finally, we validate the quality of our dataset through a survey on face pairs selected and rejected by the pipeline. We refer to this as the \emph{Post-Hoc User Survey}. This survey provides insights into the quality of our counterfactual dataset and the effectiveness of the components in our pipeline in accepting and rejecting counterfactuals according to the counterfactual requirements.

All four surveys are designed in Qualtrics and hosted on the Prolific platform. They are approved by our Institutional Review Board. The Artifact Survey and the Attribute Surveys have a median completion time of 10 minutes, and participants are compensated 2.50 USD for their responses. The Post-Hoc User Survey has a median completion time of 15 minutes, and the participants were paid 3.75 USD. All surveys include attention checks following Prolific standards, which we use to filter out lower-quality responses.

\begin{figure*}
    \centering
    \resizebox{0.7\textwidth}{!}{
    \includegraphics[width=0.8\textwidth]{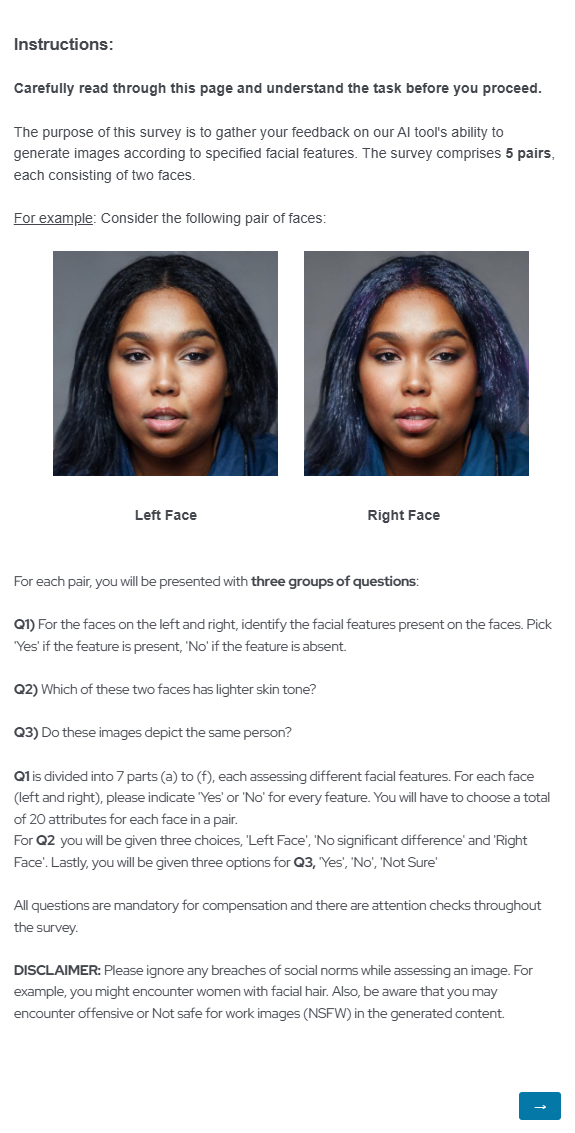}}
    \caption{Instructions for the Post-Hoc User Survey. Very similar instructions were provided in the Attribute Survey for GAN and Diffusion annotation survey.}
    \label{fig:qualtrics_instructions}
\end{figure*}

The instructions used for the attribute surveys and the Post-Hoc User Survey are in \Cref{fig:qualtrics_instructions}. 

In the following section, we provide further details on each survey.

\subsection{Artifact Survey}
\label{sec:distortion_survey}

The purpose of this survey is to assess artifacts~(distortion) in the modified source faces and use the labels to tune our artifact detector for the faces modified with SEGA. We show the participants just the modified faces. We instruct them that our AI tool generated the faces and ask them to assess each face for artifacts. The participants answer the question ``Do you think the facial features are distorted?'' and have two options (Yes \& No). The instructions provided to the participants can be found below:

\begin{tcolorbox}[colback=white!95!gray, colframe=black, title=Distortion Detector Survey Instructions,label=box:distortion_detection_survey_instructions]

Carefully read through this page and understand the task before you proceed.

The purpose of this survey is to gather your feedback on our AI tool's ability to generate a face without distorting its facial features.

The survey consists of three parts. In each part, you'll encounter different faces accompanied by the question \textit{`Do you think the facial features are distorted?'}. The flow of the survey is as follows:

\textbf{Part 1:} You will be shown 5 example faces along with a response to the question whether the faces are distorted. We also provide reasoning of why a face is \textbf{distorted} or \textbf{not-distorted}. This will help you understand what constitutes a distortion according to the context of this survey.

\textbf{Part 2:} It is a short precursor to Part 3. You'll be shown 3 faces and be asked to identify them as `Distorted' (Yes) or `Not Distorted' (No). The purpose of Part 2 is to augment and test your understanding from Part 1. Parts 1 \& 2 prepare you for Part 3.

\textbf{Part 3:} The main part of the survey presents 30 faces. For each face, we ask if it is distorted. Please examine each image carefully and respond with either `Yes' or `No'.\\

Note that the response to each question is mandatory for full compensation and there are attention questions randomly located in the survey.

\textbf{DISCLAIMER}: Please ignore any breaches of social norms while assessing an image. For example, you might encounter women with facial hair. Also, be aware that you may encounter offensive or Not safe for work images (NSFW) in the generated content.

\end{tcolorbox}

The survey consists of three stages: In the first stage, we present the participants with five examples of both distorted and non-distorted faces, along with justifications for each; in the second stage, participants evaluate three faces, where we flag them for choosing an incorrect option; in the third stage, we collect the main data, showing each participant 30 faces, including 2 attention-check questions. The first two stages help the participants in understanding the task requirements and are the same for everyone. We do not use any attention-check questions in the analysis. None of the participants fail both attention checks.

For this survey, we randomly sample 9 transformed faces per demographic for all 20 attributes, totaling 1440 faces. A total of 150 participants take part in the study, and each participant annotates 28 faces. Each face receives a maximum of 3 responses. The final label of each face is the majority vote of the annotations received from different participants of the survey. A total of 131 transformed faces out of 1368 are labeled as distorted. The lowest observed agreement between a pair of annotators annotating the same group of 28 images~(without attention checks) is 0.58, with a mean agreement of 0.8935 over all pairs of annotators.

We use these labels to tune the artifact detector's threshold for each attribute-demographic combination. As there are not enough examples with artifacts for some of the attribute-demographic combinations, we group the examples of the attributes from the same attribute-group for tuning these. The artifact detector has a minimum TPR of 0.97 for detecting artifacts for each attribute-demographic combination. In some cases, the FPR of detecting distortion is higher, but this only means the overall quality of the faces predicted as non-distorted by the detector is high.

\subsection{Attribute Surveys: Diffusion and GAN}
\label{sec:attribute_survey_appendix}
The purpose of these two surveys is to obtain annotations for a subset of faces modified using GAN-StyleCLIP and Diffusion-SEGA. A total of 700 participants annotate image pairs for a subset of attributes modified with StyleCLIP in the GAN images survey and a subset of attributes in the Diffusion images survey (\Cref{tab:attr_and_editing_tech}).

For both surveys, we sample images using weak annotations from GPT-4o to identify faces where the attribute of interest is present. This step is essential because the modification techniques do not successfully apply attributes in all instances. The goal of these surveys is to establish ground truth information for the attributes of interest and validate the selection of the best-performing detector for each attribute-demographic combination. To achieve this, we select five samples per attribute-demographic combination based on the weak labels provided by GPT-4o regarding attribute presence.

Participants in these two surveys are asked to select `Yes' or `No' via radio buttons for each face in a pair, corresponding to the attributes listed. Additionally, for the GAN survey, we ask participants a second question: which of the two faces in the given pair has a lighter skin tone? The provided options are `Right Face', `Left Face', and `No Significant Difference'.

\subsection{Post-Hoc User Survey}
\label{sec:dataset_validation_survey}
Similar to the preceding Attribute Survey, the Post-Hoc User Survey follows the same protocol: participants answer `Yes' or `No' to 18 attributes for the face pairs (in addition to `lighter skin tone', `male', and `female' annotations). We construct the dataset for this survey by randomly sampling, for every attribute–demographic combination, five images rejected and five images accepted by our RS pipeline in the strict setting. This enables us to assess the efficacy of our pipeline not only in selecting correct samples but also in rejecting incorrect ones.

The design of this survey mirrors the behavior of the attribute detectors in our pipeline—we treat participant responses as detector outputs and apply the same counterfactual criteria to determine whether a given face pair constitutes a valid counterfactual for humans. A total of 900 participants complete the survey. Each face pair also includes an identity-similarity question with the options `Yes', `Not sure', and `No', allowing us to quantify, at scale, how well the framework preserves identity.

\section{Face Recognition Performance Tables}
\label{sec:additional_tabels}

\subsection{Main Face Recognition Results (\Cref{sec:main_face_reg_results})}
\Cref{tab:aws_specificity_fr2}, \Cref{tab:facepp_specific_fr2}, \Cref{tab:adaface_ir101_msv2_specific_fr2}, \Cref{tab:magface_specificity}, \Cref{tab:facenet_specificity_fr}, and \Cref{tab:arcface_specificity_fr} comprise the full results of the face recognition performance evaluation on the 160 attribute-demographic combinations of \cface across the models AWS, Face++, AdaFace, MagFace, FaceNet, and ArcFace respectively. For easier, the tables appear one after the other with ablation results about role of confounding factors in counterfactually evaluating face recognition systems.

\subsection{Minimizing Confounding Factor Ablation Results (\Cref{sec:ablation_2})}
\label{sec:ablation_tables_2}

\Cref{tab:aws_correctness_fr2}, \Cref{tab:facepp_correctness_fr2}, \Cref{sec:adaface_correctness_v2}, \Cref{tab:magface_correctness_v2}, \Cref{tab:facenet_correctness_fr}, and \Cref{tab:arcface_correctness_fr} comprise the full results of the face recognition results, when the specificity criterion is not considered in the curation of the dataset, thereby allowing more confounding factors to appear in the modified face. 

\subsection{Model and Dataset Size Ablation Results (\Cref{sec:ablation_1})}
\label{sec:ablation_tables_1}

\Cref{tab:adaface_model_size_ablation} and \Cref{tab:adaface_dataset_size_ablation} comprise the results using \cface of the AdaFace variants: Ada-50-MS and Ada-100-Web, introduced in \Cref{sec:face_rec_ablations}.

\subsection{FMR=1.0\% Ablation Results (\Cref{sec:ablation_3})}
\label{sec:ablation_tables_3}
\Cref{tab:aws_specificity_fmr=1}, \Cref{tab:facepp_specificity_fmr=1}, \Cref{tab:adaface_ir101_msv2_specific_fmr=1}, \Cref{tab:magface_specificity_fmr=1}, \Cref{tab:facenet_specificity_fmr=1}, and \Cref{tab:arcface_specificity_fmr=1} comprise the counterfactual performance of the models at FMR=1\%.

\clearpage
\begin{table*}[]
    \centering
    \resizebox{0.65\textwidth}{!}{

\begin{tabular}{lrrrrrrrr}
    \toprule
 & AM & AF & BM & BF & WM & WF & IM & IF \\
\midrule
blond hair & \cellcolor{green!60}0.00 & \cellcolor{green!60}0.00 & \cellcolor{green!60}0.00 & \cellcolor{green!60}0.00 & \cellcolor{green!60}0.00 & \cellcolor{green!60}0.00 & \cellcolor{green!60}0.00 & \cellcolor{green!60}0.00 \\
bald & \cellcolor{green!60}0.00 & \cellcolor{lightgreen!60}2.67 & \cellcolor{green!60}0.00 & \cellcolor{green!60}0.00 & \cellcolor{green!60}0.00 & \cellcolor{green!60}0.00 & \cellcolor{green!60}0.00 & \cellcolor{lightgreen!60}4.00 \\
mustache & \cellcolor{lightgreen!60}1.33 & \cellcolor{lightgreen!60}6.67 & \cellcolor{green!60}0.00 & \cellcolor{green!60}0.00 & \cellcolor{green!60}0.00 & \cellcolor{green!60}0.00 & \cellcolor{green!60}0.00 & \cellcolor{lightgreen!60}1.33 \\
buzz cut & \cellcolor{green!60}0.00 & \cellcolor{lightgreen!60}3.64 & \cellcolor{green!60}0.00 & \cellcolor{green!60}0.00 & \cellcolor{green!60}0.00 & \cellcolor{green!60}0.00 & \cellcolor{green!60}0.00 & \cellcolor{green!60}0.00 \\
thick beard & \cellcolor{yellow!60}24.00 & \cellcolor{red!60}55.77 & \cellcolor{green!60}0.00 & \cellcolor{yellow!60}28.00 & \cellcolor{lightgreen!60}2.67 & \cellcolor{yellow!60}40.00 & \cellcolor{lightgreen!60}4.00 & \cellcolor{yellow!60}18.67 \\
red hair & \cellcolor{green!60}0.00 & \cellcolor{lightgreen!60}1.33 & \cellcolor{lightgreen!60}1.33 & \cellcolor{green!60}0.00 & \cellcolor{green!60}0.00 & \cellcolor{lightgreen!60}1.33 & \cellcolor{green!60}0.00 & \cellcolor{green!60}0.00 \\
goatee & \cellcolor{green!60}0.00 & \cellcolor{red!60}45.33 & \cellcolor{green!60}0.00 & \cellcolor{yellow!60}10.67 & \cellcolor{green!60}0.00 & \cellcolor{yellow!60}13.33 & \cellcolor{green!60}0.00 & \cellcolor{lightgreen!60}5.33 \\
\bottomrule
\end{tabular}
}
\caption{\emph{AWS} performance (FNMR @ FMR=0.1\%) using  \cface dataset}
    \label{tab:aws_specificity_fr2}
\end{table*}

\begin{table*}[h]
    \centering
    \resizebox{0.65\textwidth}{!}{
\begin{tabular}{lrrrrrrrr}
    \toprule
 & AM & AF & BM & BF & WM & WF & IM & IF \\
\midrule
blond hair & \cellcolor{green!60}0.00 & \cellcolor{green!60}0.00 & \cellcolor{green!60}0.00 & \cellcolor{green!60}0.00 & \cellcolor{green!60}0.00 & \cellcolor{green!60}0.00 & \cellcolor{green!60}0.00 & \cellcolor{green!60}0.00 \\
bald & \cellcolor{lightgreen!60}2.67 & \cellcolor{lightgreen!60}5.33 & \cellcolor{green!60}0.00 & \cellcolor{lightgreen!60}1.33 & \cellcolor{green!60}0.00 & \cellcolor{green!60}0.00 & \cellcolor{green!60}0.00 & \cellcolor{yellow!60}18.67 \\
mustache & \cellcolor{lightgreen!60}5.33 & \cellcolor{yellow!60}18.67 & \cellcolor{green!60}0.00 & \cellcolor{green!60}0.00 & \cellcolor{green!60}0.00 & \cellcolor{lightgreen!60}1.33 & \cellcolor{green!60}0.00 & \cellcolor{lightgreen!60}9.33 \\
buzz cut & \cellcolor{lightgreen!60}6.67 & \cellcolor{yellow!60}12.00 & \cellcolor{green!60}0.00 & \cellcolor{green!60}0.00 & \cellcolor{green!60}0.00 & \cellcolor{lightgreen!60}4.00 & \cellcolor{green!60}0.00 & \cellcolor{green!60}0.00 \\
thick beard & \cellcolor{red!60}66.67 & \cellcolor{red!60}97.33 & \cellcolor{yellow!60}12.00 & \cellcolor{red!60}92.00 & \cellcolor{yellow!60}28.00 & \cellcolor{red!60}82.43 & \cellcolor{lightgreen!60}9.33 & \cellcolor{red!60}72.00 \\
red hair & \cellcolor{green!60}0.00 & \cellcolor{lightgreen!60}4.00 & \cellcolor{lightgreen!60}1.33 & \cellcolor{green!60}0.00 & \cellcolor{lightgreen!60}2.67 & \cellcolor{lightgreen!60}2.67 & \cellcolor{green!60}0.00 & \cellcolor{green!60}0.00 \\
goatee & \cellcolor{lightgreen!60}2.67 & \cellcolor{red!60}57.33 & \cellcolor{green!60}0.00 & \cellcolor{yellow!60}22.67 & \cellcolor{green!60}0.00 & \cellcolor{yellow!60}20.00 & \cellcolor{green!60}0.00 & \cellcolor{yellow!60}18.67 \\
\bottomrule
\end{tabular}
}
\caption{Confounding Factors Ablation: AWS performance (FNMR@ FPR=0.1\%) using the dataset without specificity requirement.}
    \label{tab:aws_correctness_fr2}
\end{table*}

\begin{table*}[]
    \centering
    \resizebox{0.65\textwidth}{!}{
\begin{tabular}{lrrrrrrrr}
    \toprule
 & AM & AF & BM & BF & WM & WF & IM & IF \\
\midrule
blond hair & \cellcolor{green!60}0.00 & \cellcolor{green!60}0.00 & \cellcolor{green!60}0.00 & \cellcolor{green!60}0.00 & \cellcolor{green!60}0.00 & \cellcolor{green!60}0.00 & \cellcolor{green!60}0.00 & \cellcolor{green!60}0.00 \\
bald & \cellcolor{lightgreen!60}5.33 & \cellcolor{yellow!60}18.67 & \cellcolor{green!60}0.00 & \cellcolor{lightgreen!60}2.67 & \cellcolor{green!60}0.00 & \cellcolor{green!60}0.00 & \cellcolor{green!60}0.00 & \cellcolor{green!60}0.00 \\
mustache & \cellcolor{yellow!60}13.33 & \cellcolor{red!60}76.00 & \cellcolor{green!60}0.00 & \cellcolor{lightgreen!60}9.33 & \cellcolor{lightgreen!60}1.33 & \cellcolor{yellow!60}17.33 & \cellcolor{lightgreen!60}1.33 & \cellcolor{yellow!60}26.67 \\
buzz cut & \cellcolor{lightgreen!60}1.33 & \cellcolor{yellow!60}10.91 & \cellcolor{green!60}0.00 & \cellcolor{green!60}0.00 & \cellcolor{green!60}0.00 & \cellcolor{green!60}0.00 & \cellcolor{green!60}0.00 & \cellcolor{green!60}0.00 \\
thick beard & \cellcolor{red!60}48.00 & \cellcolor{red!60}75.00 & \cellcolor{lightgreen!60}9.33 & \cellcolor{red!60}48.00 & \cellcolor{lightgreen!60}6.67 & \cellcolor{red!60}70.27 & \cellcolor{lightgreen!60}1.33 & \cellcolor{yellow!60}32.00 \\
red hair & \cellcolor{lightgreen!60}2.67 & \cellcolor{lightgreen!60}5.33 & \cellcolor{lightgreen!60}1.33 & \cellcolor{lightgreen!60}1.33 & \cellcolor{green!60}0.00 & \cellcolor{lightgreen!60}1.33 & \cellcolor{green!60}0.00 & \cellcolor{green!60}0.00 \\
goatee & \cellcolor{yellow!60}16.00 & \cellcolor{red!60}77.33 & \cellcolor{green!60}0.00 & \cellcolor{yellow!60}13.33 & \cellcolor{green!60}0.00 & \cellcolor{yellow!60}32.00 & \cellcolor{green!60}0.00 & \cellcolor{lightgreen!60}6.67 \\
\bottomrule
\end{tabular}
}
\caption{\emph{Face++} performance (FNMR @ FMR=0.1\%) using  \cface dataset}
    \label{tab:facepp_specific_fr2}
\end{table*}

\begin{table*}[ht]
    \centering
    \resizebox{0.65\textwidth}{!}{
\begin{tabular}{lrrrrrrrr}
    \toprule
 & AM & AF & BM & BF & WM & WF & IM & IF \\
\midrule
blond hair & \cellcolor{green!60}0.00 & \cellcolor{green!60}0.00 & \cellcolor{green!60}0.00 & \cellcolor{green!60}0.00 & \cellcolor{green!60}0.00 & \cellcolor{green!60}0.00 & \cellcolor{green!60}0.00 & \cellcolor{green!60}0.00 \\
bald & \cellcolor{lightgreen!60}4.00 & \cellcolor{lightgreen!60}6.67 & \cellcolor{green!60}0.00 & \cellcolor{green!60}0.00 & \cellcolor{green!60}0.00 & \cellcolor{green!60}0.00 & \cellcolor{green!60}0.00 & \cellcolor{green!60}0.00 \\
mustache & \cellcolor{yellow!60}22.67 & \cellcolor{red!60}80.00 & \cellcolor{lightgreen!60}1.33 & \cellcolor{yellow!60}13.33 & \cellcolor{lightgreen!60}1.33 & \cellcolor{yellow!60}25.33 & \cellcolor{lightgreen!60}1.33 & \cellcolor{yellow!60}30.67 \\
buzz cut & \cellcolor{lightgreen!60}6.67 & \cellcolor{yellow!60}14.67 & \cellcolor{green!60}0.00 & \cellcolor{green!60}0.00 & \cellcolor{green!60}0.00 & \cellcolor{lightgreen!60}4.00 & \cellcolor{green!60}0.00 & \cellcolor{green!60}0.00 \\
thick beard & \cellcolor{red!60}78.67 & \cellcolor{red!60}97.33 & \cellcolor{yellow!60}13.33 & \cellcolor{red!60}93.24 & \cellcolor{yellow!60}34.67 & \cellcolor{red!60}91.89 & \cellcolor{lightgreen!60}2.67 & \cellcolor{red!60}72.00 \\
red hair & \cellcolor{lightgreen!60}6.67 & \cellcolor{lightgreen!60}8.00 & \cellcolor{lightgreen!60}1.33 & \cellcolor{green!60}0.00 & \cellcolor{lightgreen!60}2.67 & \cellcolor{lightgreen!60}2.67 & \cellcolor{green!60}0.00 & \cellcolor{green!60}0.00 \\
goatee & \cellcolor{yellow!60}18.67 & \cellcolor{red!60}78.67 & \cellcolor{green!60}0.00 & \cellcolor{yellow!60}26.67 & \cellcolor{green!60}0.00 & \cellcolor{red!60}42.67 & \cellcolor{green!60}0.00 & \cellcolor{yellow!60}24.00 \\
\bottomrule
\end{tabular}
}
\caption{Confounding Factors Ablation: Face++ performance (FNMR@ FMR=0.1\%) using the dataset without specificity requirement.}
\label{tab:facepp_correctness_fr2}
\end{table*}

\begin{table*}[ht]
    \centering
    \resizebox{0.65\textwidth}{!}{
    \begin{tabular}{lrrrrrrrr}
\toprule
 & AM & AF & BM & BF & WM & WF & IM & IF \\
\midrule
blond hair & \cellcolor{green!60}0.00 & \cellcolor{green!60}0.00 & \cellcolor{green!60}0.00 & \cellcolor{green!60}0.00 & \cellcolor{green!60}0.00 & \cellcolor{green!60}0.00 & \cellcolor{green!60}0.00 & \cellcolor{green!60}0.00 \\
bald & \cellcolor{lightgreen!60}1.33 & \cellcolor{lightgreen!60}4.00 & \cellcolor{green!60}0.00 & \cellcolor{lightgreen!60}2.67 & \cellcolor{green!60}0.00 & \cellcolor{green!60}0.00 & \cellcolor{green!60}0.00 & \cellcolor{lightgreen!60}2.67 \\
mustache & \cellcolor{lightgreen!60}2.67 & \cellcolor{yellow!60}13.70 & \cellcolor{green!60}0.00 & \cellcolor{lightgreen!60}2.67 & \cellcolor{green!60}0.00 & \cellcolor{lightgreen!60}5.33 & \cellcolor{green!60}0.00 & \cellcolor{lightgreen!60}4.05 \\
buzz cut & \cellcolor{green!60}0.00 & \cellcolor{lightgreen!60}9.09 & \cellcolor{green!60}0.00 & \cellcolor{green!60}0.00 & \cellcolor{green!60}0.00 & \cellcolor{lightgreen!60}1.89 & \cellcolor{green!60}0.00 & \cellcolor{green!60}0.00 \\
thick beard & \cellcolor{yellow!60}32.43 & \cellcolor{red!60}54.00 & \cellcolor{lightgreen!60}5.33 & \cellcolor{yellow!60}28.38 & \cellcolor{lightgreen!60}2.67 & \cellcolor{yellow!60}39.19 & \cellcolor{lightgreen!60}1.33 & \cellcolor{yellow!60}17.33 \\
red hair & \cellcolor{green!60}0.00 & \cellcolor{lightgreen!60}2.67 & \cellcolor{lightgreen!60}2.67 & \cellcolor{green!60}0.00 & \cellcolor{green!60}0.00 & \cellcolor{lightgreen!60}1.33 & \cellcolor{green!60}0.00 & \cellcolor{green!60}0.00 \\
goatee & \cellcolor{green!60}0.00 & \cellcolor{red!60}48.00 & \cellcolor{lightgreen!60}1.37 & \cellcolor{lightgreen!60}5.33 & \cellcolor{green!60}0.00 & \cellcolor{yellow!60}14.86 & \cellcolor{green!60}0.00 & \cellcolor{lightgreen!60}5.33 \\
\bottomrule
\end{tabular}
}
\caption{\emph{AdaFace} performance (FNMR @ FMR=0.1\%) using  \cface dataset}
    \label{tab:adaface_ir101_msv2_specific_fr2}
\end{table*}

\begin{table*}[ht]
    \centering
    \resizebox{0.65\textwidth}{!}{
\begin{tabular}{lrrrrrrrr}
\toprule
 & AM & AF & BM & BF & WM & WF & IM & IF \\
\midrule
blond hair & \cellcolor{green!60}0.00 & \cellcolor{green!60}0.00 & \cellcolor{green!60}0.00 & \cellcolor{green!60}0.00 & \cellcolor{green!60}0.00 & \cellcolor{green!60}0.00 & \cellcolor{green!60}0.00 & \cellcolor{green!60}0.00 \\
bald & \cellcolor{lightgreen!60}2.67 & \cellcolor{green!60}0.00 & \cellcolor{green!60}0.00 & \cellcolor{lightgreen!60}1.33 & \cellcolor{green!60}0.00 & \cellcolor{green!60}0.00 & \cellcolor{green!60}0.00 & \cellcolor{lightgreen!60}5.33 \\
mustache & \cellcolor{lightgreen!60}4.05 & \cellcolor{yellow!60}20.83 & \cellcolor{lightgreen!60}1.33 & \cellcolor{lightgreen!60}4.00 & \cellcolor{green!60}0.00 & \cellcolor{lightgreen!60}6.85 & \cellcolor{green!60}0.00 & \cellcolor{lightgreen!60}6.67 \\
buzz cut & \cellcolor{lightgreen!60}6.67 & \cellcolor{yellow!60}12.00 & \cellcolor{green!60}0.00 & \cellcolor{green!60}0.00 & \cellcolor{green!60}0.00 & \cellcolor{lightgreen!60}4.00 & \cellcolor{green!60}0.00 & \cellcolor{green!60}0.00 \\
thick beard & \cellcolor{red!60}61.33 & \cellcolor{red!60}93.24 & \cellcolor{lightgreen!60}9.33 & \cellcolor{red!60}80.82 & \cellcolor{yellow!60}25.33 & \cellcolor{red!60}76.39 & \cellcolor{lightgreen!60}2.67 & \cellcolor{red!60}62.67 \\
red hair & \cellcolor{green!60}0.00 & \cellcolor{lightgreen!60}1.33 & \cellcolor{lightgreen!60}1.33 & \cellcolor{green!60}0.00 & \cellcolor{lightgreen!60}4.00 & \cellcolor{lightgreen!60}2.67 & \cellcolor{green!60}0.00 & \cellcolor{green!60}0.00 \\
goatee & \cellcolor{lightgreen!60}2.67 & \cellcolor{red!60}49.33 & \cellcolor{lightgreen!60}2.74 & \cellcolor{yellow!60}12.16 & \cellcolor{green!60}0.00 & \cellcolor{yellow!60}19.18 & \cellcolor{green!60}0.00 & \cellcolor{lightgreen!60}4.05 \\
\bottomrule
\end{tabular}
}
\caption{Confounding Factors Ablation: AdaFace performance (FNMR@ FMR=0.1\%) using the dataset without specificity requirement.}
\label{sec:adaface_correctness_v2}
\end{table*}

\begin{table*}[ht]
    \centering
    \resizebox{0.65\textwidth}{!}{
\begin{tabular}{lrrrrrrrr}
\toprule
 & AM & AF & BM & BF & WM & WF & IM & IF \\
\midrule
blond hair & \cellcolor{green!60}0.00 & \cellcolor{green!60}0.00 & \cellcolor{green!60}0.00 & \cellcolor{green!60}0.00 & \cellcolor{green!60}0.00 & \cellcolor{green!60}0.00 & \cellcolor{green!60}0.00 & \cellcolor{green!60}0.00 \\
bald & \cellcolor{lightgreen!60}4.00 & \cellcolor{lightgreen!60}6.67 & \cellcolor{lightgreen!60}1.33 & \cellcolor{yellow!60}10.67 & \cellcolor{lightgreen!60}1.33 & \cellcolor{lightgreen!60}6.67 & \cellcolor{lightgreen!60}2.67 & \cellcolor{yellow!60}33.33 \\
mustache & \cellcolor{lightgreen!60}6.67 & \cellcolor{yellow!60}32.00 & \cellcolor{green!60}0.00 & \cellcolor{lightgreen!60}2.67 & \cellcolor{lightgreen!60}1.33 & \cellcolor{lightgreen!60}5.33 & \cellcolor{green!60}0.00 & \cellcolor{lightgreen!60}6.67 \\
buzz cut & \cellcolor{green!60}0.00 & \cellcolor{lightgreen!60}9.09 & \cellcolor{green!60}0.00 & \cellcolor{green!60}0.00 & \cellcolor{green!60}0.00 & \cellcolor{lightgreen!60}3.77 & \cellcolor{green!60}0.00 & \cellcolor{green!60}0.00 \\
thick beard & \cellcolor{yellow!60}38.67 & \cellcolor{red!60}71.15 & \cellcolor{lightgreen!60}9.33 & \cellcolor{red!60}45.33 & \cellcolor{lightgreen!60}5.33 & \cellcolor{red!60}52.00 & \cellcolor{lightgreen!60}5.33 & \cellcolor{yellow!60}22.67 \\
red hair & \cellcolor{green!60}0.00 & \cellcolor{lightgreen!60}4.00 & \cellcolor{lightgreen!60}2.67 & \cellcolor{green!60}0.00 & \cellcolor{green!60}0.00 & \cellcolor{lightgreen!60}1.33 & \cellcolor{green!60}0.00 & \cellcolor{green!60}0.00 \\
goatee & \cellcolor{lightgreen!60}1.33 & \cellcolor{red!60}57.33 & \cellcolor{green!60}0.00 & \cellcolor{yellow!60}17.33 & \cellcolor{green!60}0.00 & \cellcolor{yellow!60}21.33 & \cellcolor{green!60}0.00 & \cellcolor{yellow!60}13.33 \\
\bottomrule
\end{tabular}
}
\caption{\emph{MagFace} performance (FNMR @ FMR=0.1\%) using  \cface dataset}
\label{tab:magface_specificity}
\end{table*}

\begin{table*}[ht]
    \centering
    \resizebox{0.65\textwidth}{!}{
\begin{tabular}{lrrrrrrrr}
\toprule
 & AM & AF & BM & BF & WM & WF & IM & IF \\
\midrule
blond hair & \cellcolor{green!60}0.00 & \cellcolor{green!60}0.00 & \cellcolor{green!60}0.00 & \cellcolor{green!60}0.00 & \cellcolor{green!60}0.00 & \cellcolor{green!60}0.00 & \cellcolor{green!60}0.00 & \cellcolor{green!60}0.00 \\
bald & \cellcolor{lightgreen!60}5.33 & \cellcolor{yellow!60}12.00 & \cellcolor{lightgreen!60}1.33 & \cellcolor{lightgreen!60}4.00 & \cellcolor{green!60}0.00 & \cellcolor{lightgreen!60}5.33 & \cellcolor{lightgreen!60}1.33 & \cellcolor{yellow!60}34.67 \\
mustache & \cellcolor{lightgreen!60}9.33 & \cellcolor{yellow!60}37.33 & \cellcolor{lightgreen!60}1.33 & \cellcolor{lightgreen!60}2.67 & \cellcolor{lightgreen!60}1.33 & \cellcolor{yellow!60}12.00 & \cellcolor{green!60}0.00 & \cellcolor{yellow!60}14.67 \\
buzz cut & \cellcolor{lightgreen!60}6.67 & \cellcolor{yellow!60}12.00 & \cellcolor{green!60}0.00 & \cellcolor{green!60}0.00 & \cellcolor{green!60}0.00 & \cellcolor{lightgreen!60}5.33 & \cellcolor{green!60}0.00 & \cellcolor{lightgreen!60}2.67 \\
thick beard & \cellcolor{red!60}76.00 & \cellcolor{red!60}97.33 & \cellcolor{yellow!60}14.67 & \cellcolor{red!60}93.33 & \cellcolor{yellow!60}29.33 & \cellcolor{red!60}84.00 & \cellcolor{lightgreen!60}9.33 & \cellcolor{red!60}64.00 \\
red hair & \cellcolor{lightgreen!60}1.33 & \cellcolor{lightgreen!60}2.67 & \cellcolor{lightgreen!60}1.33 & \cellcolor{green!60}0.00 & \cellcolor{lightgreen!60}4.00 & \cellcolor{lightgreen!60}2.67 & \cellcolor{lightgreen!60}1.33 & \cellcolor{green!60}0.00 \\
goatee & \cellcolor{lightgreen!60}5.33 & \cellcolor{red!60}65.33 & \cellcolor{green!60}0.00 & \cellcolor{yellow!60}28.00 & \cellcolor{green!60}0.00 & \cellcolor{yellow!60}36.00 & \cellcolor{green!60}0.00 & \cellcolor{yellow!60}21.33 \\
\bottomrule
\end{tabular}
}
\caption{Confounding Factors Ablation: MagFace performance (FNMR@ FMR=0.1\%) using the dataset without specificity requirement.}
\label{tab:magface_correctness_v2}
\end{table*}

\begin{table*}[]
    \centering
    \resizebox{0.65\textwidth}{!}{
    \begin{tabular}{lrrrrrrrr}
\toprule
 & AM & AF & BM & BF & WM & WF & IM & IF \\
\midrule
blond hair & \cellcolor{green!60}0.00 & \cellcolor{lightgreen!60}1.33 & \cellcolor{green!60}0.00 & \cellcolor{green!60}0.00 & \cellcolor{lightgreen!60}6.76 & \cellcolor{green!60}0.00 & \cellcolor{green!60}0.00 & \cellcolor{green!60}0.00 \\
bald & \cellcolor{red!60}53.33 & \cellcolor{red!60}56.00 & \cellcolor{yellow!60}32.00 & \cellcolor{yellow!60}41.33 & \cellcolor{red!60}73.33 & \cellcolor{red!60}78.67 & \cellcolor{red!60}58.67 & \cellcolor{red!60}76.00 \\
mustache & \cellcolor{lightgreen!60}8.00 & \cellcolor{yellow!60}22.67 & \cellcolor{green!60}0.00 & \cellcolor{yellow!60}14.67 & \cellcolor{lightgreen!60}6.67 & \cellcolor{yellow!60}24.00 & \cellcolor{lightgreen!60}5.33 & \cellcolor{yellow!60}24.00 \\
buzz cut & \cellcolor{lightgreen!60}2.67 & \cellcolor{lightgreen!60}9.09 & \cellcolor{green!60}0.00 & \cellcolor{green!60}0.00 & \cellcolor{green!60}0.00 & \cellcolor{lightgreen!60}3.77 & \cellcolor{green!60}0.00 & \cellcolor{lightgreen!60}5.48 \\
thick beard & \cellcolor{red!60}48.00 & \cellcolor{red!60}61.54 & \cellcolor{yellow!60}22.67 & \cellcolor{red!60}61.33 & \cellcolor{yellow!60}21.33 & \cellcolor{red!60}74.67 & \cellcolor{yellow!60}29.33 & \cellcolor{red!60}50.67 \\
red hair & \cellcolor{green!60}0.00 & \cellcolor{lightgreen!60}2.67 & \cellcolor{green!60}0.00 & \cellcolor{lightgreen!60}1.33 & \cellcolor{green!60}0.00 & \cellcolor{lightgreen!60}5.33 & \cellcolor{green!60}0.00 & \cellcolor{lightgreen!60}4.00 \\
goatee & \cellcolor{yellow!60}13.33 & \cellcolor{red!60}49.33 & \cellcolor{lightgreen!60}2.74 & \cellcolor{yellow!60}14.67 & \cellcolor{lightgreen!60}2.67 & \cellcolor{red!60}44.00 & \cellcolor{green!60}0.00 & \cellcolor{lightgreen!60}9.33 \\
\bottomrule
\end{tabular}
}
\caption{\emph{FaceNet} performance (FNMR @ FMR=0.1\%) using  \cface dataset}
\label{tab:facenet_specificity_fr}
\end{table*}

\begin{table*}[]
    \centering
    \resizebox{0.65\textwidth}{!}{
    \begin{tabular}{lrrrrrrrr}
\toprule
 & AM & AF & BM & BF & WM & WF & IM & IF \\
\midrule
blond hair & \cellcolor{green!60}0.00 & \cellcolor{lightgreen!60}1.33 & \cellcolor{green!60}0.00 & \cellcolor{green!60}0.00 & \cellcolor{lightgreen!60}6.67 & \cellcolor{green!60}0.00 & \cellcolor{green!60}0.00 & \cellcolor{green!60}0.00 \\
bald & \cellcolor{red!60}46.67 & \cellcolor{red!60}38.67 & \cellcolor{red!60}34.67 & \cellcolor{yellow!60}29.33 & \cellcolor{red!60}78.67 & \cellcolor{red!60}53.33 & \cellcolor{red!60}44.00 & \cellcolor{red!60}81.33 \\
mustache & \cellcolor{yellow!60}14.67 & \cellcolor{yellow!60}30.67 & \cellcolor{lightgreen!60}1.33 & \cellcolor{yellow!60}17.33 & \cellcolor{lightgreen!60}6.67 & \cellcolor{yellow!60}30.67 & \cellcolor{lightgreen!60}6.67 & \cellcolor{yellow!60}34.67 \\
buzz cut & \cellcolor{lightgreen!60}6.67 & \cellcolor{yellow!60}13.33 & \cellcolor{green!60}0.00 & \cellcolor{lightgreen!60}1.33 & \cellcolor{green!60}0.00 & \cellcolor{yellow!60}14.67 & \cellcolor{green!60}0.00 & \cellcolor{lightgreen!60}4.00 \\
thick beard & \cellcolor{red!60}76.00 & \cellcolor{red!60}96.00 & \cellcolor{yellow!60}24.00 & \cellcolor{red!60}93.33 & \cellcolor{red!60}48.00 & \cellcolor{red!60}93.33 & \cellcolor{yellow!60}29.33 & \cellcolor{red!60}85.33 \\
red hair & \cellcolor{lightgreen!60}1.33 & \cellcolor{lightgreen!60}1.33 & \cellcolor{lightgreen!60}1.33 & \cellcolor{lightgreen!60}2.67 & \cellcolor{lightgreen!60}9.33 & \cellcolor{yellow!60}10.67 & \cellcolor{lightgreen!60}4.00 & \cellcolor{lightgreen!60}1.33 \\
goatee & \cellcolor{yellow!60}14.67 & \cellcolor{red!60}53.33 & \cellcolor{lightgreen!60}2.74 & \cellcolor{yellow!60}33.33 & \cellcolor{lightgreen!60}2.67 & \cellcolor{red!60}54.67 & \cellcolor{lightgreen!60}1.33 & \cellcolor{yellow!60}18.67 \\
\bottomrule
\end{tabular}
}
\caption{Confounding Factors Ablation: FaceNet performance (FNMR@ FMR=0.1\%) using the dataset without specificity requirement.}
\label{tab:facenet_correctness_fr}
\end{table*}

\begin{table*}[]
    \centering
    \resizebox{0.65\textwidth}{!}{
    \begin{tabular}{lrrrrrrrr}
\toprule
 & AM & AF & BM & BF & WM & WF & IM & IF \\
\midrule
blond hair & \cellcolor{yellow!60}10.67 & \cellcolor{lightgreen!60}9.33 & \cellcolor{lightgreen!60}5.33 & \cellcolor{lightgreen!60}2.67 & \cellcolor{lightgreen!60}5.41 & \cellcolor{lightgreen!60}6.67 & \cellcolor{lightgreen!60}8.00 & \cellcolor{lightgreen!60}6.67 \\
bald & \cellcolor{red!60}54.67 & \cellcolor{red!60}68.00 & \cellcolor{yellow!60}32.00 & \cellcolor{red!60}46.67 & \cellcolor{yellow!60}33.33 & \cellcolor{red!60}54.67 & \cellcolor{red!60}54.67 & \cellcolor{red!60}80.00 \\
mustache & \cellcolor{yellow!60}17.33 & \cellcolor{red!60}62.67 & \cellcolor{lightgreen!60}9.33 & \cellcolor{yellow!60}18.67 & \cellcolor{lightgreen!60}5.33 & \cellcolor{yellow!60}26.67 & \cellcolor{yellow!60}13.33 & \cellcolor{yellow!60}21.33 \\
buzz cut & \cellcolor{lightgreen!60}6.67 & \cellcolor{yellow!60}18.18 & \cellcolor{lightgreen!60}5.33 & \cellcolor{yellow!60}14.67 & \cellcolor{green!60}0.00 & \cellcolor{yellow!60}18.87 & \cellcolor{lightgreen!60}6.67 & \cellcolor{red!60}35.62 \\
thick beard & \cellcolor{red!60}61.33 & \cellcolor{red!60}82.69 & \cellcolor{yellow!60}30.67 & \cellcolor{red!60}68.00 & \cellcolor{yellow!60}18.67 & \cellcolor{red!60}70.67 & \cellcolor{yellow!60}36.00 & \cellcolor{red!60}65.33 \\
red hair & \cellcolor{lightgreen!60}5.33 & \cellcolor{yellow!60}12.00 & \cellcolor{yellow!60}10.67 & \cellcolor{lightgreen!60}8.00 & \cellcolor{lightgreen!60}2.67 & \cellcolor{lightgreen!60}6.67 & \cellcolor{yellow!60}14.67 & \cellcolor{yellow!60}14.67 \\
goatee & \cellcolor{yellow!60}12.00 & \cellcolor{red!60}58.67 & \cellcolor{lightgreen!60}2.74 & \cellcolor{yellow!60}28.00 & \cellcolor{lightgreen!60}9.33 & \cellcolor{yellow!60}29.33 & \cellcolor{lightgreen!60}6.67 & \cellcolor{yellow!60}14.67 \\
\bottomrule
\end{tabular}
}
\caption{\emph{ArcFace} performance (FNMR @ FMR=0.1\%) using  \cface dataset}
\label{tab:arcface_specificity_fr}
\end{table*}

\begin{table*}[]
    \centering
    \resizebox{0.65\textwidth}{!}{
    \begin{tabular}{lrrrrrrrr}
\toprule
 & AM & AF & BM & BF & WM & WF & IM & IF \\
\midrule
blond hair & \cellcolor{lightgreen!60}9.33 & \cellcolor{yellow!60}10.67 & \cellcolor{lightgreen!60}1.33 & \cellcolor{lightgreen!60}2.67 & \cellcolor{lightgreen!60}5.33 & \cellcolor{lightgreen!60}6.67 & \cellcolor{lightgreen!60}8.00 & \cellcolor{lightgreen!60}6.67 \\
bald & \cellcolor{red!60}57.33 & \cellcolor{red!60}70.67 & \cellcolor{yellow!60}21.33 & \cellcolor{red!60}48.00 & \cellcolor{yellow!60}36.00 & \cellcolor{yellow!60}37.33 & \cellcolor{yellow!60}42.67 & \cellcolor{red!60}80.00 \\
mustache & \cellcolor{yellow!60}24.00 & \cellcolor{red!60}64.00 & \cellcolor{lightgreen!60}9.33 & \cellcolor{yellow!60}18.67 & \cellcolor{lightgreen!60}8.00 & \cellcolor{yellow!60}32.00 & \cellcolor{yellow!60}10.67 & \cellcolor{yellow!60}30.67 \\
buzz cut & \cellcolor{yellow!60}13.33 & \cellcolor{red!60}33.33 & \cellcolor{lightgreen!60}4.00 & \cellcolor{yellow!60}16.00 & \cellcolor{lightgreen!60}5.33 & \cellcolor{yellow!60}30.67 & \cellcolor{lightgreen!60}9.33 & \cellcolor{red!60}38.67 \\
thick beard & \cellcolor{red!60}85.33 & \cellcolor{red!60}97.33 & \cellcolor{yellow!60}40.00 & \cellcolor{red!60}96.00 & \cellcolor{yellow!60}42.67 & \cellcolor{red!60}92.00 & \cellcolor{yellow!60}29.33 & \cellcolor{red!60}84.00 \\
red hair & \cellcolor{lightgreen!60}8.00 & \cellcolor{lightgreen!60}8.00 & \cellcolor{yellow!60}16.00 & \cellcolor{lightgreen!60}5.33 & \cellcolor{lightgreen!60}6.67 & \cellcolor{lightgreen!60}9.33 & \cellcolor{yellow!60}13.33 & \cellcolor{yellow!60}13.33 \\
goatee & \cellcolor{yellow!60}13.33 & \cellcolor{red!60}64.00 & \cellcolor{yellow!60}10.67 & \cellcolor{yellow!60}34.67 & \cellcolor{lightgreen!60}9.33 & \cellcolor{yellow!60}44.00 & \cellcolor{yellow!60}10.67 & \cellcolor{yellow!60}33.33 \\
\bottomrule
\end{tabular}
}
\caption{Confounding Factors Ablation: ArcFace performance (FNMR@ FMR=0.1\%) using the dataset without specificity requirement.}
\label{tab:arcface_correctness_fr}
\end{table*}

\begin{table*}[ht]
    \centering
    \resizebox{0.65\textwidth}{!}{
\begin{tabular}{lrrrrrrrr}
\toprule
 & AM & AF & BM & BF & WM & WF & IM & IF \\
\midrule
blond hair & \cellcolor{green!60}0.00 & \cellcolor{green!60}0.00 & \cellcolor{green!60}0.00 & \cellcolor{green!60}0.00 & \cellcolor{green!60}0.00 & \cellcolor{green!60}0.00 & \cellcolor{green!60}0.00 & \cellcolor{green!60}0.00 \\
bald & \cellcolor{lightgreen!60}6.67 & \cellcolor{lightgreen!60}6.67 & \cellcolor{green!60}0.00 & \cellcolor{lightgreen!60}4.00 & \cellcolor{green!60}0.00 & \cellcolor{lightgreen!60}5.33 & \cellcolor{green!60}0.00 & \cellcolor{yellow!60}22.67 \\
mustache & \cellcolor{lightgreen!60}2.67 & \cellcolor{yellow!60}16.44 & \cellcolor{green!60}0.00 & \cellcolor{lightgreen!60}2.67 & \cellcolor{green!60}0.00 & \cellcolor{lightgreen!60}6.67 & \cellcolor{green!60}0.00 & \cellcolor{lightgreen!60}5.41 \\
buzz cut & \cellcolor{green!60}0.00 & \cellcolor{lightgreen!60}9.09 & \cellcolor{green!60}0.00 & \cellcolor{green!60}0.00 & \cellcolor{green!60}0.00 & \cellcolor{lightgreen!60}1.89 & \cellcolor{green!60}0.00 & \cellcolor{green!60}0.00 \\
thick beard & \cellcolor{yellow!60}28.38 & \cellcolor{red!60}58.00 & \cellcolor{lightgreen!60}2.67 & \cellcolor{yellow!60}31.08 & \cellcolor{lightgreen!60}2.67 & \cellcolor{red!60}50.00 & \cellcolor{lightgreen!60}4.00 & \cellcolor{yellow!60}17.33 \\
red hair & \cellcolor{green!60}0.00 & \cellcolor{lightgreen!60}2.67 & \cellcolor{lightgreen!60}2.67 & \cellcolor{green!60}0.00 & \cellcolor{green!60}0.00 & \cellcolor{lightgreen!60}1.33 & \cellcolor{green!60}0.00 & \cellcolor{green!60}0.00 \\
goatee & \cellcolor{lightgreen!60}1.33 & \cellcolor{red!60}49.33 & \cellcolor{lightgreen!60}1.37 & \cellcolor{yellow!60}10.67 & \cellcolor{green!60}0.00 & \cellcolor{yellow!60}13.51 & \cellcolor{green!60}0.00 & \cellcolor{lightgreen!60}5.33 \\
\bottomrule
\end{tabular}
}
\caption{Model Size Ablation: AdaFace-50-MS performance (FNMR @ FMR=0.1\%) using \cface.}
\label{tab:adaface_model_size_ablation}
\end{table*}

\begin{table*}[ht]
    \centering
    \resizebox{0.65\textwidth}{!}{
\begin{tabular}{lrrrrrrrr}
\toprule
 & AM & AF & BM & BF & WM & WF & IM & IF \\
\midrule
blond hair & \cellcolor{green!60}0.00 & \cellcolor{green!60}0.00 & \cellcolor{green!60}0.00 & \cellcolor{green!60}0.00 & \cellcolor{green!60}0.00 & \cellcolor{green!60}0.00 & \cellcolor{green!60}0.00 & \cellcolor{green!60}0.00 \\
bald & \cellcolor{lightgreen!60}2.67 & \cellcolor{lightgreen!60}6.67 & \cellcolor{green!60}0.00 & \cellcolor{yellow!60}10.67 & \cellcolor{green!60}0.00 & \cellcolor{lightgreen!60}5.33 & \cellcolor{green!60}0.00 & \cellcolor{yellow!60}13.33 \\
mustache & \cellcolor{lightgreen!60}2.67 & \cellcolor{lightgreen!60}9.59 & \cellcolor{green!60}0.00 & \cellcolor{green!60}0.00 & \cellcolor{green!60}0.00 & \cellcolor{lightgreen!60}2.67 & \cellcolor{green!60}0.00 & \cellcolor{lightgreen!60}4.05 \\
buzz cut & \cellcolor{green!60}0.00 & \cellcolor{lightgreen!60}7.27 & \cellcolor{green!60}0.00 & \cellcolor{green!60}0.00 & \cellcolor{green!60}0.00 & \cellcolor{lightgreen!60}1.89 & \cellcolor{green!60}0.00 & \cellcolor{green!60}0.00 \\
thick beard & \cellcolor{yellow!60}31.08 & \cellcolor{red!60}58.00 & \cellcolor{lightgreen!60}2.67 & \cellcolor{yellow!60}29.73 & \cellcolor{lightgreen!60}2.67 & \cellcolor{red!60}41.89 & \cellcolor{lightgreen!60}1.33 & \cellcolor{yellow!60}16.00 \\
red hair & \cellcolor{green!60}0.00 & \cellcolor{lightgreen!60}2.67 & \cellcolor{lightgreen!60}2.67 & \cellcolor{green!60}0.00 & \cellcolor{green!60}0.00 & \cellcolor{lightgreen!60}1.33 & \cellcolor{green!60}0.00 & \cellcolor{green!60}0.00 \\
goatee & \cellcolor{green!60}0.00 & \cellcolor{yellow!60}37.33 & \cellcolor{lightgreen!60}1.37 & \cellcolor{lightgreen!60}4.00 & \cellcolor{green!60}0.00 & \cellcolor{yellow!60}13.51 & \cellcolor{green!60}0.00 & \cellcolor{lightgreen!60}2.67 \\
\bottomrule
\end{tabular}
}
\caption{Dataset Size Ablation: AdaFace-100-Web4m performance (FNMR @ FMR=0.1\%) using \cface.}
\label{tab:adaface_dataset_size_ablation}
\end{table*}

\begin{table*}[]
    \centering
    \resizebox{0.65\textwidth}{!}{
\begin{tabular}{lrrrrrrrr}
\toprule
 & AM & AF & BM & BF & WM & WF & IM & IF \\
\midrule
blond hair & \cellcolor{green!60}0.00 & \cellcolor{green!60}0.00 & \cellcolor{green!60}0.00 & \cellcolor{green!60}0.00 & \cellcolor{green!60}0.00 & \cellcolor{green!60}0.00 & \cellcolor{green!60}0.00 & \cellcolor{green!60}0.00 \\
bald & \cellcolor{green!60}0.00 & \cellcolor{green!60}0.00 & \cellcolor{green!60}0.00 & \cellcolor{green!60}0.00 & \cellcolor{green!60}0.00 & \cellcolor{green!60}0.00 & \cellcolor{green!60}0.00 & \cellcolor{green!60}0.00 \\
mustache & \cellcolor{lightgreen!60}1.33 & \cellcolor{lightgreen!60}1.33 & \cellcolor{green!60}0.00 & \cellcolor{green!60}0.00 & \cellcolor{green!60}0.00 & \cellcolor{green!60}0.00 & \cellcolor{green!60}0.00 & \cellcolor{green!60}0.00 \\
buzz cut & \cellcolor{green!60}0.00 & \cellcolor{lightgreen!60}3.64 & \cellcolor{green!60}0.00 & \cellcolor{green!60}0.00 & \cellcolor{green!60}0.00 & \cellcolor{green!60}0.00 & \cellcolor{green!60}0.00 & \cellcolor{green!60}0.00 \\
thick beard & \cellcolor{lightgreen!60}6.67 & \cellcolor{red!60}44.23 & \cellcolor{green!60}0.00 & \cellcolor{yellow!60}17.33 & \cellcolor{green!60}0.00 & \cellcolor{yellow!60}29.33 & \cellcolor{green!60}0.00 & \cellcolor{yellow!60}13.33 \\
red hair & \cellcolor{green!60}0.00 & \cellcolor{lightgreen!60}1.33 & \cellcolor{green!60}0.00 & \cellcolor{green!60}0.00 & \cellcolor{green!60}0.00 & \cellcolor{lightgreen!60}1.33 & \cellcolor{green!60}0.00 & \cellcolor{green!60}0.00 \\
goatee & \cellcolor{green!60}0.00 & \cellcolor{yellow!60}29.33 & \cellcolor{green!60}0.00 & \cellcolor{green!60}0.00 & \cellcolor{green!60}0.00 & \cellcolor{lightgreen!60}6.67 & \cellcolor{green!60}0.00 & \cellcolor{lightgreen!60}2.67 \\
\bottomrule
\end{tabular}}
\caption{\emph{AWS} performance (FNMR @ FMR=1\%) using  \cface dataset}
\label{tab:aws_specificity_fmr=1}
\end{table*}

\begin{table*}[]
    \centering
    \resizebox{0.65\textwidth}{!}{
\begin{tabular}{lrrrrrrrr}
\toprule
 & AM & AF & BM & BF & WM & WF & IM & IF \\
\midrule
blond hair & \cellcolor{green!60}0.00 & \cellcolor{green!60}0.00 & \cellcolor{green!60}0.00 & \cellcolor{green!60}0.00 & \cellcolor{green!60}0.00 & \cellcolor{green!60}0.00 & \cellcolor{green!60}0.00 & \cellcolor{green!60}0.00 \\
bald & \cellcolor{green!60}0.00 & \cellcolor{green!60}0.00 & \cellcolor{green!60}0.00 & \cellcolor{green!60}0.00 & \cellcolor{green!60}0.00 & \cellcolor{green!60}0.00 & \cellcolor{green!60}0.00 & \cellcolor{green!60}0.00 \\
mustache & \cellcolor{lightgreen!60}1.33 & \cellcolor{yellow!60}20.00 & \cellcolor{green!60}0.00 & \cellcolor{green!60}0.00 & \cellcolor{green!60}0.00 & \cellcolor{lightgreen!60}2.67 & \cellcolor{green!60}0.00 & \cellcolor{lightgreen!60}1.33 \\
buzz cut & \cellcolor{green!60}0.00 & \cellcolor{lightgreen!60}5.45 & \cellcolor{green!60}0.00 & \cellcolor{green!60}0.00 & \cellcolor{green!60}0.00 & \cellcolor{green!60}0.00 & \cellcolor{green!60}0.00 & \cellcolor{green!60}0.00 \\
thick beard & \cellcolor{yellow!60}34.67 & \cellcolor{red!60}53.85 & \cellcolor{green!60}0.00 & \cellcolor{yellow!60}20.00 & \cellcolor{lightgreen!60}1.33 & \cellcolor{yellow!60}35.14 & \cellcolor{green!60}0.00 & \cellcolor{yellow!60}13.33 \\
red hair & \cellcolor{green!60}0.00 & \cellcolor{lightgreen!60}1.33 & \cellcolor{green!60}0.00 & \cellcolor{green!60}0.00 & \cellcolor{green!60}0.00 & \cellcolor{lightgreen!60}1.33 & \cellcolor{green!60}0.00 & \cellcolor{green!60}0.00 \\
goatee & \cellcolor{lightgreen!60}2.67 & \cellcolor{red!60}48.00 & \cellcolor{green!60}0.00 & \cellcolor{lightgreen!60}1.33 & \cellcolor{green!60}0.00 & \cellcolor{yellow!60}10.67 & \cellcolor{green!60}0.00 & \cellcolor{lightgreen!60}2.67 \\
\bottomrule
\end{tabular}}
\caption{\emph{Face++} performance (FNMR @ FMR=1\%) using  \cface dataset}
\label{tab:facepp_specificity_fmr=1}
\end{table*}

\begin{table*}[ht]
    \centering
    \resizebox{0.65\textwidth}{!}{
   \begin{tabular}{lrrrrrrrr} \toprule & AM & AF & BM & BF & WM & WF & IM & IF \\ \midrule blond hair & \cellcolor{green!60}0.00 & \cellcolor{green!60}0.00 & \cellcolor{green!60}0.00 & \cellcolor{green!60}0.00 & \cellcolor{green!60}0.00 & \cellcolor{green!60}0.00 & \cellcolor{green!60}0.00 & \cellcolor{green!60}0.00 \\ bald & \cellcolor{green!60}0.00 & \cellcolor{green!60}0.00 & \cellcolor{green!60}0.00 & \cellcolor{green!60}0.00 & \cellcolor{green!60}0.00 & \cellcolor{green!60}0.00 & \cellcolor{green!60}0.00 & \cellcolor{green!60}0.00 \\ mustache & \cellcolor{green!60}0.00 & \cellcolor{lightgreen!60}4.11 & \cellcolor{green!60}0.00 & \cellcolor{green!60}0.00 & \cellcolor{green!60}0.00 & \cellcolor{green!60}0.00 & \cellcolor{green!60}0.00 & \cellcolor{green!60}0.00 \\ buzz cut & \cellcolor{green!60}0.00 & \cellcolor{lightgreen!60}5.45 & \cellcolor{green!60}0.00 & \cellcolor{green!60}0.00 & \cellcolor{green!60}0.00 & \cellcolor{lightgreen!60}1.89 & \cellcolor{green!60}0.00 & \cellcolor{green!60}0.00 \\ thick beard & \cellcolor{lightgreen!60}4.05 & \cellcolor{yellow!60}38.00 & \cellcolor{green!60}0.00 & \cellcolor{yellow!60}10.81 & \cellcolor{green!60}0.00 & \cellcolor{yellow!60}18.92 & \cellcolor{green!60}0.00 & \cellcolor{lightgreen!60}2.67 \\ red hair & \cellcolor{green!60}0.00 & \cellcolor{green!60}0.00 & \cellcolor{green!60}0.00 & \cellcolor{green!60}0.00 & \cellcolor{green!60}0.00 & \cellcolor{lightgreen!60}1.33 & \cellcolor{green!60}0.00 & \cellcolor{green!60}0.00 \\ goatee & \cellcolor{green!60}0.00 & \cellcolor{yellow!60}18.67 & \cellcolor{lightgreen!60}1.37 & \cellcolor{green!60}0.00 & \cellcolor{green!60}0.00 & \cellcolor{lightgreen!60}6.76 & \cellcolor{green!60}0.00 & \cellcolor{green!60}0.00 \\
\bottomrule \end{tabular}
}
\caption{\emph{AdaFace} performance (FNMR @ FMR=1\%) using  \cface dataset}
\label{tab:adaface_ir101_msv2_specific_fmr=1}
\end{table*}

\begin{table*}[ht]
    \centering
    \resizebox{0.65\textwidth}{!}{
\begin{tabular}{lrrrrrrrr}
\toprule
 & AM & AF & BM & BF & WM & WF & IM & IF \\
\midrule
blond hair & \cellcolor{green!60}0.00 & \cellcolor{green!60}0.00 & \cellcolor{green!60}0.00 & \cellcolor{green!60}0.00 & \cellcolor{green!60}0.00 & \cellcolor{green!60}0.00 & \cellcolor{green!60}0.00 & \cellcolor{green!60}0.00 \\
bald & \cellcolor{green!60}0.00 & \cellcolor{lightgreen!60}1.33 & \cellcolor{green!60}0.00 & \cellcolor{green!60}0.00 & \cellcolor{green!60}0.00 & \cellcolor{green!60}0.00 & \cellcolor{lightgreen!60}1.33 & \cellcolor{lightgreen!60}1.33 \\
mustache & \cellcolor{lightgreen!60}1.33 & \cellcolor{lightgreen!60}5.33 & \cellcolor{green!60}0.00 & \cellcolor{green!60}0.00 & \cellcolor{lightgreen!60}1.33 & \cellcolor{green!60}0.00 & \cellcolor{green!60}0.00 & \cellcolor{green!60}0.00 \\
buzz cut & \cellcolor{green!60}0.00 & \cellcolor{lightgreen!60}5.45 & \cellcolor{green!60}0.00 & \cellcolor{green!60}0.00 & \cellcolor{green!60}0.00 & \cellcolor{green!60}0.00 & \cellcolor{green!60}0.00 & \cellcolor{green!60}0.00 \\
thick beard & \cellcolor{yellow!60}10.67 & \cellcolor{red!60}51.92 & \cellcolor{lightgreen!60}1.33 & \cellcolor{yellow!60}20.00 & \cellcolor{lightgreen!60}1.33 & \cellcolor{yellow!60}32.00 & \cellcolor{green!60}0.00 & \cellcolor{yellow!60}10.67 \\
red hair & \cellcolor{green!60}0.00 & \cellcolor{lightgreen!60}2.67 & \cellcolor{green!60}0.00 & \cellcolor{green!60}0.00 & \cellcolor{green!60}0.00 & \cellcolor{lightgreen!60}1.33 & \cellcolor{green!60}0.00 & \cellcolor{green!60}0.00 \\
goatee & \cellcolor{green!60}0.00 & \cellcolor{yellow!60}21.33 & \cellcolor{green!60}0.00 & \cellcolor{lightgreen!60}2.67 & \cellcolor{green!60}0.00 & \cellcolor{lightgreen!60}9.33 & \cellcolor{green!60}0.00 & \cellcolor{lightgreen!60}1.33 \\
\bottomrule
\end{tabular}
}

\caption{\emph{MagFace} performance (FNMR @ FMR=1\%) using  \cface dataset}
\label{tab:magface_specificity_fmr=1}
\end{table*}

\begin{table*}[]
    \centering
    \resizebox{0.65\textwidth}{!}{
   \begin{tabular}{lrrrrrrrr}
\toprule
 & AM & AF & BM & BF & WM & WF & IM & IF \\
\midrule
blond hair & \cellcolor{green!60}0.00 & \cellcolor{green!60}0.00 & \cellcolor{green!60}0.00 & \cellcolor{green!60}0.00 & \cellcolor{green!60}0.00 & \cellcolor{green!60}0.00 & \cellcolor{green!60}0.00 & \cellcolor{green!60}0.00 \\
bald & \cellcolor{lightgreen!60}1.33 & \cellcolor{lightgreen!60}2.67 & \cellcolor{lightgreen!60}1.33 & \cellcolor{lightgreen!60}2.67 & \cellcolor{yellow!60}10.67 & \cellcolor{yellow!60}22.67 & \cellcolor{lightgreen!60}1.33 & \cellcolor{yellow!60}17.33 \\
mustache & \cellcolor{green!60}0.00 & \cellcolor{lightgreen!60}4.00 & \cellcolor{green!60}0.00 & \cellcolor{lightgreen!60}2.67 & \cellcolor{lightgreen!60}1.33 & \cellcolor{lightgreen!60}2.67 & \cellcolor{lightgreen!60}1.33 & \cellcolor{green!60}0.00 \\
buzz cut & \cellcolor{green!60}0.00 & \cellcolor{lightgreen!60}3.64 & \cellcolor{green!60}0.00 & \cellcolor{green!60}0.00 & \cellcolor{green!60}0.00 & \cellcolor{lightgreen!60}1.89 & \cellcolor{green!60}0.00 & \cellcolor{lightgreen!60}1.37 \\
thick beard & \cellcolor{yellow!60}20.00 & \cellcolor{red!60}48.08 & \cellcolor{lightgreen!60}2.67 & \cellcolor{yellow!60}29.33 & \cellcolor{lightgreen!60}5.33 & \cellcolor{red!60}41.33 & \cellcolor{lightgreen!60}2.67 & \cellcolor{yellow!60}14.67 \\
red hair & \cellcolor{green!60}0.00 & \cellcolor{green!60}0.00 & \cellcolor{green!60}0.00 & \cellcolor{green!60}0.00 & \cellcolor{green!60}0.00 & \cellcolor{lightgreen!60}1.33 & \cellcolor{green!60}0.00 & \cellcolor{green!60}0.00 \\
goatee & \cellcolor{green!60}0.00 & \cellcolor{yellow!60}18.67 & \cellcolor{green!60}0.00 & \cellcolor{lightgreen!60}4.00 & \cellcolor{lightgreen!60}2.67 & \cellcolor{yellow!60}13.33 & \cellcolor{green!60}0.00 & \cellcolor{lightgreen!60}1.33 \\
\bottomrule
\end{tabular}
}
\caption{\emph{FaceNet} performance (FNMR @ FMR=1\%) using  \cface dataset}
\label{tab:facenet_specificity_fmr=1}
\end{table*}

\begin{table*}[ht]
    \centering
    \resizebox{0.65\textwidth}{!}{
   \begin{tabular}{lrrrrrrrr}
\toprule
 & AM & AF & BM & BF & WM & WF & IM & IF \\
\midrule
blond hair & \cellcolor{lightgreen!60}2.67 & \cellcolor{green!60}0.00 & \cellcolor{lightgreen!60}1.33 & \cellcolor{green!60}0.00 & \cellcolor{green!60}0.00 & \cellcolor{lightgreen!60}2.67 & \cellcolor{lightgreen!60}1.33 & \cellcolor{lightgreen!60}1.33 \\
bald & \cellcolor{lightgreen!60}9.33 & \cellcolor{yellow!60}20.00 & \cellcolor{lightgreen!60}2.67 & \cellcolor{yellow!60}10.67 & \cellcolor{lightgreen!60}1.33 & \cellcolor{yellow!60}13.33 & \cellcolor{lightgreen!60}9.33 & \cellcolor{yellow!60}14.67 \\
mustache & \cellcolor{lightgreen!60}4.00 & \cellcolor{yellow!60}12.00 & \cellcolor{green!60}0.00 & \cellcolor{green!60}0.00 & \cellcolor{lightgreen!60}1.33 & \cellcolor{lightgreen!60}1.33 & \cellcolor{lightgreen!60}2.67 & \cellcolor{lightgreen!60}2.67 \\
buzz cut & \cellcolor{lightgreen!60}1.33 & \cellcolor{lightgreen!60}5.45 & \cellcolor{green!60}0.00 & \cellcolor{lightgreen!60}2.67 & \cellcolor{green!60}0.00 & \cellcolor{lightgreen!60}5.66 & \cellcolor{lightgreen!60}1.33 & \cellcolor{lightgreen!60}5.48 \\
thick beard & \cellcolor{yellow!60}10.67 & \cellcolor{red!60}46.15 & \cellcolor{lightgreen!60}2.67 & \cellcolor{yellow!60}17.33 & \cellcolor{lightgreen!60}1.33 & \cellcolor{yellow!60}28.00 & \cellcolor{lightgreen!60}5.33 & \cellcolor{yellow!60}17.33 \\
red hair & \cellcolor{lightgreen!60}1.33 & \cellcolor{lightgreen!60}2.67 & \cellcolor{lightgreen!60}1.33 & \cellcolor{green!60}0.00 & \cellcolor{green!60}0.00 & \cellcolor{lightgreen!60}2.67 & \cellcolor{green!60}0.00 & \cellcolor{lightgreen!60}1.33 \\
goatee & \cellcolor{lightgreen!60}1.33 & \cellcolor{yellow!60}14.67 & \cellcolor{lightgreen!60}1.37 & \cellcolor{green!60}0.00 & \cellcolor{lightgreen!60}1.33 & \cellcolor{lightgreen!60}2.67 & \cellcolor{lightgreen!60}1.33 & \cellcolor{lightgreen!60}1.33 \\
\bottomrule
\end{tabular}
}
\caption{\emph{ArcFace} performance (FNMR @ FMR=1\%) using  \cface dataset}
\label{tab:arcface_specificity_fmr=1}
\end{table*}

\end{document}